\documentclass[10pt]{article} 
\usepackage[accepted]{tmlr}

\usepackage{amsmath,amsfonts,bm}

\def\eqref#1{equation~\ref{#1}}

\def\1{\bm{1}}

\DeclareMathAlphabet{\mathsfit}{\encodingdefault}{\sfdefault}{m}{sl}
\SetMathAlphabet{\mathsfit}{bold}{\encodingdefault}{\sfdefault}{bx}{n}

\newcommand{\mcS}{\mathcal{S}}
\newcommand{\mcO}{\mathcal{O}}

\newcommand{\bfx}{\mathbf{x}}
\newcommand{\bfy}{\mathbf{y}}
\newcommand{\bfo}{\mathbf{o}}

\newcommand{\pretrain}{p_\text{pre-train}}
\newcommand{\xtest}{\mathbf{x}_{\text{test}}}
\newcommand{\delim}{o^\text{delim}}

\usepackage{hyperref}
\usepackage{url}
\usepackage{xparse}
\usepackage{graphicx}
\usepackage[normalem]{ulem}
\usepackage{subcaption}
\usepackage{amsthm}
\usepackage{amsmath}
\newtheorem{assumption}{Assumption}
\newtheorem{theorem}{Theorem}

\usepackage{xparse}
\usepackage{wrapfig}
\usepackage{booktabs}
 \usepackage{multirow}

\title{Understanding Emergent In-Context Learning from a Kernel Regression Perspective}

\author{\name Chi Han \email chihan3@illinois.edu \\
       \name Ziqi Wang \email ziqiw9@illinois.edu \\
       \name Han Zhao \email hanzhao@illinois.edu \\
       \name Heng Ji \email hengji@illinois.edu \\
       \addr Siebel School of Computing and Data Science \\
       University of Illinois Urbana-Champaign
}

\def\month{09}  
\def\year{2025} 
\def\openreview{\url{https://openreview.net/forum?id=6rD50Q6yYz}} 

\begin{document}

\maketitle
\NewDocumentCommand{\heng}
{ mO{} }{\textcolor{red}{\textsuperscript{\textit{heng}}\textsf{\textbf{\small[#1]}}}}
\NewDocumentCommand{\hengsolved}
{ mO{} }{\textcolor{red}{\textsuperscript{\textit{heng}}\textsf{\sout{\small[#1]}}}}
\NewDocumentCommand{\hengarchived}
{ mO{} }{}
\NewDocumentCommand{\chihan}
{ mO{} }{\textcolor{blue}{\textsuperscript{\textit{chi}}[#1]}}
\NewDocumentCommand{\han}
{ mO{} }{\textcolor{magenta}{\textsuperscript{\textit{Han}}[#1]}}
\NewDocumentCommand{\pengfei}
{ mO{} }{\textcolor{teal}{\textsuperscript{\textit{Pengfei}}[#1]}}

\begin{abstract}
    Large language models (LLMs) have initiated a paradigm shift in transfer learning. In contrast to the classic pretraining-then-finetuning procedure, in order to use LLMs for downstream prediction tasks, one only needs to provide a few demonstrations, known as in-context examples, without adding more or updating existing model parameters. This in-context learning (ICL) capability of LLMs is intriguing, and it is not yet fully understood how pretrained LLMs acquire such capabilities.
In this paper, we investigate the reason why a transformer-based language model can accomplish in-context learning after pre-training on a general language corpus by proposing a kernel-regression perspective of understanding LLMs' ICL bahaviors when faced with in-context examples. More concretely, we first prove that Bayesian inference on in-context prompts can be asymptotically understood as kernel regression $\hat y = \sum_i y_i K(x, x_i)/\sum_i K(x, x_i)$ as the number of in-context demonstrations grows. Then, we empirically investigate the in-context behaviors of language models. We find that during ICL, the attention and hidden features in LLMs match the behaviors of a kernel regression. Finally, our theory provides insights into multiple phenomena observed in the ICL field: why retrieving demonstrative samples similar to test samples can help, why ICL performance is sensitive to the output formats, and why ICL accuracy benefits from selecting in-distribution and representative samples. Code and resources are publicly available at \url{https://github.com/Glaciohound/Explain-ICL-As-Kernel-Regression}.
\end{abstract}

\section{Introduction}
\label{sec:introduction}

Pre-trained large language models (LLMs) have emerged as powerful tools in the field of natural language processing, demonstrating remarkable performance across a broad range of applications~\citep{weiemergent, kojimalarge, weichain, brown2020language,hierarchicalschema2023}. They have been used to tackle diverse tasks such as text summarization, sentiment analysis, schema induction and translation, among others~\citep{brown2020language, radford2019language,hierarchicalschema2023}. One of the most fascinating capabilities of LLMs is their ability to perform in-context learning (ICL), a process in which a language model can make predictions on a test sample based on a few demonstrative examples provided in the input context~\citep{logan2022cutting}.  This feature makes LLMs particularly versatile and adaptive to different tasks.
Studies have found ICL to emerge especially when the size of LLM is large enough and pre-trained over a massive corpus~\citep{wei2023larger}.

Although intuitive for human learners, ICL poses a mystery for optimization theories because of the significant format shift between ICL prompts and pre-training corpus. There have been lots of efforts to provide a theoretical understanding of how LLMs implement ICL. Some work~\citep{xieexplanation, wang2023large} approaches this problem from a data perspective: they claim that ICL is possible if a model masters Bayesian inference on pre-training distribution. Our paper also follows the setting in \citet{xieexplanation} that models the natural language as a mixture of sub-distributions and that LLMs conduct Bayesian inference on it while focusing on providing a tractable approximation of ICL inference via kernel regression along with insights into ICL behaviors.
Another stream of work conjectures that under a simple linear setting: $[x, y]$ where $y = w^\top x$ and the input sequence $x$ only has length 1, \hengarchived{what does linear setting mean? elaborate?} they can construct a Transformer~\citep{vaswani2017attention} to implement gradient descent (GD) algorithm over ICL prompt~\citep{akyrek2023what, von2022transformers, gargcan}. However, this constrained setting diverges from the most interesting part of ICL, as state-of-the-art LLMs work with linguistic tasks \hengarchived{are you indicating some linguistic tasks are not linear? what does that mean? do you mean tasks like parsing and structured prediction? also linear in input/output?} where the sequential textual inputs have complex semantic structures, and ICL emerges from pre-training on general-purpose corpus instead of explicit ICL training.

In this work, we delve deeper into the question of \textit{how to understand the mechanism that enables Transformer-based pre-trained LLMs to accomplish in-context learning on sequential data}. We specifically explore the hypothesis that LLMs employ a kernel regression algorithm when confronted with in-context prompts. Kernel regression adopts a non-parametric form 
\begin{equation}
\label{eq:kernel_regression}
    \hat{y}=\frac{\sum_i  y_iK(x, x_i)}{\sum_i K(x ,x_i)}
\end{equation}
when making predictions, where $K(x, x_i)$ is a kernel that measures the similarity between inputs $x$ and $x_i$. In plain words, it estimates the output $\hat{y}$ on $x$ by drawing information from similar other data points $x_i$ and taking a weighted sum on their $y_i$.


We first provide a theoretical analysis demonstrating that Bayesian inference predictions on in-context prompts converge to a kernel regression in Section~\ref{sec:theory}. In Section~\ref{subsec:insights}, our results also shed light on numerous phenomena observed in previous empirical studies, such as the advantage of retrieving in-context examples that are similar to the test sample, \hengarchived{what do you mean by similar here? similar between in-context examples and test examples, or similar among in-context examples? if the latter it's counter-intuitive since usually we prefer diverse examples. the most important thing is probably to make sure the examples represent the task well. how to measure that?} the sensitivity of ICL performance to the output formats, and why using a group of in-distribution and representative samples improves ICL accuracy.
\hengarchived{give examples on which tasks are not well suited for ICL? for example, in our recent work on fine-grained event detection, ICL does not work because this task requires many examples to represent the meaning of 3000+ event types: https://arxiv.org/abs/2303.09093}

Following our theoretical investigation, in Section~\ref{sec:empirical} we conduct empirical studies to verify our explanation of in-context learning of LLMs in more detail. Our results reveal that during LLM ICL, the attention map used by the last token to predict the next token \hengarchived{attention maps between words?} is allocated in accordance with our explanation. By plugging attention values into our equation, we are also able to reconstruct the model's output with over 80\% accuracy. Moreover, we are able to reveal how information necessary to kernel regression is computed in intermediate LLM layers.


    


\begin{figure*}[t!]
\centering
\includegraphics[width=\textwidth]{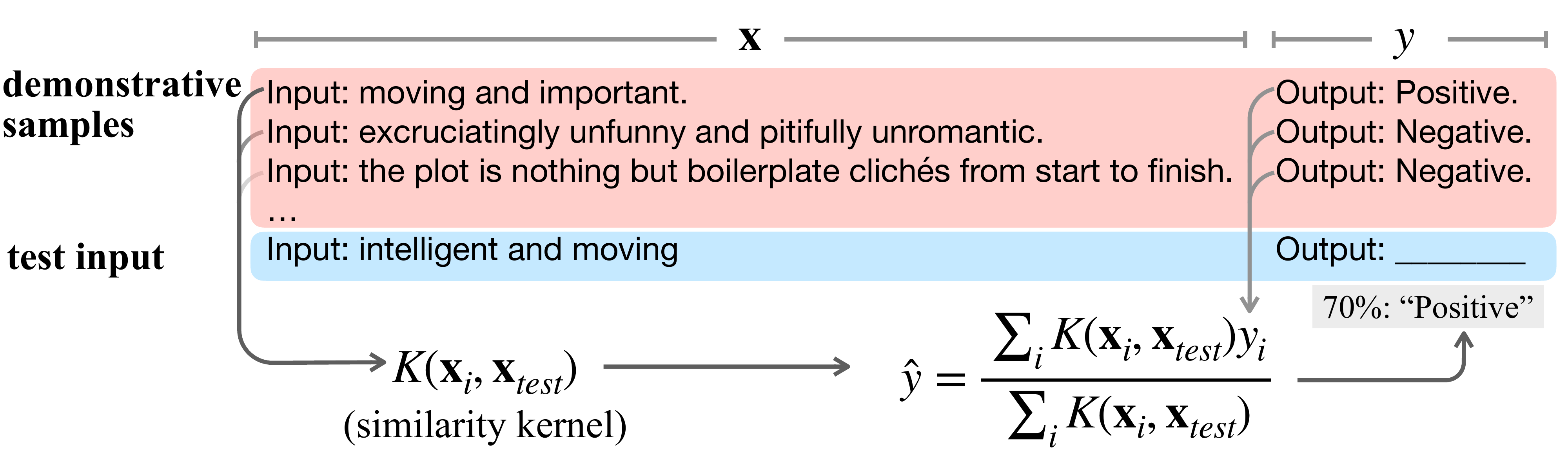}

\caption{Our results suggests that LLMs might be conducting kernel regression on ICL prompts.}
\label{fig:overview}
\end{figure*}

\section{Related Work}
\label{sec:related}

\subsection{In-Context Learning}

As an intriguing property of large language models, in-context learning has attracted high attention in the research community. There have been numerous studies on empirical analysis of in-context learning, including format and effects of in-context samples~\citep{min2022rethinking, min2022noisy, zhao2021calibrate}, selection of in-context samples~\citep{liu2022makes, lu2022fantastically}, characteristics of in-context learning behaviors~\citep{zhao2021calibrate, lu2022fantastically, khashabi2022reframing, wei2023larger}, and the relation between ICL and pre-training dataset~\citep{wu2022insights, chan2022data}.

Theoretical analysis of ICL is another active research direction. One branch of studies investigates ICL from a data perspective: under some general assumptions of the data, \citep{xie2021gx, wang2023large} demonstrate that with sufficient pre-training data and a good enough generative model of the data, ICL could be understood as performing Bayesian inference. However, they do not fully explain if such Bayesian inference is computationally feasible in practical language models since such Bayesian inference involves unbounded depth computational graphs as the number of samples increases. Our study builds on top of some similar assumptions but goes further to explain how ICL can be accomplished with the attention mechanism in Transformers~\cite{vaswani2017attention}. Recently, there is controversy regarding whether LLMs always conduct Bayesian inference in the ICL setting, and \citet{falck2024context, raventos2023pretraining, park2024competition} show that there exist scenarios where LLMs' predictions deviate from the predictions of Bayesian inference. This distinction can be quantified as an error term in Equation~\ref{eq:convergence} related to $\epsilon_\theta$ which we use to measure the deviation between ICL task and pre-training data. \cite{raventos2023pretraining, lu2024asymptotic} additionally show that under certain conditions, this divergence enables Transformers to achieve stronger performance than Bayesian inference.  Both \cite{raventos2023pretraining, park2024competition} discover that there is a threshold needed where ICL behaviors become less similar to Bayesian inference.

Another perspective to explain ICL is by analyzing what algorithms might be implemented in LLMs under ICL settings, which is closely related to our study. Representative studies include \citep{akyrek2023what, gargcan, von2022transformers, dai2023can, mahankalione, ren2023towards}, with the majority of them proposing gradient descent (GD) algorithm as a promising candidate answer. More recently,~\citet{fu2024transformers} show that transformers can learn to approximate second-order optimization methods for ICL, such as the Iterative Newton's method, which is exponentially faster than GD. These studies are mostly constructive, in the sense that there exist parameters of a transformer that can be used to approximate certain algorithms. On the other hand, theoretical explanations of whether or why such parameters might be learned during pre-training on language data are less explored. Furthermore, such settings usually focus on a specific type of question, such as linear regression, to construct the algorithm. 

Going beyond the linear setting, \cite{guotransformers} constructs solutions of ridge regression assuming non-linear representations on synthetic datasets.
On top of individual algorithms, \cite{bai2024transformers} explores the possibility of Transformers selecting from a pool of algorithms.
Some following-up work analyzed how training dynamics converge Transformer to ICL solutions on linear functions~\cite{zhang2024trained, huangcontext, ahn2023transformers}, and when linear-attention Transformers is trained on linear tasks~\cite{lu2024asymptotic}. \cite{kimtransformers} analyzed the training landscape of Transformers on ICL inputs, including the avoidance of saddle points and proximity to global optima.
These attempts in explicit constructions of algorithms in Transformers mostly assume a relatively simplified setting with input lengths equal to 1, and evaluate Transformers after training on a synthetic dataset (including ~\citep{akyrek2023what, gargcan}). This is different from ICL's main advantage as an emergent ability on language pre-training, and LLMs can work on textual data which involves sentences with lengths longer than 1.

\subsection{Emergent Ability of LLMs}

This is a larger topic that in-context learning is also highly related to. \citep{weiemergent, brown2020language, kojimalarge, weichain} showed that abilities including reasoning, in-context learning, few-shot learning, instruction understanding, and multilingualism emerge in large language models after pre-training on massive language data. These impressive and mysterious capacities have boosted significant progress in natural language processing as well as artificial intelligence, but still baffle theoretical analysis. In this work, we make a preliminary step towards understanding ICL as a special case of LM capacity emergence.

\subsection{Associating Attention with Kernel Regression}
\cite{chen2023calibrating} presents a related view of associating self-attention with kernel ridge regressions. They focus on a different problem and propose to substitute attention mechanisms with a kernel to tackle the uncertainty calibration problem. Another blog, \cite{olsson2022context}, also relates attention to kernel regression but defines ICL as  ``decreasing loss at increasing token indices'', rather than the more widely adopted definition of learning from in-context demonstrations as in ours. Another related stream of work is the topic of associative memory in deep learning networks. ~\cite{ramsauerhopfield} revisits Hopfield networks which is capable of reusing past patterns for inference on new data. \cite{arorazoology, bietti2023birth} explore components in Transformers which might be responsible for associative recalling.

\section{Formulation}
\label{sec:formulation}

\subsection{Preliminaries: Hidden Markov Models}

Following the setting of \citep{xieexplanation}, we assume that the pre-training corpus can be modeled by a mixture of HMMs. Each HMM corresponds to a certain task $\theta\in\Theta$.
Assuming a large finite number of tasks, one can include all task-specific HMMs into one single HMM. In this unified HMM,
let $\mcS$ be the set of states, and $\mcO$ be the set of observations where $|\mcO| = m$.
At each time step, state $s_t$ randomly emits one observation $o_t$ and then transits to the next state $s_{t+1}$.
Quantities $p_\text{pre-train}$, $P(s_{t+1}=s'|s_t=s)$ and $P(o_t=o|s_t=s)$ denote the pre-training initial distribution, transition distribution, and emission distribution respectively. Under an arbitrary ordering of $\mathcal{S}$ and $\mathcal{O}$, we can define the transition matrix $T: T(s, s')=P(s' | s)$, and emission matrix $B: B(s, o)=P(o | s)$, respectively.
We also let $\bfo=(o_0,\cdots)$ be the full observation sequence, and $\bfo_{[0:l]}$ denote its first $l$ tokens.


\subsection{In-Context Learning}
\label{subsec:icl_formulation}

In this work, we consider the following formulation of in-context learning (ICL). Let $\Theta$ be the set of tasks.
We follow the ICL prompt formulation, the HMM mixture formulation, and assumptions in \citep{xieexplanation} as follows.
The distribution of sequences generated by each individual task in the HMM together composes the pre-training distribution. Specifically, each task $\theta\in\Theta$ is associated with a distinct initial state $s_\theta\in\mathcal{S}$ with transition rate lower bound $\epsilon_d$ (the Assumption 5 ``Regularity'' in \cite{xieexplanation}), and the set of all such initial states $\mcS_\text{start}=\{s_\theta| \theta \in \Theta\}$ forms the support of $\pretrain$.
For a test task $\theta^\star$, the in-context learning prompt follows the format:

\begin{equation}
\label{eq:prompt}
    \bfo_{ICL} = [S_n, \xtest]=[\bfx_1,y_1,o^\text{delim}, \bfx_2, y_2, o^\text{delim}, \cdots,  \bfx_n, y_n, o^\text{delim}, \xtest],
\end{equation}
where the input-output pairs $[\bfx_i, y_i]$ are i.i.d. demonstrate samples sampled from $\theta^\star$, and $\delim$ is the delimiter token with emission rate lower bound $\epsilon_r$ (the Assumption 5 ``Regularity'' in \cite{xieexplanation}).
We further make some connections between in-context learning and the HMM model. Note that the probability of generating a sequence from the initial distribution $p_0$ can be expressed as follows\citep{jaeger2000observable}:
\begin{equation}
\label{eq:inference}
    P(\bfo_{[0:l]} | \pretrain) = \pretrain^\top \left(\prod_{i=0}^{l-1} \text{diag}(\mathbf{p}_{o_i}) T \right) \text{diag}(\mathbf{p}_{o_l}) \mathbf{1},
\end{equation}
where $\mathbf{p}_{o}$ is vector of emission probabilities $P(o|s\in\mathcal{S})$ for $o$.
We denote the intermediate matrices in Equation~\ref{eq:inference} as one operator for a (sub)sequence $\bfo$: 
\begin{equation}
\label{eq:T_x}    
T_{\bfo}=\prod_{i=0}^{l-1} \text{diag}(\mathbf{p}_{o_i}) T.
\end{equation}
We use a matrix $\Sigma_{p,l}$ to denote the covariance between all of its $d^2$ elements of $\mathrm{vec}(T_{\bfo_{[0:l-1]}})$ when $\bfo_{[0:l-1]}$ is generated from initial distribution $p$. For each individual task, we denote $\epsilon_\theta = \inf_l \rho(\Sigma_{\pretrain, l}^{-1} - \Sigma_{s_\theta,l}^{-1})$ to quantify the difference between sequences generated by $s_\theta$ and those from pre-training distribution, where $\rho$ denotes the spectral radius of a matrix.
Let $\eta=\sup_{\bfo_{[0:l-1]}} \|T_{\bfo_{[0:l]}}\|_F$ be the upper bound of $T_{\bfo_{[0:l]}}$'s Frobenius-norm.

\subsection{Assumptions}
\label{subsec:assumptions}

We go on and present the assumptions we borrow from \cite{xieexplanation}.


\begin{assumption}
\label{assumption:anchor}
    (Delimiter Tokens) The delimiter token indicates the start of sampling of new sequences:
    \[
        P(s_{t+1} = s \mid o^\text{delim}) > 0 \rightarrow s \in \mcS_\text{start}
    \]
\end{assumption}
\textbf{Remark}: this means that the delimiter tokens  are indicative enough, such as the start of a new line before the beginning of a paragraph.


\begin{assumption}
\label{assumption:KL}
    (Task Distinguishability) The Kullback–Leibler divergence (KL divergence) between the first $l$ tokens between two distinct tasks $\theta\ne \theta'$ is lower-bounded by:
    \[
        \inf_{\theta,\theta'} D_{KL}\left(P(\bfo_{[0:l]}|\theta') ~\|~P(\bfo_{[0:l]}|\theta)\right) = \epsilon_{KL} > \ln\frac{1}{\epsilon_r\epsilon_d}.
    \]
\end{assumption}
\textbf{Remark}: this requires that tasks are distinguishable from each other. As KL-divergence is non-decreasing with length $l$, it suffices to increase the length $l$ to provide sufficient task information.


\section{Theoretical Analysis}
\label{sec:theory}

\subsection{Explaining ICL as Kernel Regression}
\label{subsec:explanation}

Within the framework presented in Section~\ref{sec:formulation}, we pose the following result. The basic idea is that, as the number of samples $n$ increases, inference on the in-context learning prompt converges to a kernel-regression form.

\begin{theorem}
\label{thm:kernel}
    Let us denote a kernel 
    \begin{equation}    
    \label{eq:kernel}
    \mathcal{K}(\bfx, \bfx') = \mathrm{vec}(T_{\bfx})^\top \Sigma_{\pretrain, }^{-1} \mathrm{vec}(T_{\bfx'}).
    \end{equation}
    $T_{\bfx}$ is defined in Equation~\ref{eq:T_x}.
    Let $\mathbf{e}(y)$ be the one-hot vector for index $y$. Then the difference between the following logit vector in the form of kernel regression:
    \begin{equation}
    \label{eq:construction}
        \hat{\mathbf{y}} = \frac{\sum_{i=1}^n  \mathbf{e}(y_i) \mathcal{K}(\xtest, \bfx_i)}{\sum_{i=1}^n \mathcal{K}(\xtest, \bfx_i)}
    \end{equation}
    and it converges polynomially to the conditional likelihood of $Y$ conditioned on ICL prompt $\bfo_{ICL}$, $P(Y \mid \bfo_{ICL}, \pretrain)$, with probability $1-\delta$:
    \begin{equation}
    \label{eq:convergence}
        \|\hat{\bfy}
        -
        P(Y\mid \bfo_{ICL}, \pretrain)\|_\infty
        =  \eta^2\epsilon_\theta + O\left(\sqrt{\frac{1}{n} \ln\frac{4m}{\delta}}\right)
    \end{equation}
    
\end{theorem}

Equation~\ref{eq:construction} can be interpreted as follows: it calculates the semantic similarity between the test input $\xtest$ and each sample $\bfx_i$ and aggregates their outputs to compute a most likely prediction for the test sample. This is natural to the motivation of ICL: we encourage the LLM to leverage the pattern provided in demonstrative samples and mimic the pattern to predict the test input.
Equation~\ref{eq:construction} is also similar to the form of attention mechanism used in Transformer decoder models:
\begin{equation}
\label{eq:attention}
    h 
    = \text{softmax}(q^\top K) V^\top
    = \frac{\sum_{i} v_i e^{\langle q, k_i\rangle}}{\sum_{i} e^{\langle q, k_i\rangle}}
\end{equation}
where $q$ is the query vector corresponding to the last token, $k, K$ are the key vectors and matrix, and $v, V$ are the value vectors and matrix used in the Transformer, respectively. The only difference is that $e^{<q, k_i>}$ is replaced with a dot product in Equation~\ref{eq:construction}, which can be regarded as a kernel trick. We assume that previously hidden layers are responsible for learning the semantic vectors of sample inputs $\mathrm{vec}(T_\bfx)$. We can then make the following loose analogy between our kernel regression explanation (Equation~\ref{eq:construction}) and the attention mechanism (Equation~\ref{eq:attention}):
\begin{itemize}
    \item Label information $\mathbf{e}(y_i)$ corresponds with the \textit{value} vector $v_i$
    \item The similarity kernel $\mathrm{vec}(T_{\bfx})^\top \Sigma_{\pretrain, }^{-1} \mathrm{vec}(T_{\bfx'})$ loosely corresponds to the attention value $e^{\langle q, k_i\rangle}$, where:
    \item the semantic information $\mathrm{vec}(T_{\bfx_i})$ corresponds to the \textit{key} vectors $k_i$ and \textit{query} vectors $q_i$ for samples $[\bfx, y]$.
\end{itemize}

Here we further explain how the kernel is effected by the LLM. Equation~\ref{eq:construction} measures similarity between $\mathbf{x}$ and $\mathbf{x}'$ on the space of $\text{vec}(T_{\mathbf{x}})$ and $\text{vec}(T_{\mathbf{x}'})$. This flattened vector of $T_{\mathbf{x}}$ defines the ``belief state'' in HMMs, which determines the conditional probability $P(\cdot | \mathbf{x})$ after prefix $\mathbf{x}$. This is the pre-training objective functionality of LLMs. Therefore, if $x_i$ and $x_{test}$ have similar follow-up conditional probabilities under the LLM, their similarity value will be larger, and vice versa.

One might argue that it is also theoretically possible to directly compute the next token likelihood in Equation~\ref{eq:inference}.
However, this form involves $2n$ consecutive matrix multiplications. When $n$ increases, this is infeasible for a practical Transformer architecture which is composed of a fixed number of layers. In comparison, Equation~\ref{eq:construction} only requires semantic information for each sample $\bfx$ to be provided beforehand and then applies kernel regression (which can be done by one attention layer) to get the answer. Learning to represent $T_{\bfx}$ is probable for preceding layers, as it is also used for ordinary inference $P(y\mid \bfx) = \pretrain^\top T_{\bfx}$. In experiments in Section~\ref{sec:empirical} we demonstrate that this analogy can explain the ICL behaviors of LLMs to an extent.

\vspace{-2mm}
\begin{figure*}[t!]
\centering
\includegraphics[width=\textwidth]{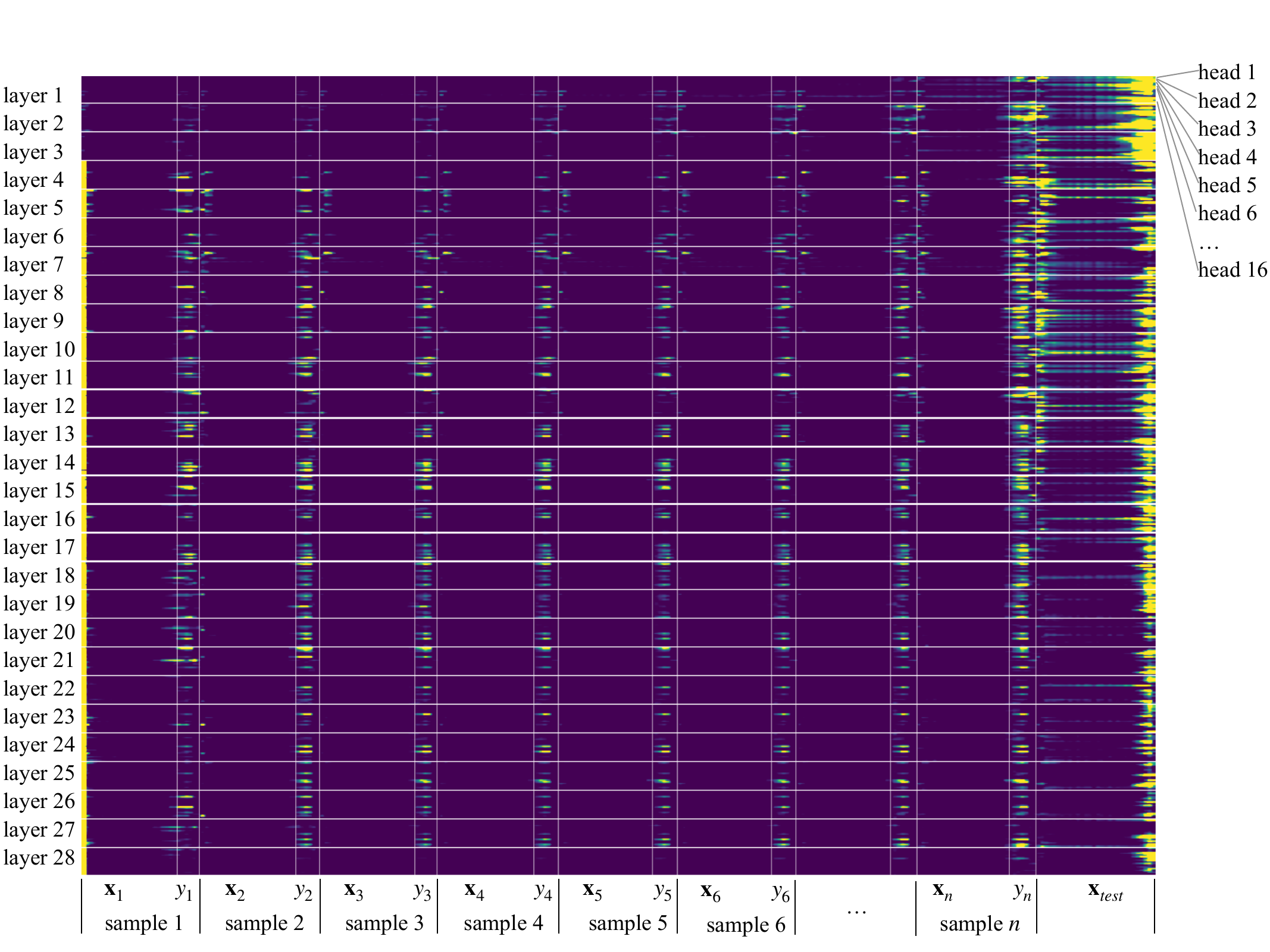}

\vspace{-3mm}
\caption{Averaged attention map over GLUE-sst2 test set. A portion of attention on demonstrative samples is generally focused on label positions $y_i$. This conforms to the intuition in Theorem~\ref{thm:kernel} that the inference on in-context learning prompts is a weighted average over sample labels.}
\vspace{-4mm}
\label{fig:overall_attention}
\end{figure*}

\subsection{Insights Provided by the Explanation}
\label{subsec:insights}



Theorem~\ref{thm:kernel} can provide insights into multiple phenomena in the ICL field observed by previous studies. This is helpful for understanding and predicting the behaviors of ICL, and providing heuristics for future development.

\textbf{Retrieving Similar Samples}
It is empirically observed~\citep{rubin2022learning, liu2022makes} that retrieving demonstrative samples $\bfx_i$ that is similar to the test input $\bfx_\text{test}$ can benefit ICL performance. This phenomenon is understandable from our explanation. Encouraging the selection of similar samples can be understood as limiting the cosine distance between demonstrative samples $\bfx_i$ and test sample $\bfx_\text{test}$ in sentence embedding space. This is similar to selecting a smaller ``bandwidth'' in kernel regression and sampling only from a local window, which reduces the bias in kernel regression. Therefore, devising better retrieval techniques for selecting samples, especially those with similar representations as LLMs, is a promising direction for further boosting ICL scores.

\textbf{Sensitivity to Label Format}
~\citep{min2022rethinking} also observes that the ICL performance relies on the label format. Replacing the label set with another random set will reduce the performance of ordinary auto-regressive LLMs.
This can be explained in our theory that the model's output comes from a weighted voting of demonstrative sample labels $\{y_i\}$. If the label space is changed, the next token will also be misled to a different output space. So it is generally beneficial for ICL to ensure the demonstrative samples and test samples share an aligned label space and output format.

\begin{wraptable}{l}{8cm}
\centering
\caption{Effect of OOD inputs on ICL on SST2 dataset.}
\begin{tabular}{cccccc}
\toprule
OOD-type & None & rare word & complex & typo \\
\midrule
accuracy & 0.805 & 0.677 & 0.788 & 0.534 \\
\bottomrule
\end{tabular}
\end{wraptable} 

\paragraph{Sample Selection}
In Equation~\ref{eq:convergence}, the existence of the $\eta^2\epsilon_\theta$ term implies that the demonstrations $[\bfx_i, y_i]$ should be sampled in a way close to $\pretrain$. To verify this, we conduct an experiment by converting test inputs to semantically similar but rarer sentences. Specifically, on the SST2 dataset, we prompt GPT-3.5-turbo to generate semantically similar while rarer (i.e. OOD) expressions of inputs and use them to substitute the original dataset inputs. We consider three types of OOD types. ``Rare word`` is where words are substituted with rare synonyms. ``Complex`` is to express the original sentence in a more complex structure. ``Typo`` is where we require adding typos to the original input. The results are listed below. We see that compared with original scores (``None`` ), these OOD types more or less decrease the ICL accuracy, with ``typo`` having the largest effect, dropping the accuracy to near random.


\paragraph{Bias from Pre-training Data} 
Theorem~\ref{thm:kernel} implies that the final prediction depends both on in-context examples and prior knowledge from pre-training. This can be seen from the $\eta^2 \epsilon_\theta$ term in Equation~\ref{eq:convergence}, which comes from the pre-training information in Equation~\ref{eq:prior_info_term} in Appendix~\ref{appsec:proof}. Specifically, $\epsilon_\theta$ measures the similarity between $\theta$'s distribution and pre-training distribution, and $\eta^2 \epsilon_\theta$ describes how the pre-training distribution's bias affects the ICL prediction. When the task $\theta$ emits a sequence distribution similar to the pretraining distribution, $\epsilon_\theta$ will be small, and the ICL prediction $\hat{y}$ will converge to $P(Y| \bfo_{ICL}, p_{\pretrain})$ with a smaller error $\eta^2 \epsilon_\theta$, and vise versa.  This partially elucidates \cite{kossen2023context}'s finding that ``ICL cannot overcome prediction preferences from pre-training.'' 

\paragraph{Remaining Challenges}

However, we need to point out that there are still phenomena not explainable by our framework, as well as most previous explanations. One most mysterious one is the sensitivity to sample ordering~\citep{lu2022fantastically} as a kernel regression should be order-ignorant, which no existing explanations (including ~\citep{xieexplanation, akyrek2023what}) take into account. Another intriguing question is that some work finds LLMs robust to perturbed or random labels\citep{kossen2023context} while others find the opposite \citep{min2022rethinking}. We attribute such phenomena to the fact that LLMs also rely on a large portion of implicit reasoning in text generation and might benefit from linguistic cues in text prompts. Our theory provides a partial explanation, which needs to be combined with this implicit ability of LLMs to form a more comprehensive understanding of ICL.

\section{Empirical Analysis}
\label{sec:empirical}
In this section, we conduct empirical analysis on LLMs in order to verify our hypothesis. Because Equation~\ref{eq:construction} is only one special solution among infinitely many of its equivalent forms, and it also relies on the unknown HMM structure, it is infeasible to directly evaluate it on data. However, we can verify if it can predict observable behaviors on LLMs in experiments. Limited by computation resources in an academic lab, we analyze the GPT-J 6B model\citep{gpt-j} on one Tesla V100. It employs a decoder-only Transformer architecture. In this section, we use the validation set of the sst2 dataset as a case study, while results on more tasks can be found in Appendix~\ref{appsec:more_results}. 
We investigate the ICL behavior of LLMs from shallow to deep levels, and sequentially ask the following 4 questions: Do the attention heads collect label information $\mathbf{e}(y_i)$ as predicted? Does the attention-kernel analogy explain LLM's prediction? Can we actually explain the attention values as a kind of similarity? Can we find where algorithmic features $\mathbf{e}(y_i), T_{\bfx_i}$ are stored? The following sections answer these questions one by one.

\begin{figure}
     \centering
     \begin{subfigure}[b]{0.45\textwidth}
         \centering
         \includegraphics[width=\textwidth]{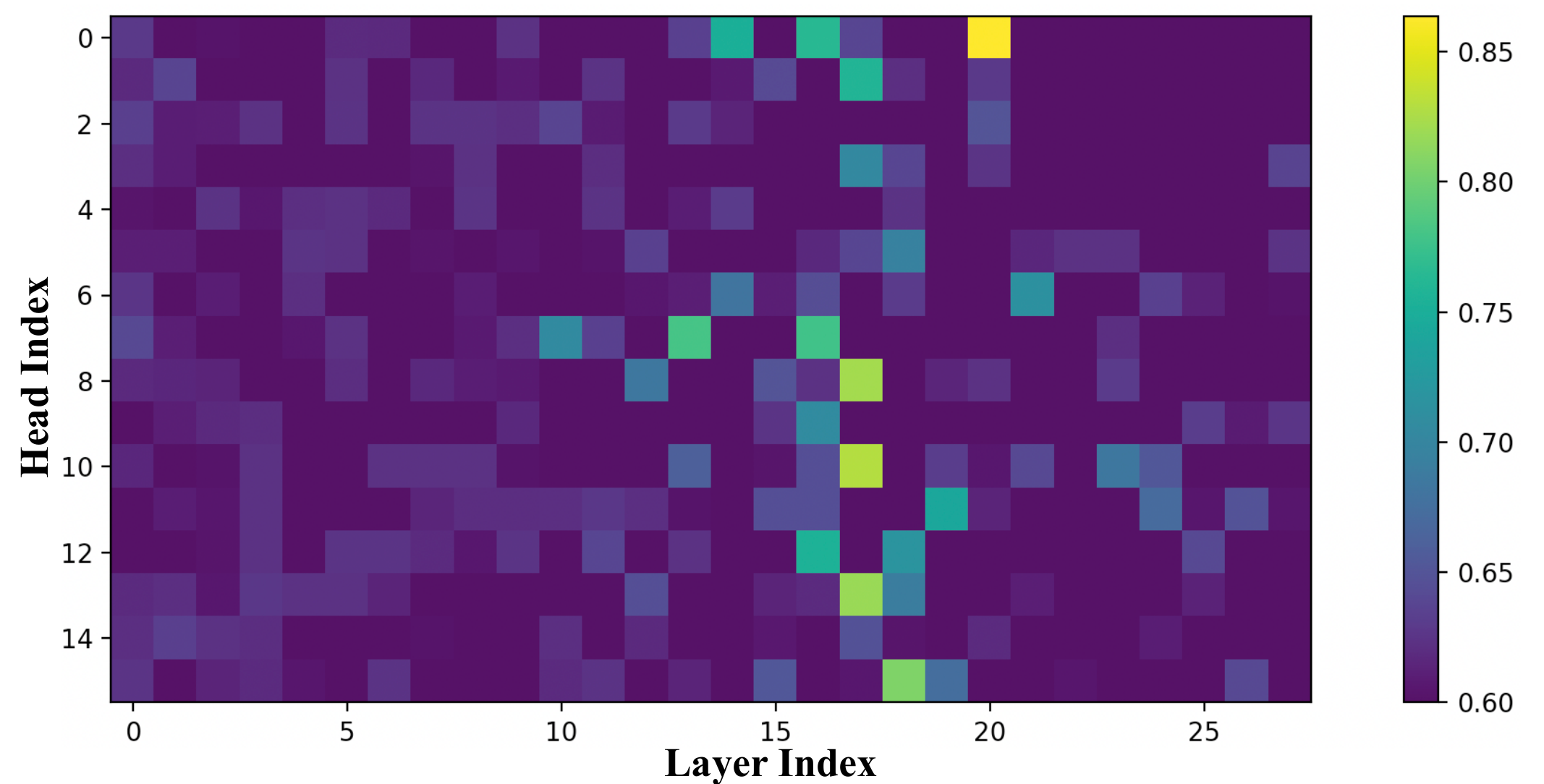}
         \vspace{-2mm}
         \caption{Accuracy compared with model output $\hat y$.}
         \label{fig:attention_reconstruction_accuracy}
     \end{subfigure}
     \hfill
     \begin{subfigure}[b]{0.45\textwidth}
         \centering
         \includegraphics[width=\textwidth]{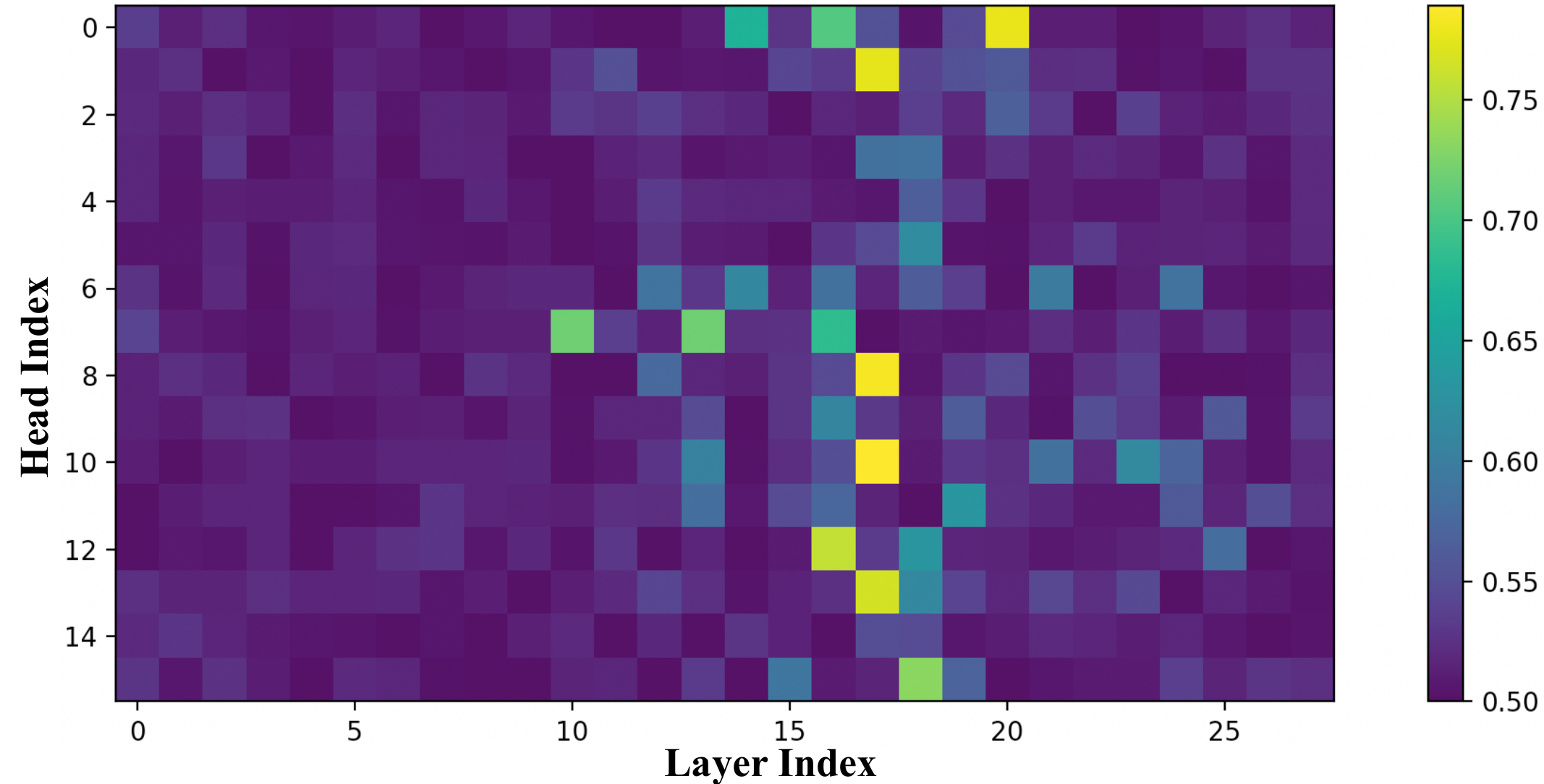}
         \vspace{-2mm}
         \caption{Accuracy compared with ground truth label $y_\text{test}$.}
         \label{fig:attention_reconstruction_label_accuracy}
     \end{subfigure}
\caption{We use each head's attention weights on demonstrative samples to manually average sample labels $y_i$.
These figures show that ``reconstructed'' outputs in some heads from layers 16$\sim$21 matches LLM prediction with as high as 89.2\% accuracy, and matches ground truth with 86.4\% accuracy.}
\label{fig:attention_reconstruction}
\end{figure}

\subsection{Where Are Attentions Distributed During ICL?}
\label{subsec:attention_distribution}

First, we notice that Equation~\ref{eq:prompt} implies that the LLM takes a weighted average over sample labels $y_i$ in ICL.
Figure ~\ref{fig:overall_attention} shows how attention weights are distributed on in-context learning inputs $[S_n, \xtest]$. On each test point, we sample one ICL prompt and collect the attention map over previous tokens for predicting the next token. After getting the attention maps, as ICL samples $\bfx_i$ may have varied lengths, we re-scale the attentions on each $\bfx$ from $|\bfx|$ to a fixed 30-token length with linear interpolation. After aligning the attention lengths, we average all attention maps. The horizontal axis is the aligned positions on prompt $[S_n, \xtest]$. Each bar corresponds to one of 28 Transformer layers. Within each bar, each thin line is 1 out of 16 attention heads. Darker (blue) areas mean smaller averaged attention, while brighter areas indicate high attention.


In Figure~\ref{fig:overall_attention}, there are three major locations of attention masses. First, a majority of attention is focused on the final few tokens in $\xtest$, especially in the first 3 layers. This accords with previous observations that Transformer attentions tend to locate in a local window to construct \hengarchived{this word is misspelled, do you mean reconstruction?} local semantic feature for $\xtest$. Secondly, as also observed in previous studies, LLMs tend to allocate much attention on the first few tokens in a sequence to collect starter information. Finally and most intriguingly, we observe concentrated attention on each sample label tokens $\{y_i\}$. This phenomenon confirms an aggregation of label information in LLM ICL, in line with the prediction by Equation~\ref{eq:construction}. Note that our explanation does not specify or limit the model from implementing kernel regression in a particular layer. In fact, an equivalent mechanism can occur in one or more layers as long as these (possibly redundant) results can be passed with skip connections to the final layer and aggregated.

\subsection{Can Attentions Be Interpreted as Kernel Functions?}
\label{subsec:prediction_reconstruction}

Now that we observe expected locations of attention weights on labels, we go on to verify if the LLM really predicts by averaging on labels as suggested by Theorem~\ref{thm:kernel}. We iterate over 16 heads and 28 layers, and insert their attention weights into Equation~\ref{eq:construction} to manually average the label distribution. This is similar in concept to a mind-reading experiment to predict one's next word using brain waves only~\citep{eeg2022}. Specifically, for each attention head, we use the maximal attention value $a_i$ within the range of $[\bfx_i, y_i]$ as the kernel weight. Then on the ICL samples, we reconstruct the prediction as follows:
\[
    \tilde y = \arg\max \frac{\sum_{i=1}^n \mathbf{e}(y_i) a_i}{\sum_{i=1}^n a_i}
\]
The resulting ``reconstructed output'' $\Tilde{y}$ is compared for both LLM's actual prediction $\hat{y}$ and ground truth label $y_\text{test}$ to calculate its accuracy.
Figure~\ref{fig:attention_reconstruction_accuracy} and \ref{fig:attention_reconstruction_label_accuracy} plot the accuracy between $\Tilde{y}$ and $\hat{y}$ and between $\Tilde{y}$ and $y_\text{test}$ respectively.
Interestingly, we spot the existence of several heads in layers 18$\sim$21 which demonstrate high accuracy in reconstruction. The highest of them (layer 17, head 10) achieves 89.2\% accuracy on $\hat{y}$ and 86.4\% accuracy on $y_\text{test}$. This result validates our hypothesis that some components in Transformer-based LLMs implement kernel regression. Note that this phenomenon happens within a few adjacent layers in the middle of the model. This is similar to our prediction in Section~\ref{subsec:explanation}: not many attention layers are needed for kernel regression, as long as the required features have been computed by preceding layers. It is enough for the higher layers to only pass on the computed results.

\begin{wrapfigure}{l}{0.6\textwidth}
\centering
\includegraphics[width=0.6\textwidth]{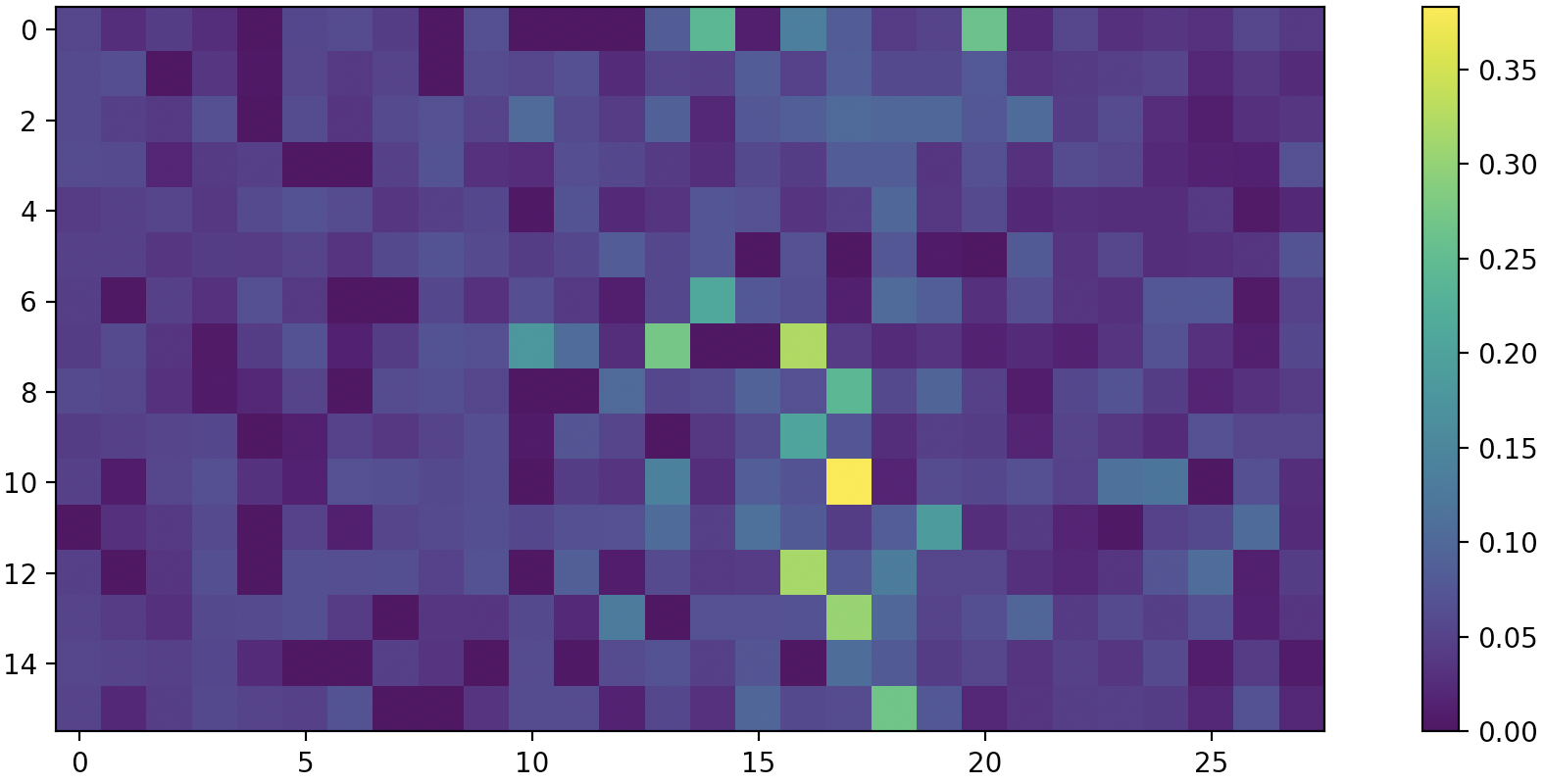}

\vspace{-1mm}
\caption{Pearson correlation between sample's attentions and \textit{prediction similarity} $\text{sim}_\text{pred}(\bfx_\text{test}, \bfx_i)$ (Equation~\ref{eq:prediction_similarity}). $x$-axis are layers and $y$-axis are heads in each layer. Note the resemblence between this heatmap and Figure~\ref{fig:attention_reconstruction}.}
\vspace{-2mm}
\label{fig:attention_logit_correlation}
\end{wrapfigure}

\subsection{Which Samples Receive High Attention?}
\label{subsec:interpret_attention}

We go on and ask the question: if the LLMs use attention to implement kernel regression, \textit{what kind of similarity does this kernel function evaluate?} From Equation~\ref{eq:construction}, we see that the dot product is measuring similarity between $T_{\bfx}$, which encodes information necessary for HMM inference: $p(o|\bfx) = \pretrain^\top T_\bfx B$. Therefore, we conjecture that the attention value $a_i$ between $\bfx_\text{test}$ and $\bfx_i$ correlates with their prediction similarity. Specifically, we define the \textit{prediction similarity} as follows:
\begin{equation}
\label{eq:prediction_similarity}
    \text{sim}(\bfx_1, \bfx_2) = P(o | \bfx_1)^\top P(o | \bfx_2),
\end{equation}
which is measured by applying LLMs on these texts \textit{alone}, rather than in ICL prompt. Finally, we compute the Pearson correlation coefficient between $\text{sim}(\bfx_\text{tes}, \bfx_i)$ and each attention values on samples for each attention head. The results are shown in Figure~\ref{fig:attention_logit_correlation}. The absolute value of correlation is not high, as $P(o | \bfx)$ is a dimension reduction to $T_\bfx$ and can lose and mix information. However, we can still note a striking similarity between it and Figure~\ref{fig:attention_reconstruction}. This means that the heads responsible for ICL mostly attend to \textit{prediction-similar} samples.

\subsection{Do Intermediate Features Store Information Useful for Kernel Regression?}
\label{subsec:interpret_features}

\begin{figure}
     \centering
     \begin{subfigure}[b]{0.45\textwidth}
         \centering
         \includegraphics[width=\textwidth]{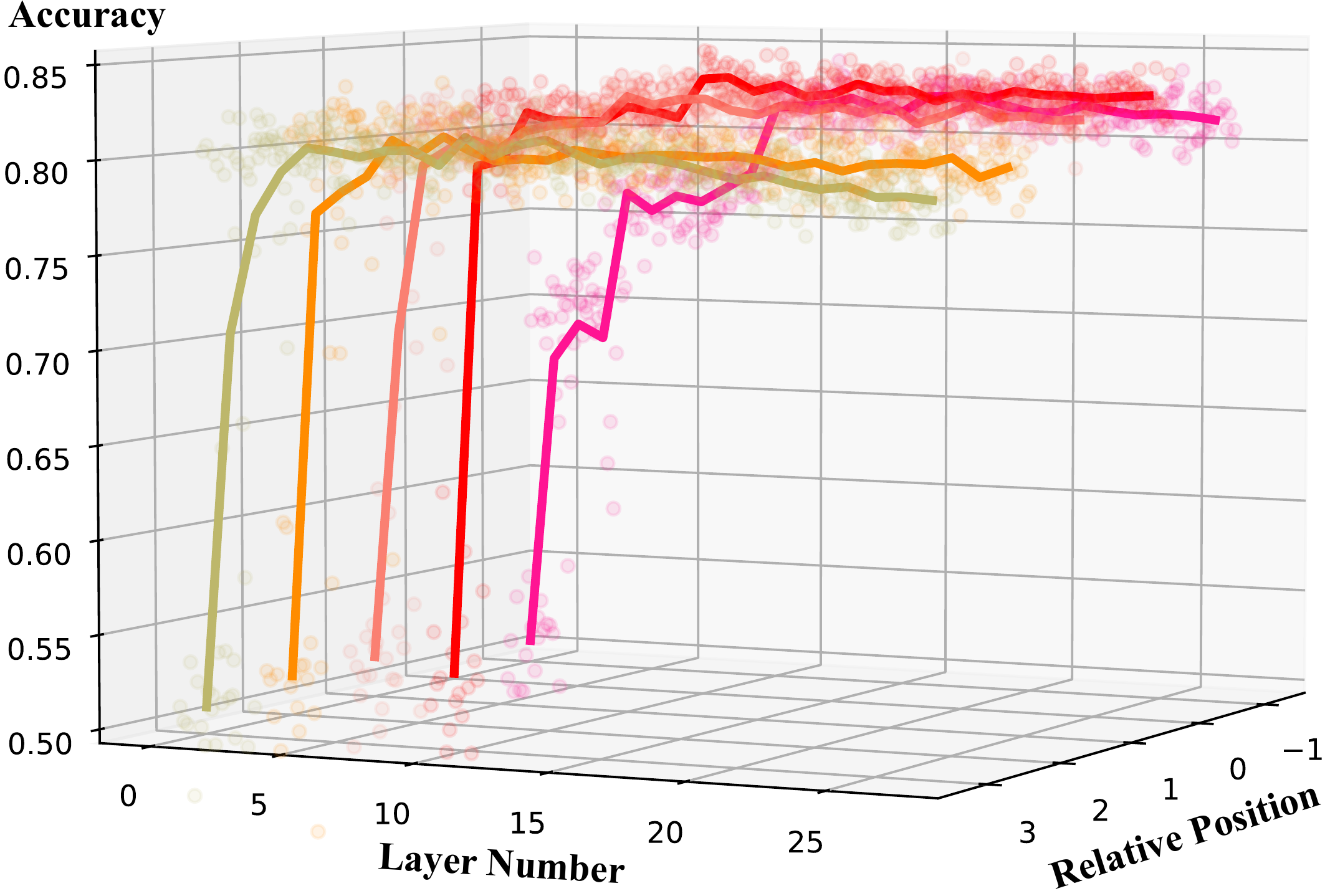}
         \caption{Predicting $\arg\max_o P(o|\bfx_i)$ with \textit{key} vectors.}
         \label{fig:key_recon_acc}
     \end{subfigure}
     \hfill
     \begin{subfigure}[b]{0.45\textwidth}
         \centering
         \includegraphics[width=\textwidth]{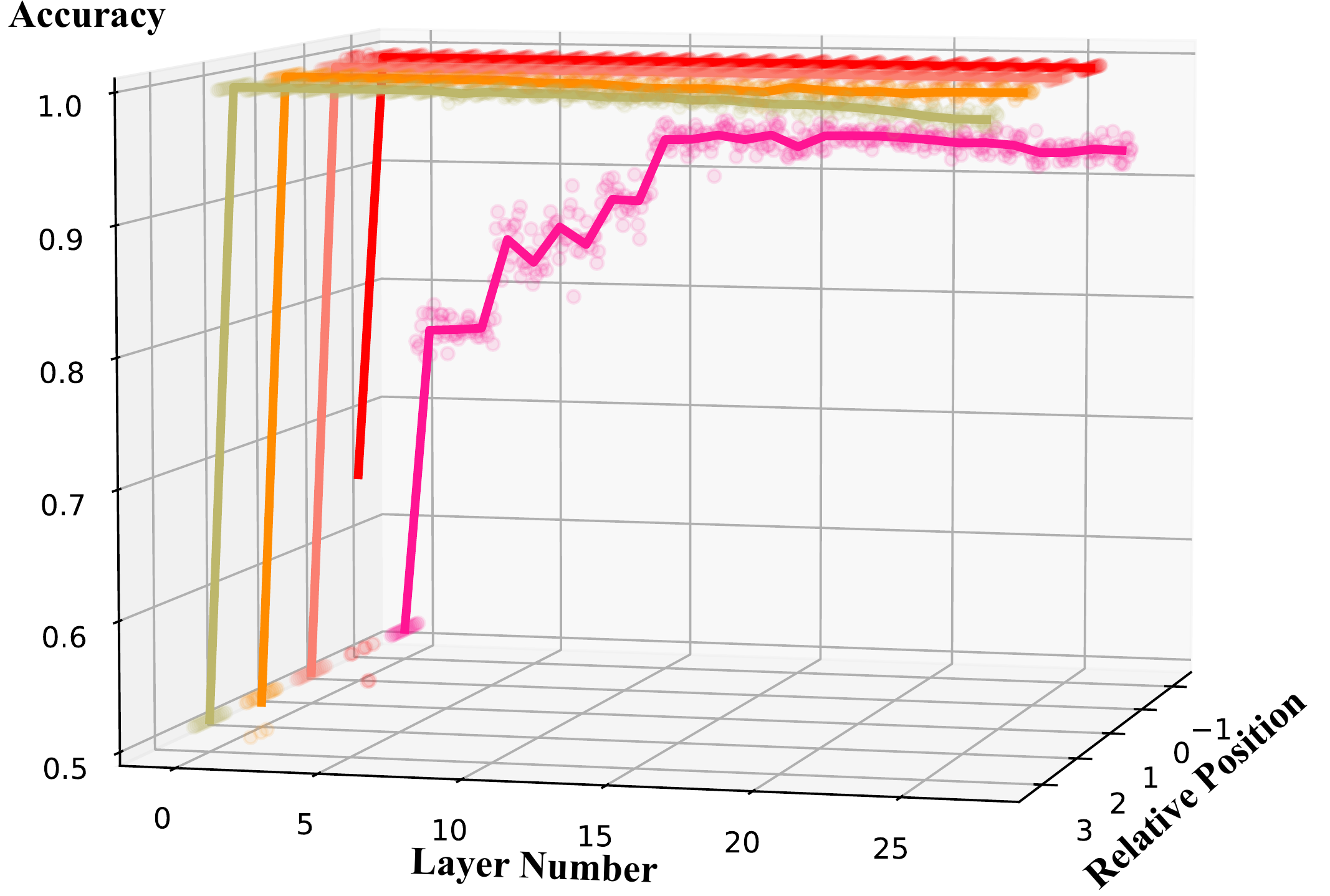}
         \caption{Predicting $y_i$ with \textit{value} vectors.}
         \label{fig:value_label_acc}
     \end{subfigure}
\caption{Key and value vectors encode label and LLM prediction information at high-attention position.
Here y-axis denotes the relative position to the high-attention position in each sample.
Each sphere is an attention head. The curve shows average accuracy within each layer.}
\label{fig:element_interpretation}
\end{figure}

In this section, we go into a more detailed level, and investigate the question: \textit{where do Transformer-based LMs store the algorithmic information needed by kernel regression?} To this end, we take out the intermediate \textit{key} and \textit{value} features in all layer heads and see if the correct information is stored in the correct locations. Note in Section~\ref{subsec:attention_distribution}, we observe that a major part of attention weights are located at the label position, so we focus on positions within $[-1, 3]$ relative to this position. Noticing the analogy we made at Section~\ref{subsec:explanation} that $k_i \sim \mathrm{vec}(T_{\bfx_j})$ and $v_j \sim y_j$, we study two sub-questions: (1) whether \textit{value} vectors encode label information $y_i$; and (2) whether \textit{key} vectors encode LLM prediction information $P(o|\bfx_i)$. For each head, we conduct Ridge regression with $\lambda=0.01$ to fit the task in these 2 questions. Results are presented in Figure~\ref{fig:element_interpretation}. We can observe that, generally the high-attention position (y-axis = 0) indeed achieves the best accuracy. Figure~\ref{fig:value_label_acc} is intuitive, as tokens at a position later than the label token $y_i$ can easily access the information of $y_i$ by self-attention. The slight drop at position +3 means that a longer distance introduces more noise to this information flow. Results in Figure~\ref{fig:key_recon_acc} tell us that, although sentence $\bfx_i$'s starting position in ICL prompt is shifted and different from 0, $k_i$ is still strongly correlated with $P(o | \bfx_i)$, which indicates a sense of translation invariance. Overall, the results mean that, with the attention map distributed in Figure~\ref{fig:overall_attention}, LLM is able to use the attention mechanism to extract information regarding $T_{\bfx_i}$ and $y_i$ from \textit{key} and \textit{value} vectors effectively just as we described.

\subsection{How Well Can Our Explanation Reconstruct ICL Prediction?}
\label{subsec:reconstruction}

Finally, we numerically evaluate our explanation on its ability to reconstruct the ICL predictions and tasks. We uniformly use 700 data in each validation set to represent tasks in a balanced way. We select the attention head that has the highest correlation with model predictions. This head is then evaluated on ICL prediction reconstruction and task performance on a held-out set of 300 data per task. In Table~\ref{tab:reconstruction_task} we see that the ICL output reconstruction has an accuracy from 68\% to 80\% except for the harder task of MNLI. The task performance of reconstructed outputs matches the model’s performance level. It also achieves similar or superior performance than kernel regression on sentence encoders such as all-MiniLM-L6-v2 and bert-base-nli-mean-tokens~\citep{reimers-2019-sentence-bert}.
\begin{table*}[t!]
\centering
\small
\linespread{1}

\caption{Performance of explicit kernel regression (KR) and LLM ICL on downstream tasks.}
\vspace{-2mm}
\setlength{\tabcolsep}{2mm}{
\begin{tabular}{lcccccc}
\toprule

\multirow{2}{*}{\textbf{Method}} & \multirow{2}{*}{\textbf{sst2}} & \multirow{2}{*}{\textbf{mnli}} & \textbf{rotten-} & \textbf{tweet\_eval} & \textbf{tweet\_eval} & \textbf{tweet\_eval} \\

& & & \textbf{tomatoes} & \textbf{(hate)} & \textbf{(irony)} & \textbf{(offensive)} \\

\midrule

\textbf{GPT-J-6B ICL} & 0.805 & 0.383 & 0.671 & 0.539 & 0.519 & 0.542 \\

\midrule

\textbf{ICL output reconstruction} & 0.797 & 0.597 & 0.683 & 0.713 & 0.722 & 0.753 \\
\textbf{Reconstructed performance} & 0.750 & 0.360 & 0.527 & 0.587 & 0.522 & 0.513 \\

\midrule

\textbf{all-MiniLM-L6-v2} & 0.503 & 0.321 & 0.478 & 0.548 & 0.491 & 0.588 \\
\textbf{bert-base-nli-mean-tokens KR} & 0.523 & 0.325 & 0.502 & 0.545 & 0.479 & 0.597 \\

\bottomrule

\end{tabular}
\label{tab:reconstruction_task}
}
\end{table*}

\section{Conclusions and Future Work}
\label{sec:conclusions}
In conclusion, our work provides a novel theoretical view to understand the intriguing in-context learning (ICL) capabilities of Transformer-based large language models (LLMs). We propose that LLMs ICL can be understood as kernel regression. Our empirical investigations into the in-context behaviors of LLMs reveal that the model's attention and hidden features during ICL are congruent with the behaviors of kernel regression. Furthermore, our theory also explains several observable phenomena in the field of ICL: why the retrieval of demonstrations similar to the test sample can enhance performance, the sensitivity of ICL to output formats, and the benefit from selecting in-distribution and representative samples.
There are still remaining challenges in this topic, such as understanding the effect of sample orderings and the robustness to perturbed labels. These questions, along with understanding other perspectives of LLMs, are exciting questions for future research.


\section{Limitations}
\label{sec:limitations}

This work is based on several assumptions, such as the HMM mixture setting described in Section~\ref{sec:formulation}, and more specifically the assumptions in Section~\ref{subsec:assumptions}. Though theoretically, the HMM can be exhaustively constructed to express any finite-length distribution (e.g., by using the sequences of prefix tokens as hidden states $\mathcal{S} = \{[o_0, o_1, \ldots, o_i] | i\in\mathbb{N}^\star, o_i\in \mathcal{O} \}$), it might not be an elegant model of natural language. It does not fully capture the full complexity and generality of natural language with unbounded lengths, and might not be the most efficient model considering the large number of hidden states needed.

There is also controversy that challenges the Bayesian inference view of ICL behaviors. Some work such as ~\cite{falck2024context} is discussed in Section~\ref{sec:related}. Besides, not all ICL phenomena can be explained as Bayesian inference, such as those discussed in Section~\ref{subsec:insights}. The divergence from kernel-regression behaviors can be partly explained by the first error term in Equation~\ref{eq:convergence}, as detailed in Section~\ref{sec:related}.

\section*{Acknowledgments}
\label{sec:acknowledgments}

This work was supported in part by US DARPA KAIROS Program No. FA8750-19-2-1004 and AIDA Program No. FA8750-18-2-0014.
The views and conclusions contained in this document are those of the authors and should not be interpreted as representing the official policies, either expressed or implied, of the U.S. Government. The U.S. Government is authorized to reproduce and distribute reprints for Government purposes notwithstanding any copyright notation here on.

\bibliography{custom}

\begin{thebibliography}{50}
\providecommand{\natexlab}[1]{#1}
\providecommand{\url}[1]{\texttt{#1}}
\expandafter\ifx\csname urlstyle\endcsname\relax
  \providecommand{\doi}[1]{doi: #1}\else
  \providecommand{\doi}{doi: \begingroup \urlstyle{rm}\Url}\fi

\bibitem[Ahn et~al.(2023)Ahn, Cheng, Daneshmand, and Sra]{ahn2023transformers}
Kwangjun Ahn, Xiang Cheng, Hadi Daneshmand, and Suvrit Sra.
\newblock Transformers learn to implement preconditioned gradient descent for
  in-context learning.
\newblock \emph{Advances in Neural Information Processing Systems},
  36:\penalty0 45614--45650, 2023.

\bibitem[Aky{\"u}rek et~al.(2023)Aky{\"u}rek, Schuurmans, Andreas, Ma, and
  Zhou]{akyrek2023what}
Ekin Aky{\"u}rek, Dale Schuurmans, Jacob Andreas, Tengyu Ma, and Denny Zhou.
\newblock What learning algorithm is in-context learning? investigations with
  linear models.
\newblock In \emph{The Eleventh International Conference on Learning
  Representations}, 2023.
\newblock URL \url{https://openreview.net/forum?id=0g0X4H8yN4I}.

\bibitem[Arora et~al.()Arora, Eyuboglu, Timalsina, Johnson, Poli, Zou, Rudra,
  and Re]{arorazoology}
Simran Arora, Sabri Eyuboglu, Aman Timalsina, Isys Johnson, Michael Poli, James
  Zou, Atri Rudra, and Christopher Re.
\newblock Zoology: Measuring and improving recall in efficient language models.
\newblock In \emph{The Twelfth International Conference on Learning
  Representations}.

\bibitem[Bai et~al.(2024)Bai, Chen, Wang, Xiong, and Mei]{bai2024transformers}
Yu~Bai, Fan Chen, Huan Wang, Caiming Xiong, and Song Mei.
\newblock Transformers as statisticians: Provable in-context learning with
  in-context algorithm selection.
\newblock \emph{Advances in neural information processing systems}, 36, 2024.

\bibitem[Bietti et~al.(2023)Bietti, Cabannes, Bouchacourt, Jegou, and
  Bottou]{bietti2023birth}
Alberto Bietti, Vivien Cabannes, Diane Bouchacourt, Herve Jegou, and Leon
  Bottou.
\newblock Birth of a transformer: A memory viewpoint.
\newblock \emph{Advances in Neural Information Processing Systems},
  36:\penalty0 1560--1588, 2023.

\bibitem[Brown et~al.(2020)Brown, Mann, Ryder, Subbiah, Kaplan, Dhariwal,
  Neelakantan, Shyam, Sastry, Askell, et~al.]{brown2020language}
Tom Brown, Benjamin Mann, Nick Ryder, Melanie Subbiah, Jared~D Kaplan, Prafulla
  Dhariwal, Arvind Neelakantan, Pranav Shyam, Girish Sastry, Amanda Askell,
  et~al.
\newblock Language models are few-shot learners.
\newblock \emph{Advances in neural information processing systems},
  33:\penalty0 1877--1901, 2020.

\bibitem[Chan et~al.(2022)Chan, Santoro, Lampinen, Wang, Singh, Richemond,
  McClelland, and Hill]{chan2022data}
Stephanie Chan, Adam Santoro, Andrew Lampinen, Jane Wang, Aaditya Singh, Pierre
  Richemond, James McClelland, and Felix Hill.
\newblock Data distributional properties drive emergent in-context learning in
  transformers.
\newblock \emph{Advances in Neural Information Processing Systems},
  35:\penalty0 18878--18891, 2022.

\bibitem[Chen \& Li(2023)Chen and Li]{chen2023calibrating}
Wenlong Chen and Yingzhen Li.
\newblock Calibrating transformers via sparse gaussian processes.
\newblock \emph{arXiv preprint arXiv:2303.02444}, 2023.

\bibitem[Dai et~al.(2022)Dai, Sun, Dong, Hao, Ma, Sui, and Wei]{dai2023can}
Damai Dai, Yutao Sun, Li~Dong, Yaru Hao, Shuming Ma, Zhifang Sui, and Furu Wei.
\newblock Why can gpt learn in-context? language models implicitly perform
  gradient descent as meta-optimizers.
\newblock In \emph{ICLR 2023 Workshop on Mathematical and Empirical
  Understanding of Foundation Models}, 2022.

\bibitem[Falck et~al.(2024)Falck, Wang, and Holmes]{falck2024context}
Fabian Falck, Ziyu Wang, and Christopher~C Holmes.
\newblock Is in-context learning in large language models bayesian? a
  martingale perspective.
\newblock In \emph{International Conference on Machine Learning}, pp.\
  12784--12805. PMLR, 2024.

\bibitem[Fu et~al.(2024)Fu, Chen, Jia, and Sharan]{fu2024transformers}
Deqing Fu, Tian-Qi Chen, Robin Jia, and Vatsal Sharan.
\newblock Transformers learn to achieve second-order convergence rates for
  in-context linear regression.
\newblock In \emph{The Thirty-eighth Annual Conference on Neural Information
  Processing Systems}, 2024.

\bibitem[Garg et~al.(2023)Garg, Tsipras, Liang, and Valiant]{gargcan}
Shivam Garg, Dimitris Tsipras, Percy Liang, and Gregory Valiant.
\newblock What can transformers learn in-context? a case study of simple
  function classes.
\newblock In \emph{Advances in Neural Information Processing Systems}, 2023.

\bibitem[Grattafiori et~al.(2024)Grattafiori, Dubey, Jauhri, Pandey, Kadian,
  Al-Dahle, Letman, Mathur, Schelten, Vaughan, et~al.]{grattafiori2024llama}
Aaron Grattafiori, Abhimanyu Dubey, Abhinav Jauhri, Abhinav Pandey, Abhishek
  Kadian, Ahmad Al-Dahle, Aiesha Letman, Akhil Mathur, Alan Schelten, Alex
  Vaughan, et~al.
\newblock The llama 3 herd of models.
\newblock \emph{arXiv preprint arXiv:2407.21783}, 2024.

\bibitem[Guo et~al.()Guo, Hu, Mei, Wang, Xiong, Savarese, and
  Bai]{guotransformers}
Tianyu Guo, Wei Hu, Song Mei, Huan Wang, Caiming Xiong, Silvio Savarese, and
  Yu~Bai.
\newblock How do transformers learn in-context beyond simple functions? a case
  study on learning with representations.
\newblock In \emph{The Twelfth International Conference on Learning
  Representations}.

\bibitem[Huang et~al.()Huang, Cheng, and Liang]{huangcontext}
Yu~Huang, Yuan Cheng, and Yingbin Liang.
\newblock In-context convergence of transformers.
\newblock In \emph{Forty-first International Conference on Machine Learning}.

\bibitem[Jaeger(2000)]{jaeger2000observable}
Herbert Jaeger.
\newblock Observable operator models for discrete stochastic time series.
\newblock \emph{Neural computation}, 12\penalty0 (6):\penalty0 1371--1398,
  2000.

\bibitem[Khashabi et~al.(2022)Khashabi, Baral, Choi, and
  Hajishirzi]{khashabi2022reframing}
Daniel Khashabi, Chitta Baral, Yejin Choi, and Hannaneh Hajishirzi.
\newblock Reframing instructional prompts to gptk’s language.
\newblock In \emph{Findings of the Association for Computational Linguistics:
  ACL 2022}, pp.\  589--612, 2022.

\bibitem[Kim \& Suzuki()Kim and Suzuki]{kimtransformers}
Juno Kim and Taiji Suzuki.
\newblock Transformers learn nonlinear features in context: Nonconvex
  mean-field dynamics on the attention landscape.
\newblock In \emph{Forty-first International Conference on Machine Learning}.

\bibitem[Kojima et~al.(2023)Kojima, Gu, Reid, Matsuo, and Iwasawa]{kojimalarge}
Takeshi Kojima, Shixiang~Shane Gu, Machel Reid, Yutaka Matsuo, and Yusuke
  Iwasawa.
\newblock Large language models are zero-shot reasoners.
\newblock In \emph{Advances in Neural Information Processing Systems}, 2023.

\bibitem[Kossen et~al.(2023)Kossen, Rainforth, and Gal]{kossen2023context}
Jannik Kossen, Tom Rainforth, and Yarin Gal.
\newblock In-context learning in large language models learns label
  relationships but is not conventional learning.
\newblock \emph{arXiv preprint arXiv:2307.12375}, 2023.

\bibitem[Li et~al.(2023)Li, Zhao, Li, Ji, Callison-Burch, and
  Han]{hierarchicalschema2023}
Sha Li, Ruining Zhao, Manling Li, Heng Ji, Chris Callison-Burch, and Jiawei
  Han.
\newblock Open-domain hierarchical event schema induction by incremental
  prompting and verification.
\newblock In \emph{Proc. The 61st Annual Meeting of the Association for
  Computational Linguistics (ACL2023)}, 2023.

\bibitem[Liu et~al.(2022)Liu, Shen, Zhang, Dolan, Carin, and
  Chen]{liu2022makes}
Jiachang Liu, Dinghan Shen, Yizhe Zhang, William~B Dolan, Lawrence Carin, and
  Weizhu Chen.
\newblock What makes good in-context examples for gpt-3?
\newblock In \emph{Proceedings of Deep Learning Inside Out (DeeLIO 2022): The
  3rd Workshop on Knowledge Extraction and Integration for Deep Learning
  Architectures}, pp.\  100--114, 2022.

\bibitem[Logan~IV et~al.(2022)Logan~IV, Bala{\v{z}}evi{\'c}, Wallace, Petroni,
  Singh, and Riedel]{logan2022cutting}
Robert Logan~IV, Ivana Bala{\v{z}}evi{\'c}, Eric Wallace, Fabio Petroni, Sameer
  Singh, and Sebastian Riedel.
\newblock Cutting down on prompts and parameters: Simple few-shot learning with
  language models.
\newblock In \emph{Findings of the Association for Computational Linguistics:
  ACL 2022}, pp.\  2824--2835, 2022.

\bibitem[Lu et~al.(2022)Lu, Bartolo, Moore, Riedel, and
  Stenetorp]{lu2022fantastically}
Yao Lu, Max Bartolo, Alastair Moore, Sebastian Riedel, and Pontus Stenetorp.
\newblock Fantastically ordered prompts and where to find them: Overcoming
  few-shot prompt order sensitivity.
\newblock In \emph{Proceedings of the 60th Annual Meeting of the Association
  for Computational Linguistics (Volume 1: Long Papers)}, pp.\  8086--8098,
  2022.

\bibitem[Lu et~al.(2024)Lu, Letey, Zavatone-Veth, Maiti, and
  Pehlevan]{lu2024asymptotic}
Yue~M Lu, Mary~I Letey, Jacob~A Zavatone-Veth, Anindita Maiti, and Cengiz
  Pehlevan.
\newblock Asymptotic theory of in-context learning by linear attention.
\newblock \emph{arXiv preprint arXiv:2405.11751}, 2024.

\bibitem[Mahankali et~al.()Mahankali, Hashimoto, and Ma]{mahankalione}
Arvind~V Mahankali, Tatsunori Hashimoto, and Tengyu Ma.
\newblock One step of gradient descent is provably the optimal in-context
  learner with one layer of linear self-attention.
\newblock In \emph{The Twelfth International Conference on Learning
  Representations}.

\bibitem[Min et~al.(2022{\natexlab{a}})Min, Lewis, Hajishirzi, and
  Zettlemoyer]{min2022noisy}
Sewon Min, Mike Lewis, Hannaneh Hajishirzi, and Luke Zettlemoyer.
\newblock Noisy channel language model prompting for few-shot text
  classification.
\newblock In \emph{Proceedings of the 60th Annual Meeting of the Association
  for Computational Linguistics (Volume 1: Long Papers)}, pp.\  5316--5330,
  2022{\natexlab{a}}.

\bibitem[Min et~al.(2022{\natexlab{b}})Min, Lyu, Holtzman, Artetxe, Lewis,
  Hajishirzi, and Zettlemoyer]{min2022rethinking}
Sewon Min, Xinxi Lyu, Ari Holtzman, Mikel Artetxe, Mike Lewis, Hannaneh
  Hajishirzi, and Luke Zettlemoyer.
\newblock Rethinking the role of demonstrations: What makes in-context learning
  work?
\newblock \emph{arXiv preprint arXiv:2202.12837}, 2022{\natexlab{b}}.

\bibitem[Olsson et~al.(2022)Olsson, Elhage, Nanda, Joseph, DasSarma, Henighan,
  Mann, Askell, Bai, Chen, et~al.]{olsson2022context}
Catherine Olsson, Nelson Elhage, Neel Nanda, Nicholas Joseph, Nova DasSarma,
  Tom Henighan, Ben Mann, Amanda Askell, Yuntao Bai, Anna Chen, et~al.
\newblock In-context learning and induction heads.
\newblock \emph{arXiv preprint arXiv:2209.11895}, 2022.

\bibitem[Park et~al.(2024)Park, Lubana, Pres, and Tanaka]{park2024competition}
Core~Francisco Park, Ekdeep~Singh Lubana, Itamar Pres, and Hidenori Tanaka.
\newblock Competition dynamics shape algorithmic phases of in-context learning.
\newblock \emph{arXiv preprint arXiv:2412.01003}, 2024.

\bibitem[Radford et~al.(2023)Radford, Wu, Child, Luan, Amodei, Sutskever,
  et~al.]{radford2019language}
Alec Radford, Jeffrey Wu, Rewon Child, David Luan, Dario Amodei, Ilya
  Sutskever, et~al.
\newblock Language models are unsupervised multitask learners.
\newblock 2023.

\bibitem[Ramsauer et~al.()Ramsauer, Sch{\"a}fl, Lehner, Seidl, Widrich, Gruber,
  Holzleitner, Adler, Kreil, Kopp, et~al.]{ramsauerhopfield}
Hubert Ramsauer, Bernhard Sch{\"a}fl, Johannes Lehner, Philipp Seidl, Michael
  Widrich, Lukas Gruber, Markus Holzleitner, Thomas Adler, David Kreil,
  Michael~K Kopp, et~al.
\newblock Hopfield networks is all you need.
\newblock In \emph{International Conference on Learning Representations}.

\bibitem[Ravent{\'o}s et~al.(2023)Ravent{\'o}s, Paul, Chen, and
  Ganguli]{raventos2023pretraining}
Allan Ravent{\'o}s, Mansheej Paul, Feng Chen, and Surya Ganguli.
\newblock Pretraining task diversity and the emergence of non-bayesian
  in-context learning for regression.
\newblock \emph{Advances in neural information processing systems},
  36:\penalty0 14228--14246, 2023.

\bibitem[Reimers \& Gurevych(2019)Reimers and
  Gurevych]{reimers-2019-sentence-bert}
Nils Reimers and Iryna Gurevych.
\newblock Sentence-bert: Sentence embeddings using siamese bert-networks.
\newblock In \emph{Proceedings of the 2019 Conference on Empirical Methods in
  Natural Language Processing}. Association for Computational Linguistics, 11
  2019.
\newblock URL \url{http://arxiv.org/abs/1908.10084}.

\bibitem[Ren \& Liu(2023)Ren and Liu]{ren2023towards}
Ruifeng Ren and Yong Liu.
\newblock Towards understanding how transformers learn in-context through a
  representation learning lens.
\newblock \emph{arXiv preprint arXiv:2310.13220}, 2023.

\bibitem[Rubin et~al.(2022)Rubin, Herzig, and Berant]{rubin2022learning}
Ohad Rubin, Jonathan Herzig, and Jonathan Berant.
\newblock Learning to retrieve prompts for in-context learning.
\newblock In \emph{Proceedings of the 2022 Conference of the North American
  Chapter of the Association for Computational Linguistics: Human Language
  Technologies}, pp.\  2655--2671, 2022.

\bibitem[Touvron et~al.(2023)Touvron, Martin, Stone, Albert, Almahairi, Babaei,
  Bashlykov, Batra, Bhargava, Bhosale, et~al.]{touvron2023llama}
Hugo Touvron, Louis Martin, Kevin Stone, Peter Albert, Amjad Almahairi, Yasmine
  Babaei, Nikolay Bashlykov, Soumya Batra, Prajjwal Bhargava, Shruti Bhosale,
  et~al.
\newblock Llama 2: Open foundation and fine-tuned chat models.
\newblock \emph{arXiv preprint arXiv:2307.09288}, 2023.

\bibitem[Vaswani et~al.(2017)Vaswani, Shazeer, Parmar, Uszkoreit, Jones, Gomez,
  Kaiser, and Polosukhin]{vaswani2017attention}
Ashish Vaswani, Noam Shazeer, Niki Parmar, Jakob Uszkoreit, Llion Jones,
  Aidan~N Gomez, {\L}ukasz Kaiser, and Illia Polosukhin.
\newblock Attention is all you need.
\newblock \emph{Advances in neural information processing systems}, 30, 2017.

\bibitem[von Oswald et~al.(2022)von Oswald, Niklasson, Randazzo, Sacramento,
  Mordvintsev, Zhmoginov, and Vladymyrov]{von2022transformers}
Johannes von Oswald, Eyvind Niklasson, Ettore Randazzo, Jo{\~a}o Sacramento,
  Alexander Mordvintsev, Andrey Zhmoginov, and Max Vladymyrov.
\newblock Transformers learn in-context by gradient descent.
\newblock \emph{arXiv preprint arXiv:2212.07677}, 2022.

\bibitem[Wang \& Komatsuzaki(2021)Wang and Komatsuzaki]{gpt-j}
Ben Wang and Aran Komatsuzaki.
\newblock {GPT-J-6B: A 6 Billion Parameter Autoregressive Language Model}.
\newblock \url{https://github.com/kingoflolz/mesh-transformer-jax}, May 2021.

\bibitem[Wang et~al.(2023)Wang, Zhu, and Wang]{wang2023large}
Xinyi Wang, Wanrong Zhu, and William~Yang Wang.
\newblock Large language models are implicitly topic models: Explaining and
  finding good demonstrations for in-context learning.
\newblock \emph{arXiv preprint arXiv:2301.11916}, 2023.

\bibitem[Wang \& Ji(2022)Wang and Ji]{eeg2022}
Zhenhailong Wang and Heng Ji.
\newblock Open vocabulary electroencephalography-to-text decoding and zero-shot
  sentiment classification.
\newblock In \emph{Proc. Thirty-Sixth AAAI Conference on Artificial
  Intelligence (AAAI2022)}, 2022.

\bibitem[Wei et~al.(2022{\natexlab{a}})Wei, Tay, Bommasani, Raffel, Zoph,
  Borgeaud, Yogatama, Bosma, Zhou, Metzler, et~al.]{weiemergent}
Jason Wei, Yi~Tay, Rishi Bommasani, Colin Raffel, Barret Zoph, Sebastian
  Borgeaud, Dani Yogatama, Maarten Bosma, Denny Zhou, Donald Metzler, et~al.
\newblock Emergent abilities of large language models.
\newblock \emph{Transactions on Machine Learning Research}, 2022{\natexlab{a}}.

\bibitem[Wei et~al.(2022{\natexlab{b}})Wei, Wang, Schuurmans, Bosma, Xia, Chi,
  Le, Zhou, et~al.]{weichain}
Jason Wei, Xuezhi Wang, Dale Schuurmans, Maarten Bosma, Fei Xia, Ed~H Chi,
  Quoc~V Le, Denny Zhou, et~al.
\newblock Chain-of-thought prompting elicits reasoning in large language
  models.
\newblock In \emph{Advances in Neural Information Processing Systems},
  2022{\natexlab{b}}.

\bibitem[Wei et~al.(2023)Wei, Wei, Tay, Tran, Webson, Lu, Chen, Liu, Huang,
  Zhou, et~al.]{wei2023larger}
Jerry Wei, Jason Wei, Yi~Tay, Dustin Tran, Albert Webson, Yifeng Lu, Xinyun
  Chen, Hanxiao Liu, Da~Huang, Denny Zhou, et~al.
\newblock Larger language models do in-context learning differently.
\newblock \emph{arXiv preprint arXiv:2303.03846}, 2023.

\bibitem[Wu et~al.(2022)Wu, Li, and Liang]{wu2022insights}
Yuhuai Wu, Felix Li, and Percy~S Liang.
\newblock Insights into pre-training via simpler synthetic tasks.
\newblock \emph{Advances in Neural Information Processing Systems},
  35:\penalty0 21844--21857, 2022.

\bibitem[Xie et~al.(2022)Xie, Raghunathan, Liang, and Ma]{xieexplanation}
Sang~Michael Xie, Aditi Raghunathan, Percy Liang, and Tengyu Ma.
\newblock An explanation of in-context learning as implicit bayesian inference.
\newblock In \emph{International Conference on Learning Representations}, 2022.

\bibitem[Xie(2021)]{xie2021gx}
Wanying Xie.
\newblock Gx at semeval-2021 task 2: Bert with lemma information for mcl-wic
  task.
\newblock In \emph{Proceedings of the 15th International Workshop on Semantic
  Evaluation (SemEval-2021)}, pp.\  706--712, 2021.

\bibitem[Zhang et~al.(2024)Zhang, Frei, and Bartlett]{zhang2024trained}
Ruiqi Zhang, Spencer Frei, and Peter~L Bartlett.
\newblock Trained transformers learn linear models in-context.
\newblock \emph{Journal of Machine Learning Research}, 25\penalty0
  (49):\penalty0 1--55, 2024.

\bibitem[Zhao et~al.(2021)Zhao, Wallace, Feng, Klein, and
  Singh]{zhao2021calibrate}
Zihao Zhao, Eric Wallace, Shi Feng, Dan Klein, and Sameer Singh.
\newblock Calibrate before use: Improving few-shot performance of language
  models.
\newblock In \emph{International Conference on Machine Learning}, pp.\
  12697--12706. PMLR, 2021.

\end{thebibliography}
\bibliographystyle{tmlr}

\appendix
\section{Proofs}
\label{appsec:proof}

\begin{proof}

First, we denote the kernel regression function
\begin{equation}    
    \hat{\bfy} = \frac{1}{n} \sum_{i=1}^n \left< \text{vec}(T_{\bfx_\text{test}}), \Sigma_{p_0}^{-1} \text{vec}(T_{\bfx_i}) \right> \mathbf{e}(y_i),
\end{equation}

In expectation,

\begin{equation}
    \mathbb{E}_{\bfx_i} \left< \text{vec}(T_{\bfx_\text{test}}), \Sigma_{p_{\theta^\star}}^{-1} ~ \text{vec}(T_{\bfx_i}) \right> \text{vec}(T_{\bfx_i})^\top 
    = \text{vec}(T_{\bfx_\text{test}})^\top \Sigma_{p_{\theta^\star}}^{-1} \mathbb{E}_{\bfx_i} \text{vec}(T_{\bfx_i}) \text{vec}(T_{\bfx_i})^\top
    = \text{vec}(T_{\bfx_\text{test}})^\top.
\end{equation}

As $\text{vec}(\cdot)$ is a linear function, if we replace $\text{vec}(T_{\bfx_i})^\top$ with $T_{\bfx_i}$:
\begin{equation}
    \mathbb{E}_{\bfx_i} \left< \text{vec}(T_{\bfx_\text{test}}), \Sigma_{p_{\theta^\star}}^{-1} ~~ \text{vec}(T_{\bfx_i}) \right> T_{\bfx_i} = T_{\bfx_\text{test}}
\end{equation}
and then

\begin{align}
    &\mathbb{E}_{\bfx_i} \left< \text{vec}(T_{\bfx_\text{test}}), \Sigma_{p_{\theta^\star}}^{-1} \text{vec}(T_{\bfx_i}) \right> P(Y|\bfx_i, p_{\theta^\star}) \\
    =& P(Y|\mathcal{S})^\top \mathbb{E}_{\bfx_i} \left< \text{vec}(T_{\bfx_\text{test}}), \Sigma_{p_{\theta^\star}}^{-1} ~~ \text{vec}(T_{\bfx_i}) \right> T_{\bfx_i} p_{\theta^\star} \\
    =& P(Y|\mathcal{S})^\top T_{\bfx_\text{test}} p_{\theta^\star} \\
    =& P(Y|\bfx_\text{test}, \theta^\star)
\end{align}

As $(\bfx_i, \mathbf{e}(y_i))$ can be seen as independent samples from $P(Y|\bfx_i, p_{\theta^\star})$, we can use Hoeffding's inequality and bound that, with $1-\frac{\delta}{2}$ probability,

\[
    \|\hat{\bfy}
    -
    \mathbb{E}_{\bfx_i} \left< \text{vec}(T_{\bfx_\text{test}}), \Sigma_{p_{\theta^\star}}^{-1} \text{vec}(T_{\bfx_i}) \right> P(Y|\bfx_i, p_{\theta^\star})\|_\infty
    \leq \sqrt{\frac{1}{2n} \ln\frac{4m}{\delta}}
\]

Considering the difference between $\Sigma_{p_{\theta^\star}}$ and $\Sigma_{p_0}$, we see that
\begin{align}
\label{eq:prior_info_term}
    &|\left< \text{vec}(T_{\bfx_\text{test}}), \Sigma_{p_{\theta^\star}}^{-1} \text{vec}(T_{\bfx_i}) \right>
    -
    \left< \text{vec}(T_{\bfx_\text{test}}), \Sigma_{p_0}^{-1} \text{vec}(T_{\bfx_i}) \right>| \\
    =& |\text{vec} (T_{\bfx_\text{test}})^\top (\Sigma_{p_{\theta^\star}}^{-1} - \Sigma_{p_0}^{-1}) \text{vec}(T_{\bfx_i})| \\
    \leq & \eta^2 \epsilon_\theta
\end{align}

Therefore,
\[
    \|\hat{\bfy}
    -
    P(Y|\bfx_\text{test}, \theta^\star)\|_\infty
    \leq \sqrt{\frac{1}{2n} \ln\frac{4m}{\delta}} + \eta^2\epsilon_\theta
\]

Next we bridge $P(Y|\bfx_\text{test}, \theta^\star)$ with $P(Y|\bfo_{ICL}, \pretrain)$. Let $s_\text{test}$ be the hidden state corresponding to first token of $\bfx_\text{test}$, i.e., $\bfx_{\text{test},0}$. We see that, the likelihood of $s_\text{test}=s_{\theta^\star}$ is lower bounded by:
\begin{align}
    P(s_\text{test}=s_{\theta^\star}, S_n | \pretrain)
    &= \sum_{\theta\in\Theta} P(s_\text{test}=s_{\theta^\star}| S_n, s_\theta) P(S_n|s_\theta) P(s_\theta|\pretrain) \\
    &= P(s_\text{test}=s_{\theta^\star}| S_n, s_{\theta^\star}) P(S_n|s_{\theta^\star}) P(s_{\theta^\star}|\pretrain) \\
    &\geq P(S_n|s_{\theta^\star}) P(s_{\theta^\star}|\pretrain)\epsilon_r \\
    \text{(Markov property)}
    &\geq \left(\prod_{i=1}^n P([\bfx_i, y_i,\delim]|s_{\theta^\star}) P(s_{\theta^\star}|[\bfx_i, y_i, \delim],s_{\theta^\star}) \right) P(s_{\theta^\star}|\pretrain)\epsilon_r \\
    \text{(by Assumption~\ref{assumption:anchor})}
    &\geq \left(\prod_{i=1}^n P([\bfx_i, y_i]|s_{\theta^\star}) \epsilon_d \epsilon_r \right) P(s_{\theta^\star}|\pretrain)\epsilon_r \\
    &\geq \left(\prod_{i=1}^n P([\bfx_i, y_i]|s_{\theta^\star}) \right) P(s_{\theta^\star}|\pretrain) \epsilon_r^{n+1}\epsilon_d^n
\end{align}

For another task $\theta'$, $s_\text{test}$ is unlikely to be $s_{\theta'}$ because:
\begin{align}
    P(s_\text{test}=s_{\theta'}, S_n | \pretrain)
    &= \sum_{\theta\in\Theta} P(s_\text{test}=s_{\theta'}| S_n, s_\theta) P(S_n|s_\theta) P(s_\theta|\pretrain) \\
    &= P(s_\text{test}=s_{\theta'}| S_n, s_{\theta'}) P(S_n|s_{\theta'}) P(s_{\theta'}|\pretrain) \\
    \text{(by Assumption~\ref{assumption:anchor})}
    &\leq \left(\prod_{i=1}^n P([\bfx_i, y_i, \delim]|\theta')\right) P(s_{\theta'}|\pretrain) \\
    &\leq \left(\prod_{i=1}^n P([\bfx_i, y_i]|\theta')\right) P(s_{\theta'}|\pretrain)
\end{align}

Therefore, the Bayesian inference over $s_\text{test}$, is:

\begin{align}
    &P(s_\text{test}=s_{\theta^\star}| \bfo_{ICL}, \pretrain) \\
    =& \frac{P(s_\text{test}=s_{\theta^\star}, \bfo_{ICL}| \pretrain)}{P(\bfo_{ICL}| \pretrain)} \\
    =& \frac{P(s_\text{test}=s_{\theta^\star}, \bfo_{ICL}| \pretrain)}{\sum_{\theta} P(s_\text{test}=s_\theta, \bfo_{ICL}| \pretrain)} \\
    =& \left(\sum_{\theta}\frac{\left(P(\xtest | \theta')\prod_{i=1}^n P([\bfx_i, y_i]|\theta')\right) P(s_{\theta'}|\pretrain)}{P(s_{\theta^\star}, \bfo_{ICL}| \pretrain)} \right)^{-1} \\
    \geq& \left(1+\min_{\theta\ne\theta^\star}\text{exp}\left(  \ln\frac{P(\xtest|\theta)}{P(\xtest|\theta^\star)} +  \sum_{i=1}^n\ln\frac{P([\bfx_i,y_i]|\theta)}{P([\bfx_i,y_i]|\theta^\star)} +n\ln\frac{1}{\epsilon_d}+(n+1)\ln\frac{1}{\epsilon_r}+\ln\frac{1}{\|p_0\|_{-\infty}}\right)\right)^{-1} \\
    & (\text{with } 1-\frac{\delta}{2} \text{prob.}) \\
    \geq & \left(1+\min_{\theta\ne\theta^\star}\text{exp}\left(  -n\epsilon_{KL} + \sqrt{\frac{1}{n}\ln\frac{4}{\delta}} +n\ln\frac{1}{\epsilon_d}+(n+1)\ln\frac{1}{\epsilon_r}+\ln\frac{1}{\|p_0\|_{-\infty}}\right)\right)^{-1} \\
    \geq & 1-\text{exp}\left(  -n\epsilon_{KL} + \sqrt{\frac{1}{n}\ln\frac{4}{\delta}} +n\ln\frac{1}{\epsilon_d}+(n+1)\ln\frac{1}{\epsilon_r}+\ln\frac{1}{\|p_0\|_{-\infty}}\right)
\end{align}

So that

\begin{align}
    & \|\hat{\bfy}
    -
    P(Y|\bfo_{ICL}, \pretrain)\|_\infty \\
    \leq & \sqrt{\frac{1}{n} \ln\frac{4m}{\delta}} + \eta^2\epsilon_\theta + \frac{1}{\epsilon_r \|p_0\|_{-\infty}} \text{exp}\left(  -n(\epsilon_{KL}+\ln(\epsilon_d \epsilon_r) + \sqrt{\frac{1}{n}\ln\frac{4}{\delta}}\right) \\
    &= O\left(\sqrt{\frac{1}{n} \ln\frac{4m}{\delta}}\right) + \eta^2\epsilon_\theta
\end{align}





\end{proof}

\section{Results on more tasks}
\label{appsec:more_results}

Besides the case study on SST2 dataset in Section~\ref{sec:empirical}, in this section we also provide experiment results on other tasks. In specific, we experiment on 
Rotten Tomatoes\footnote{\url{https://huggingface.co/datasets/rotten_tomatoes/}},
Tweet Eval\footnote{\url{https://huggingface.co/datasets/tweet_eval/}}'s (\texttt{hate}, \texttt{irony} and \texttt{offensive} subtasks)
and MNLI\footnote{\url{https://huggingface.co/datasets/glue/viewer/mnli_matched/test}}. The results are as follows.


\subsection{Rotten Tomatoes}
\begin{figure*}[ht]
\centering
\includegraphics[width=\textwidth]{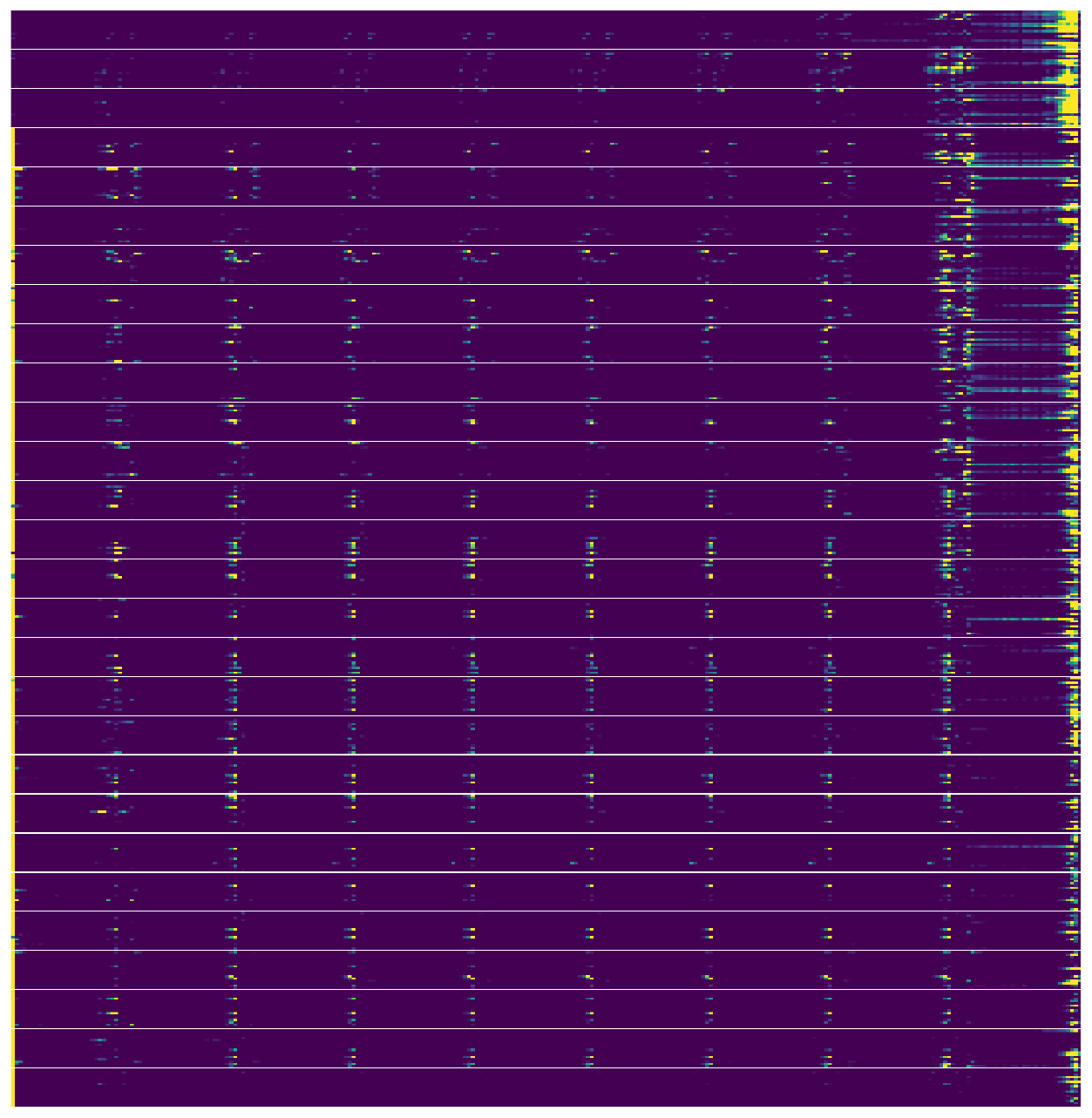}
\caption{Averaged attention map over Rotten Tomatoes test set.}
\end{figure*}

\begin{figure}
     \centering
     \begin{subfigure}[b]{0.47\textwidth}
         \centering
         \includegraphics[width=\textwidth]{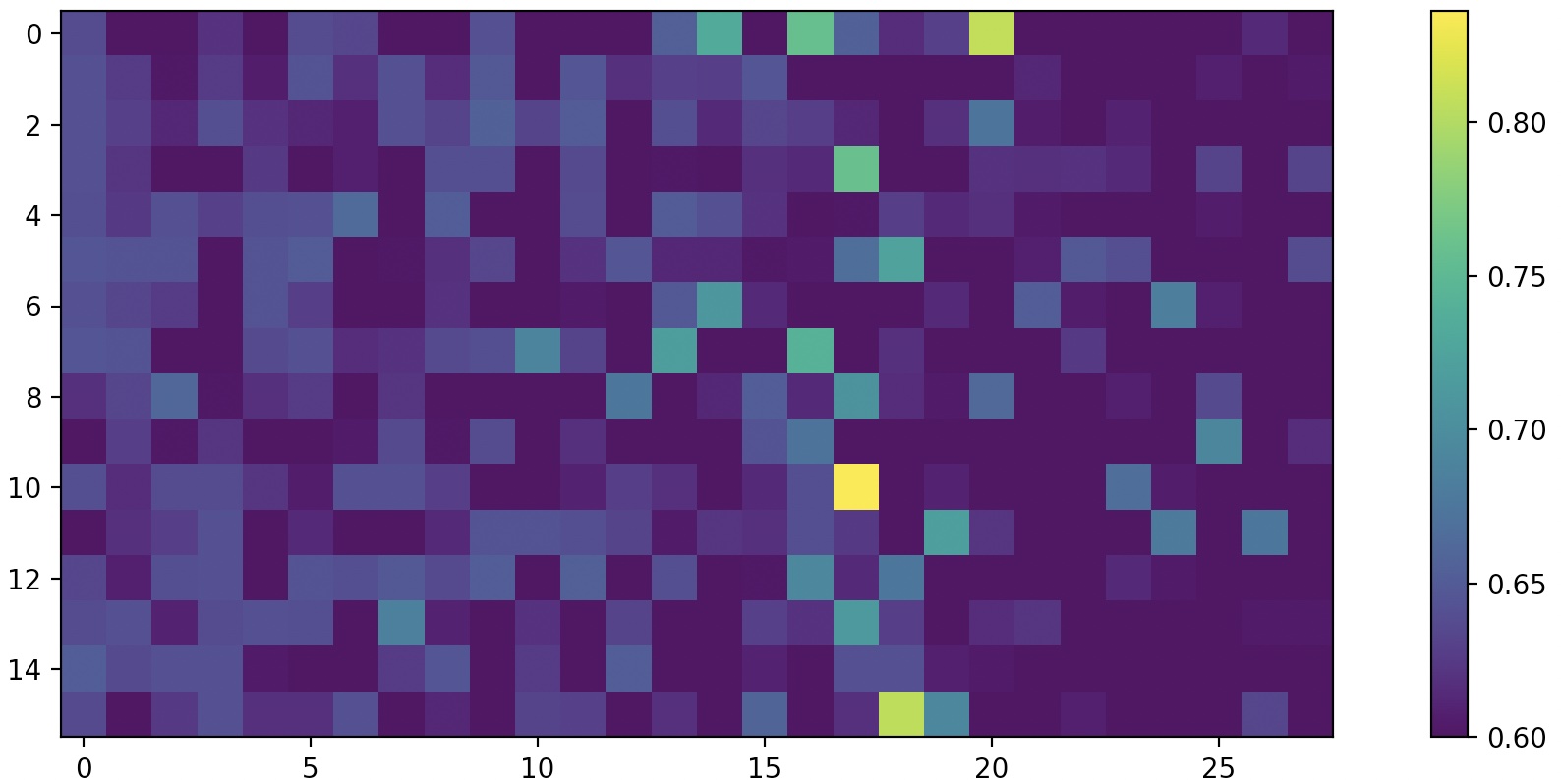}
         \caption{Accuracy on reconstruction of $\hat y$ by interpreting attention as kernel weights.}
     \end{subfigure}
     \hfill
     \begin{subfigure}[b]{0.47\textwidth}
         \centering
         \includegraphics[width=\textwidth]{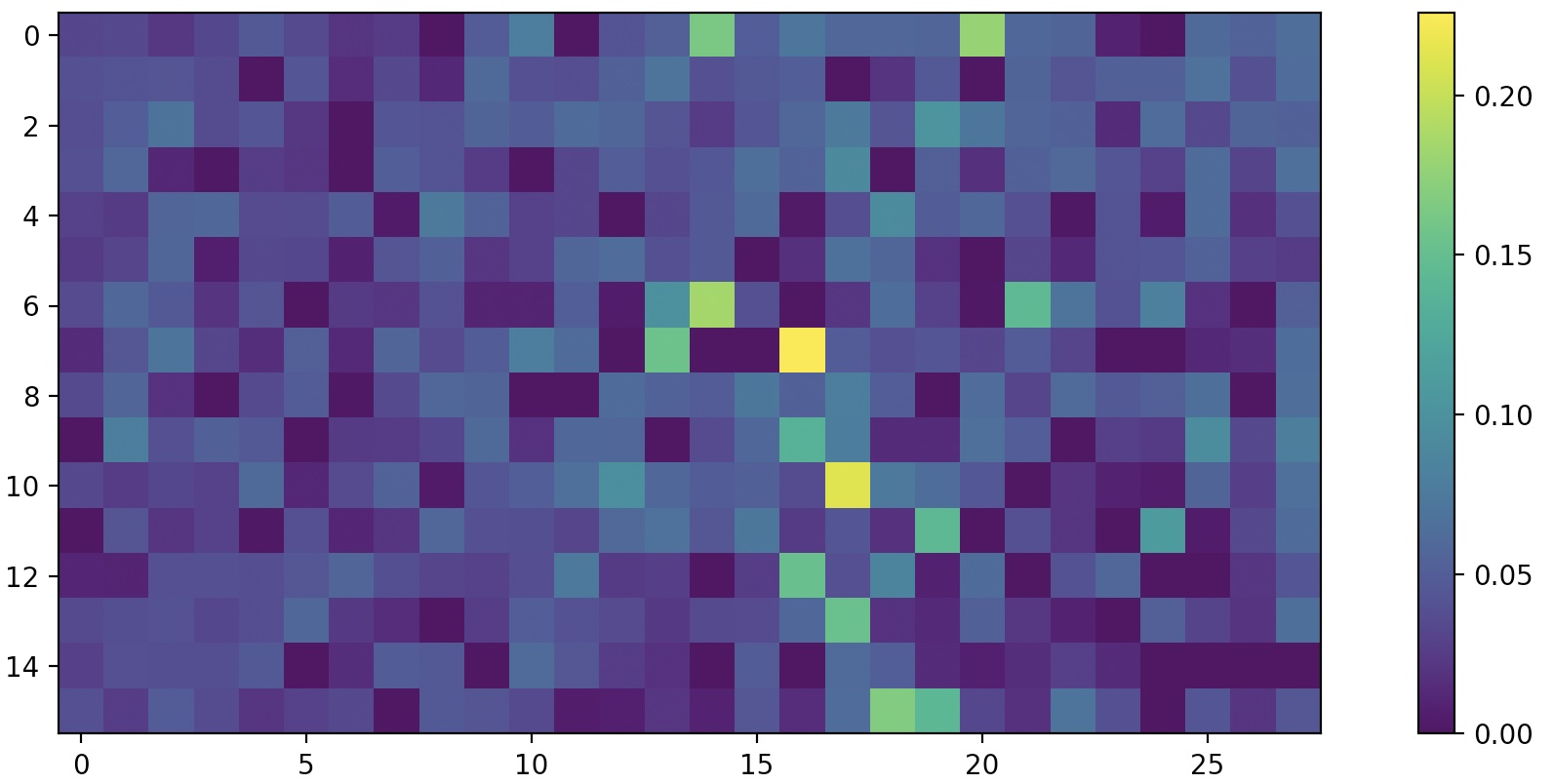}
         \caption{Pearson correlation between attention and logit similarity.}
     \end{subfigure}
\caption{Interpreting attention values from kernerl regression perspective on Rotten Tomatoes dataset.}
\end{figure}

\begin{figure}
     \centering
     \begin{subfigure}[b]{0.47\textwidth}
         \centering
         \includegraphics[width=\textwidth]{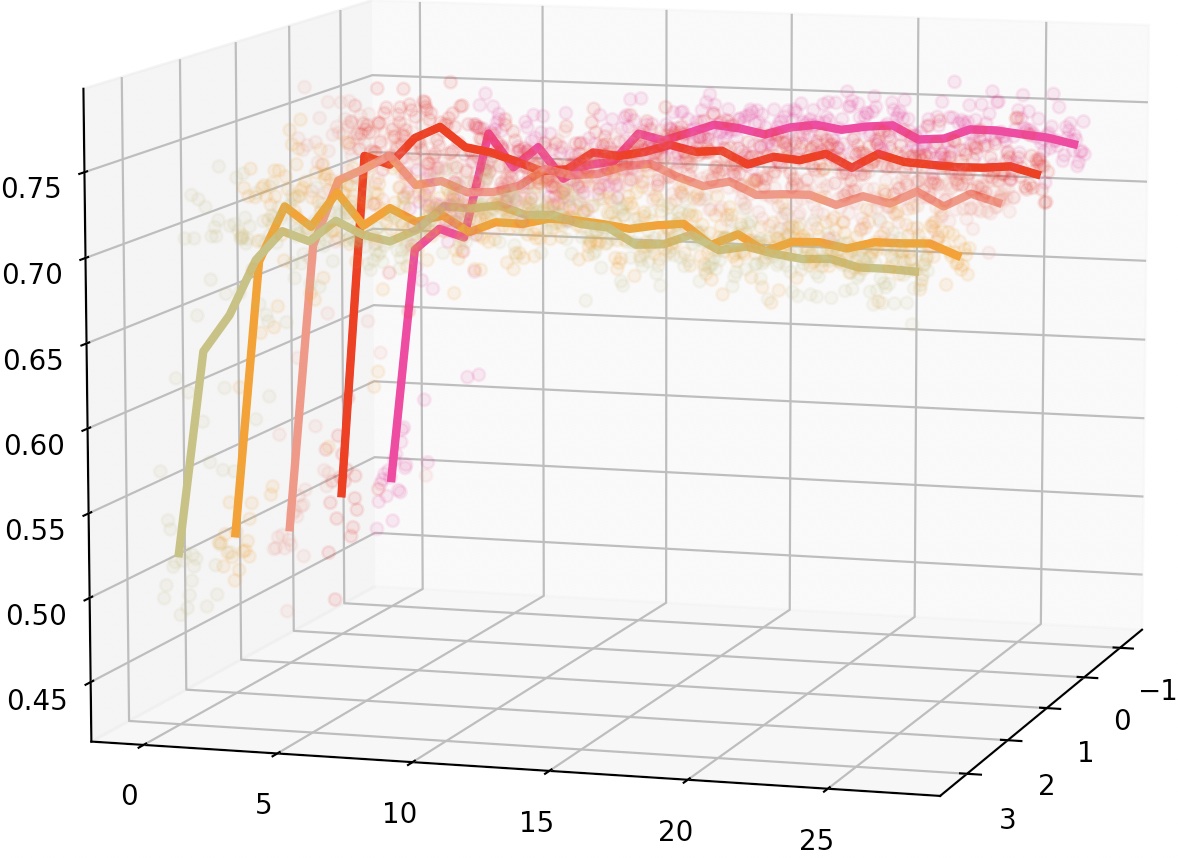}
         \caption{Predicting $\arg\max_o P(o|\bfx_i)$ with \textit{key} vectors.}
     \end{subfigure}
     \hfill
     \begin{subfigure}[b]{0.47\textwidth}
         \centering
         \includegraphics[width=\textwidth]{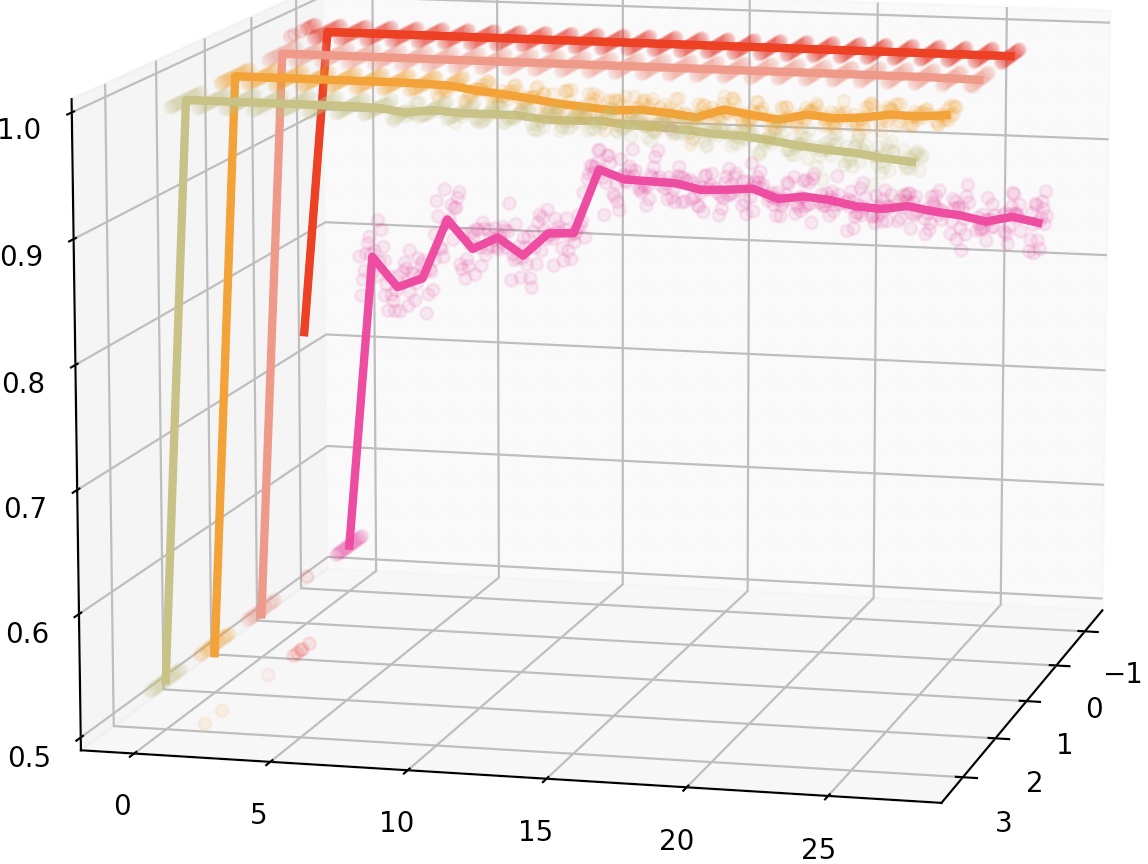}
         \caption{Predicting $y_i$ with \textit{value} vectors.}
     \end{subfigure}
\caption{Investigating information in key and value vectors on Rotten Tomatoes dataset.}
\end{figure}


\newpage

\subsection{Tweet Eval (Hate)}

\begin{figure*}[ht]
\centering
\includegraphics[width=\textwidth]{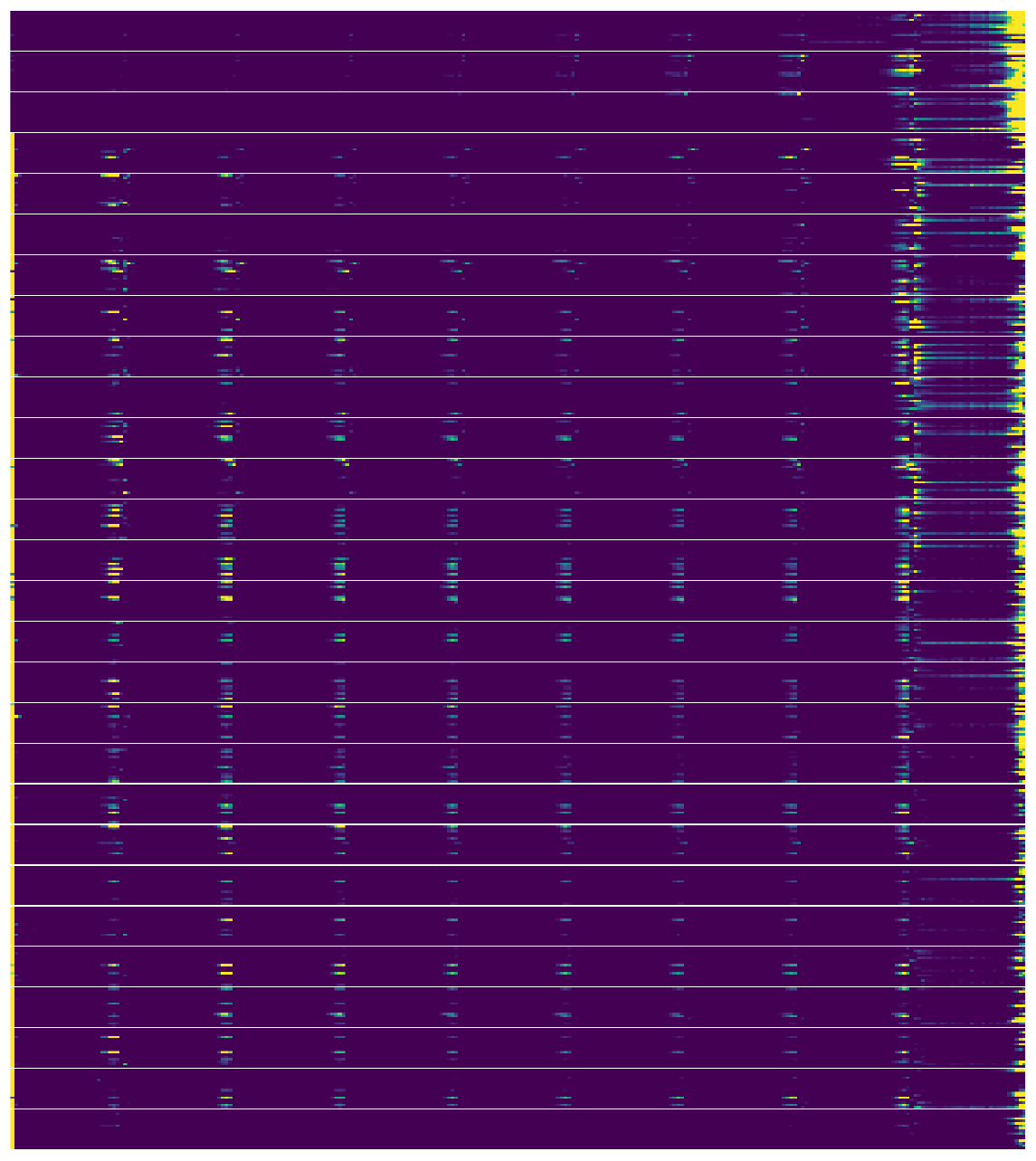}
\caption{Averaged attention map over Tweet Eval (Hate) test set.}
\end{figure*}

\begin{figure}
     \centering
     \begin{subfigure}[b]{0.47\textwidth}
         \centering
         \includegraphics[width=\textwidth]{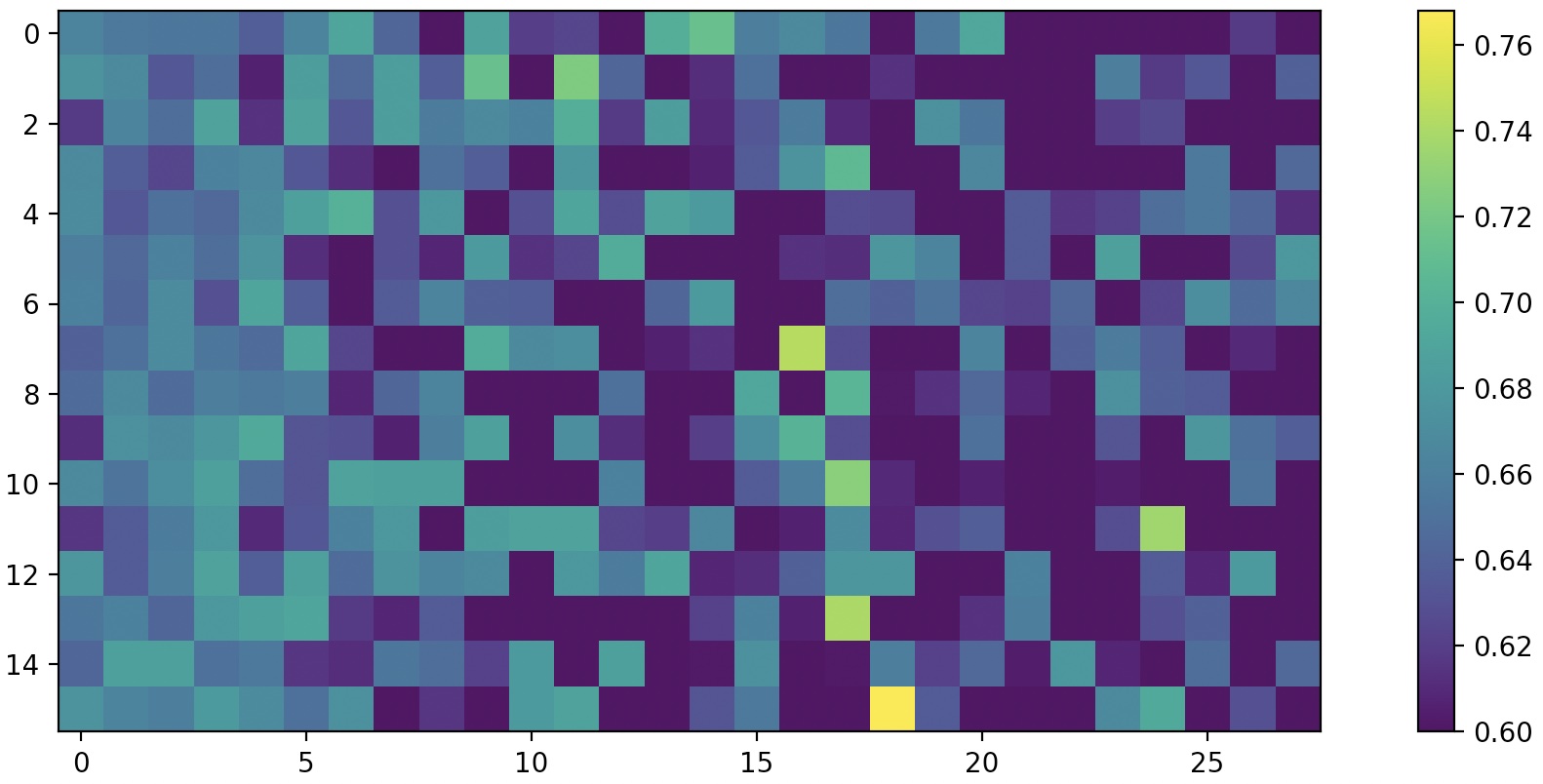}
         \caption{Accuracy on reconstruction of $\hat y$ by interpreting attention as kernel weights.}
     \end{subfigure}
     \hfill
     \begin{subfigure}[b]{0.47\textwidth}
         \centering
         \includegraphics[width=\textwidth]{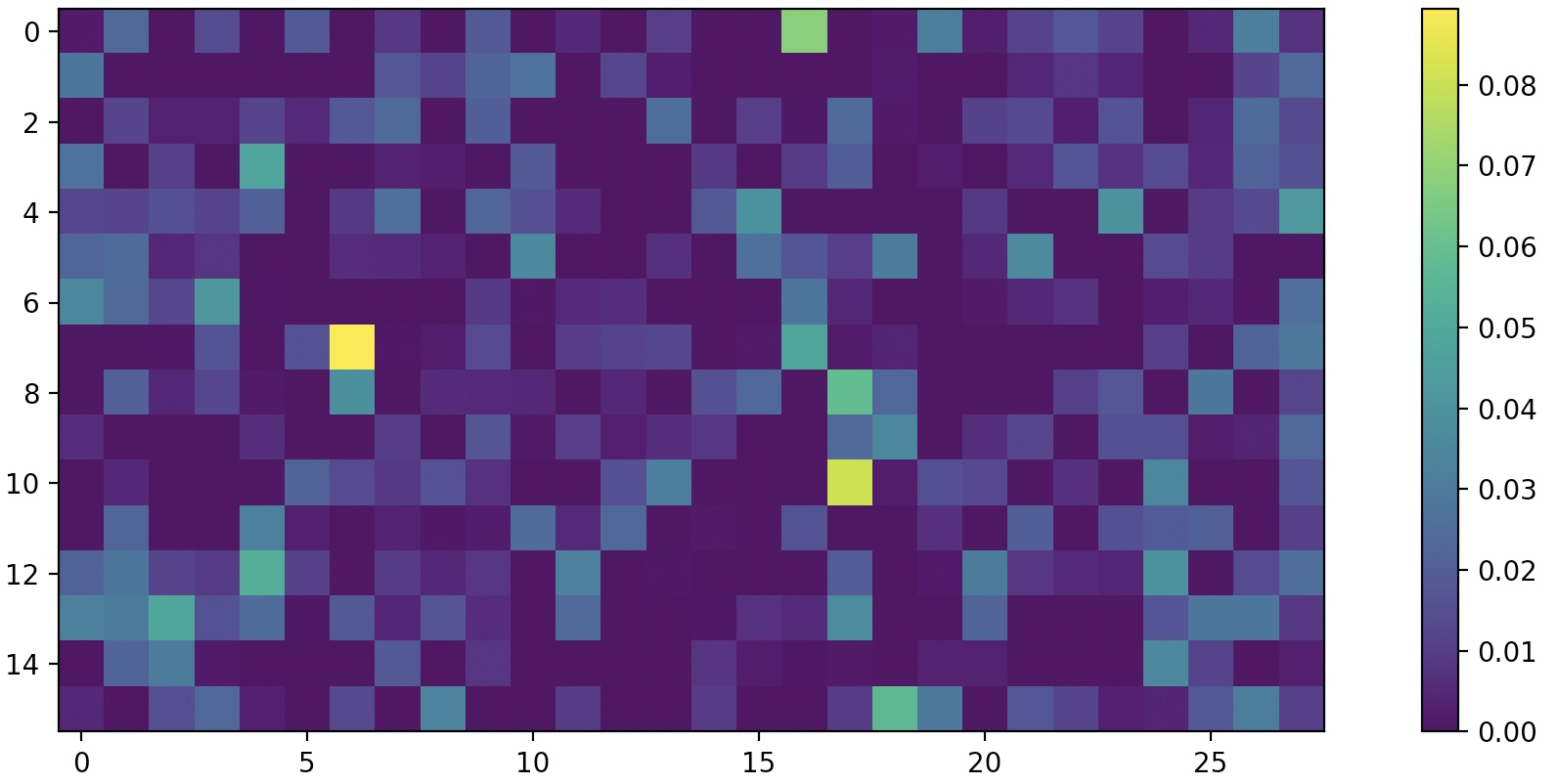}
         \caption{Pearson correlation between attention and logit similarity.}
     \end{subfigure}
\caption{Interpreting attention values from kernerl regression perspective on Tweet Eval (Hate) dataset.}
\end{figure}

\begin{figure}
     \centering
     \begin{subfigure}[b]{0.47\textwidth}
         \centering
         \includegraphics[width=\textwidth]{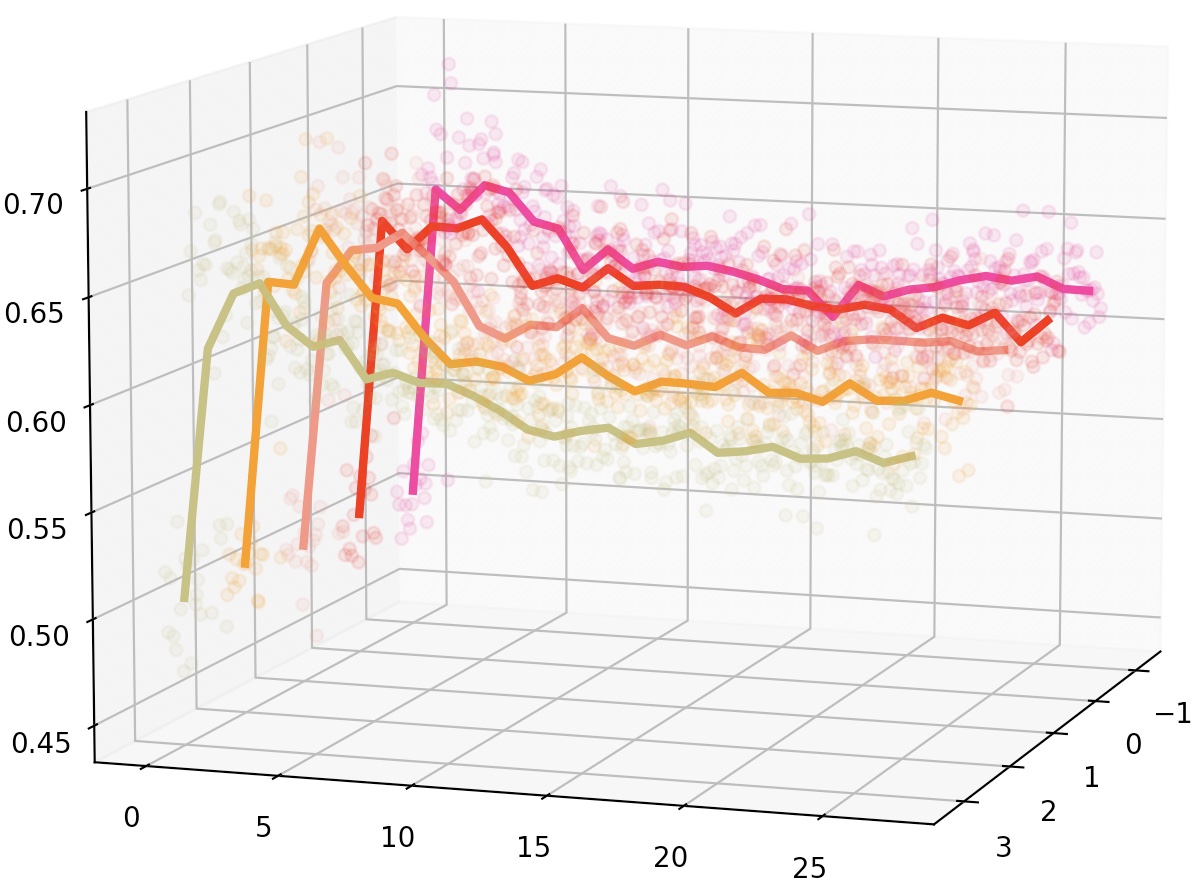}
         \caption{Predicting $\arg\max_o P(o|\bfx_i)$ with \textit{key} vectors.}
     \end{subfigure}
     \hfill
     \begin{subfigure}[b]{0.47\textwidth}
         \centering
         \includegraphics[width=\textwidth]{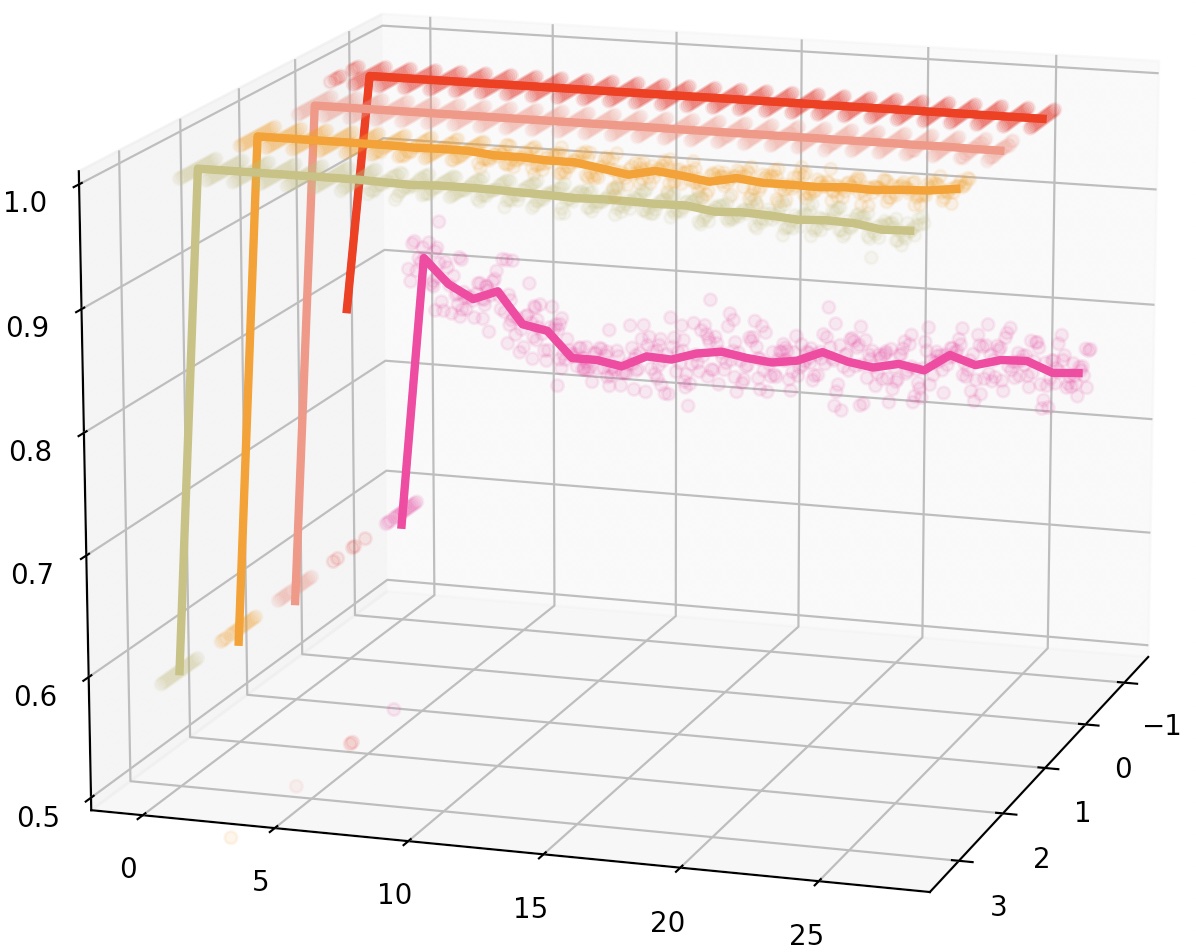}
         \caption{Predicting $y_i$ with \textit{value} vectors.}
     \end{subfigure}
\caption{Investigating information in key and value vectors on Tweet Eval (Hate) dataset.}
\end{figure}


\newpage

\subsection{Tweet Eval (Irony)}

\begin{figure*}[ht]
\centering
\includegraphics[width=\textwidth]{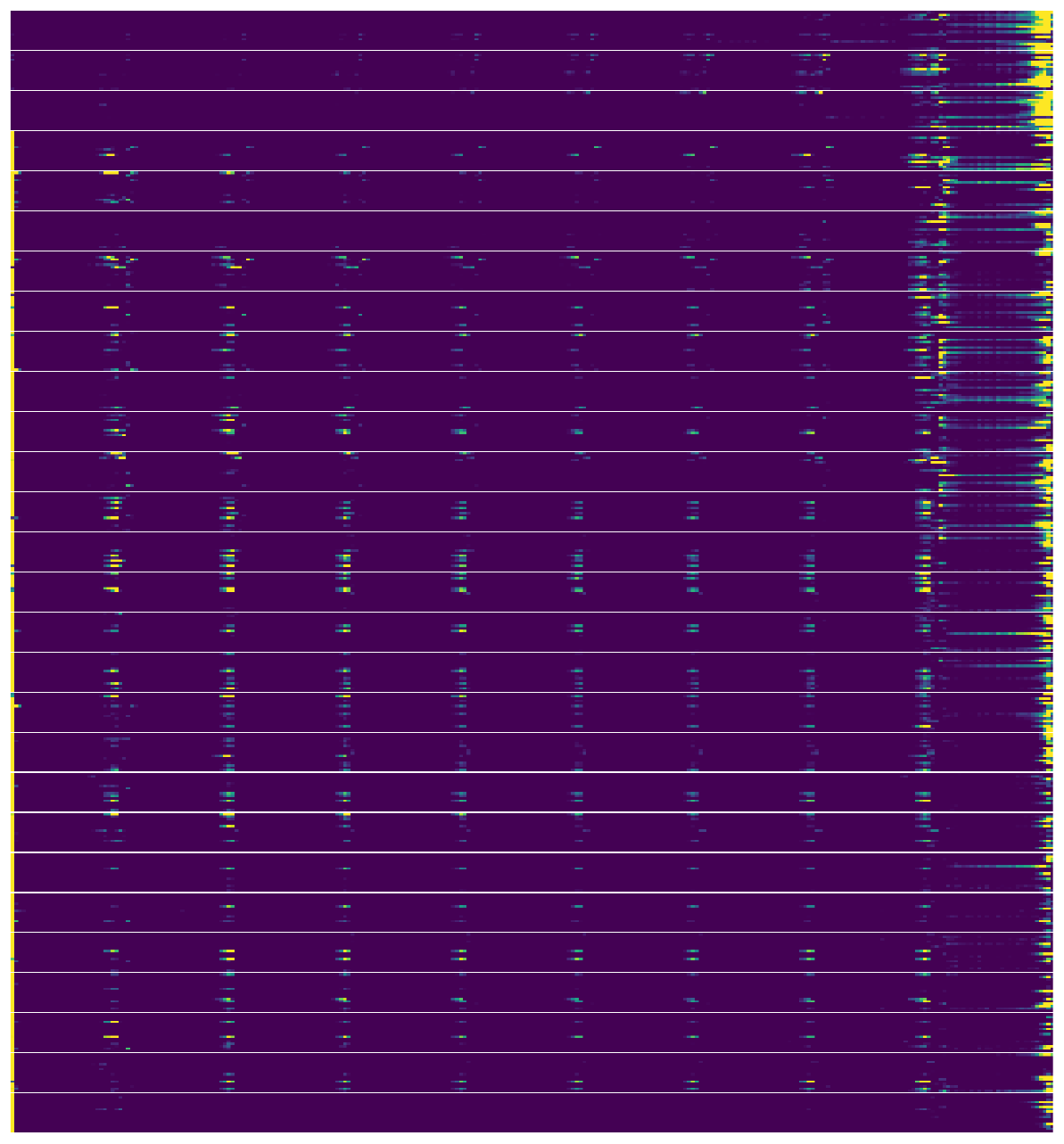}
\caption{Averaged attention map over Tweet Eval (Irony) test set.}
\end{figure*}

\begin{figure}
     \centering
     \begin{subfigure}[b]{0.47\textwidth}
         \centering
         \includegraphics[width=\textwidth]{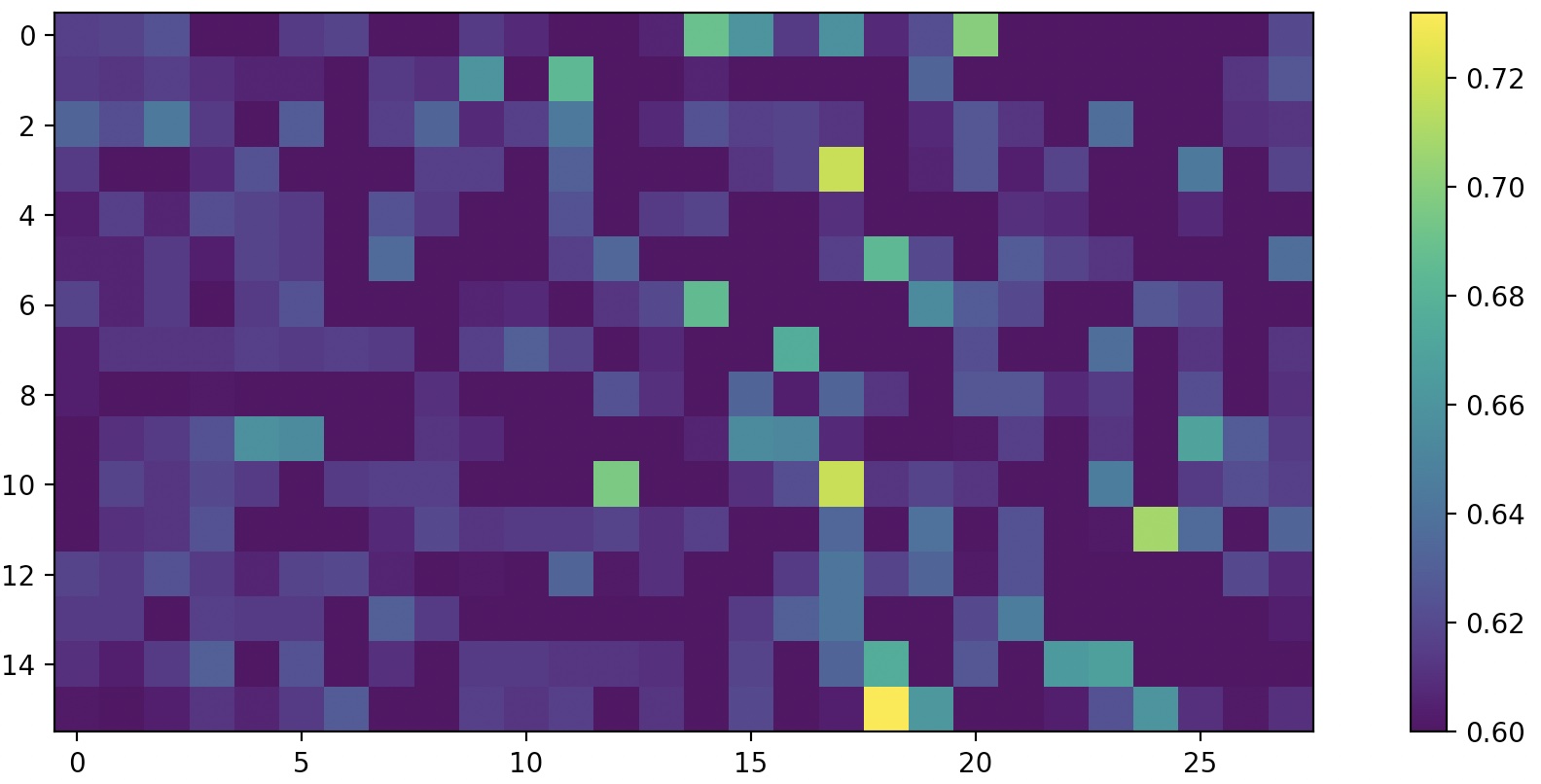}
         \caption{Accuracy on reconstruction of $\hat y$ by interpreting attention as kernel weights.}
     \end{subfigure}
     \hfill
     \begin{subfigure}[b]{0.47\textwidth}
         \centering
         \includegraphics[width=\textwidth]{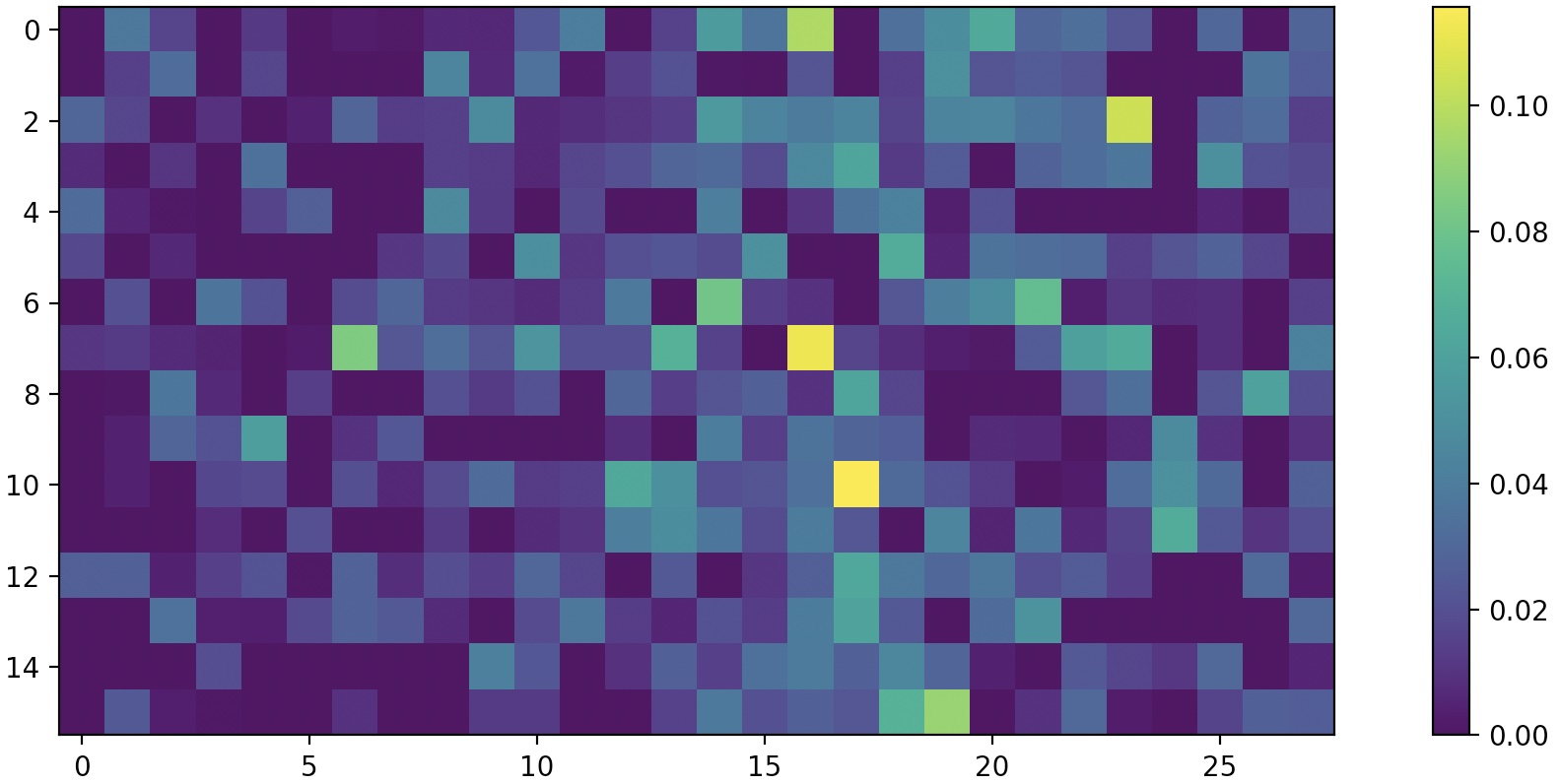}
         \caption{Pearson correlation between attention and logit similarity.}
     \end{subfigure}
\caption{Interpreting attention values from kernerl regression perspective on Tweet Eval (Irony) dataset.}
\end{figure}

\begin{figure}
     \centering
     \begin{subfigure}[b]{0.47\textwidth}
         \centering
         \includegraphics[width=\textwidth]{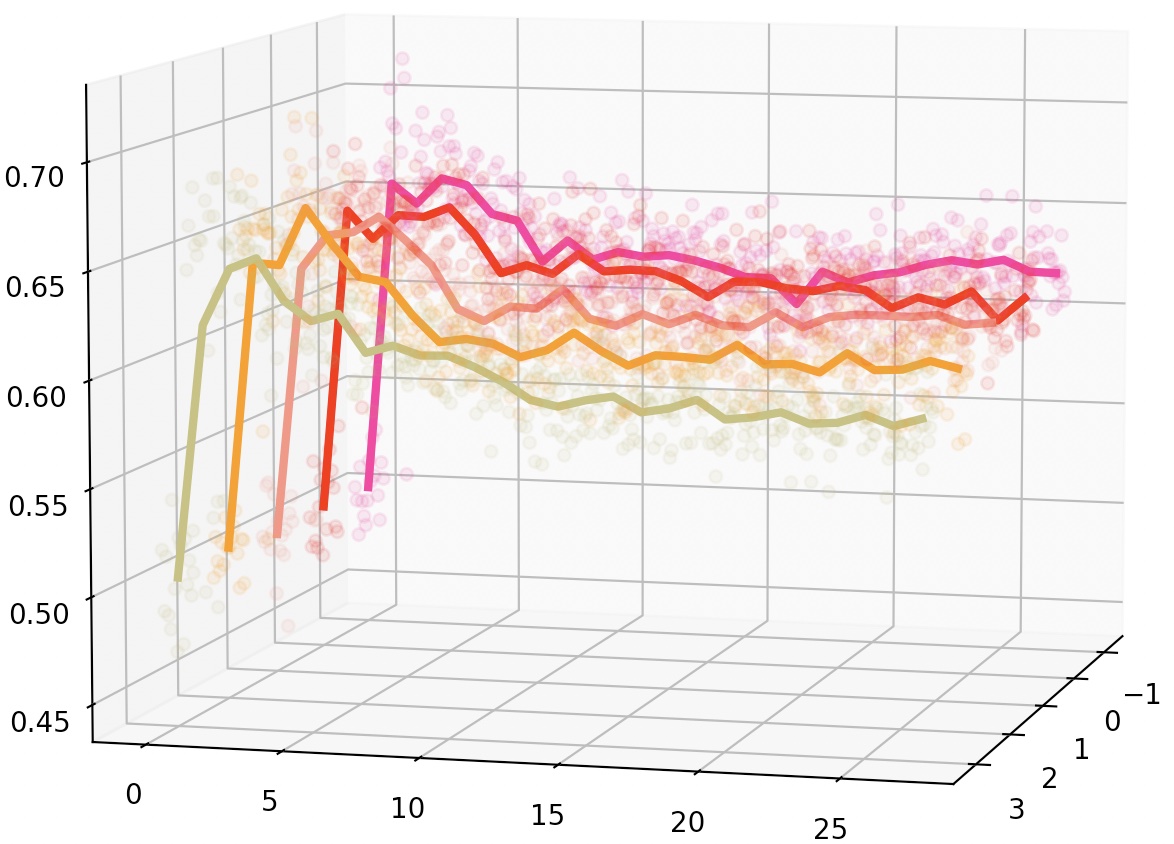}
         \caption{Predicting $\arg\max_o P(o|\bfx_i)$ with \textit{key} vectors.}
     \end{subfigure}
     \hfill
     \begin{subfigure}[b]{0.47\textwidth}
         \centering
         \includegraphics[width=\textwidth]{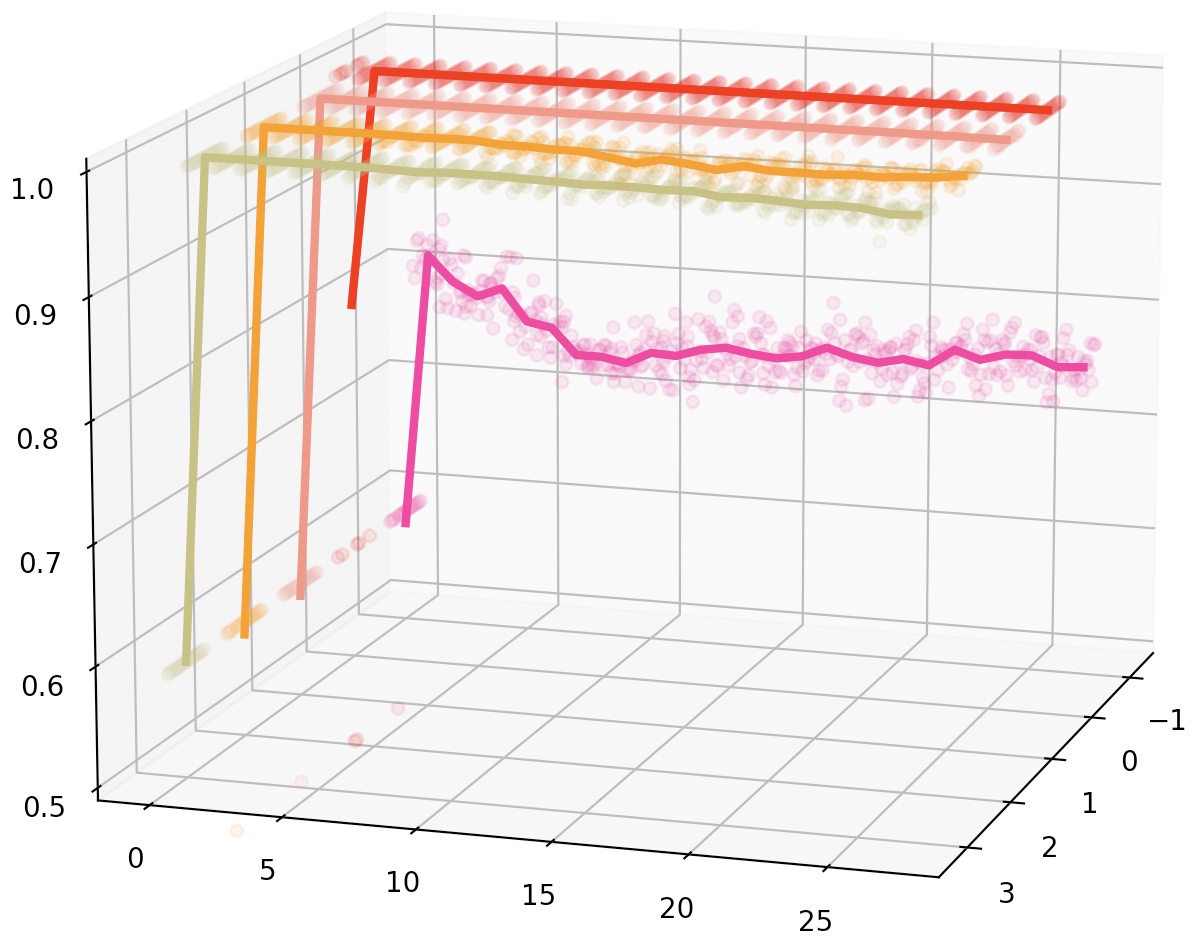}
         \caption{Predicting $y_i$ with \textit{value} vectors.}
     \end{subfigure}
\caption{Investigating information in key and value vectors on Tweet Eval (Irony) dataset.}
\end{figure}


\newpage

\subsection{Tweet Eval (Offensive)}

\begin{figure*}[ht]
\centering
\includegraphics[width=\textwidth]{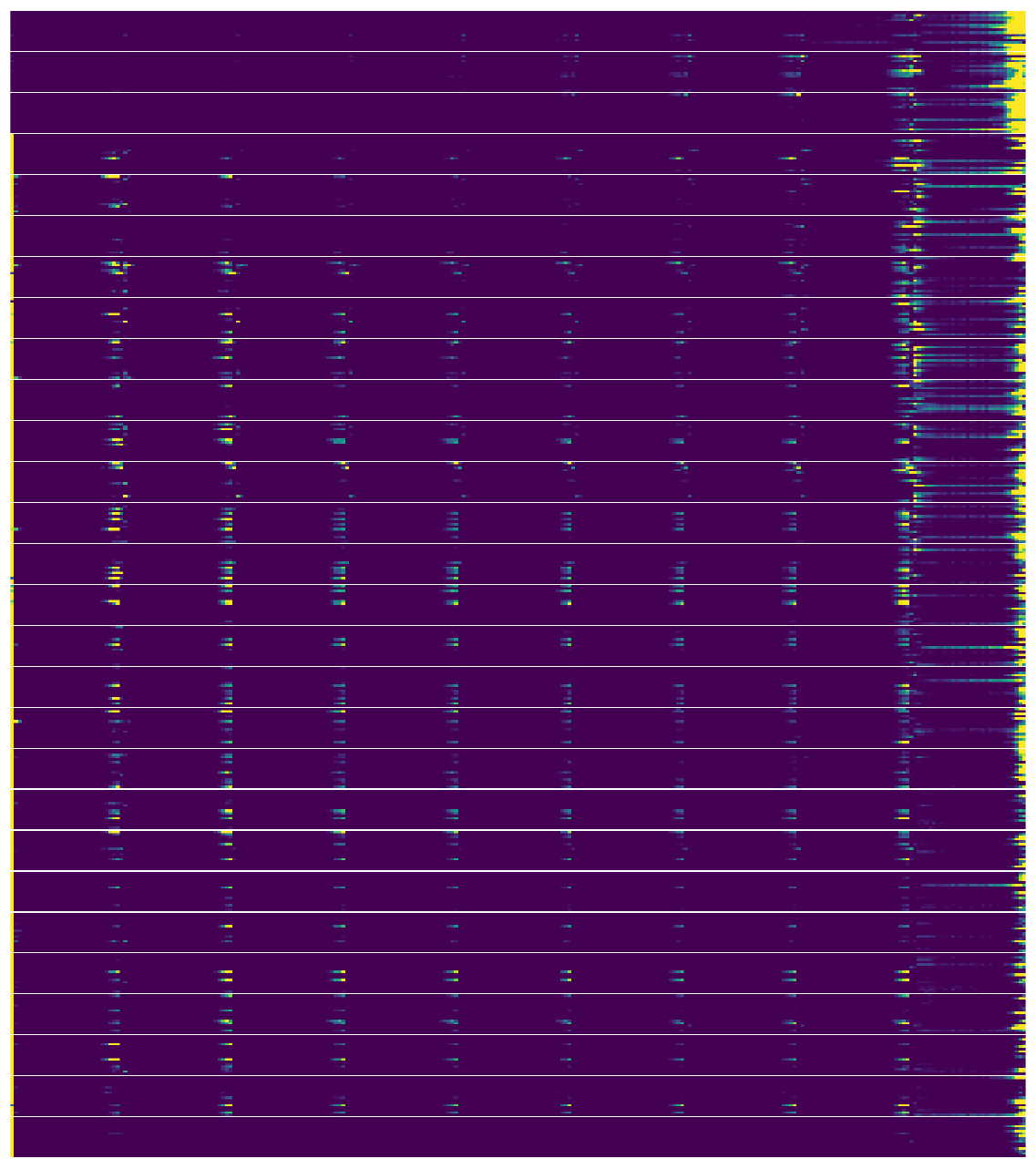}
\caption{Averaged attention map over Tweet Eval (Offensive) test set.}
\end{figure*}

\begin{figure}
     \centering
     \begin{subfigure}[b]{0.47\textwidth}
         \centering
         \includegraphics[width=\textwidth]{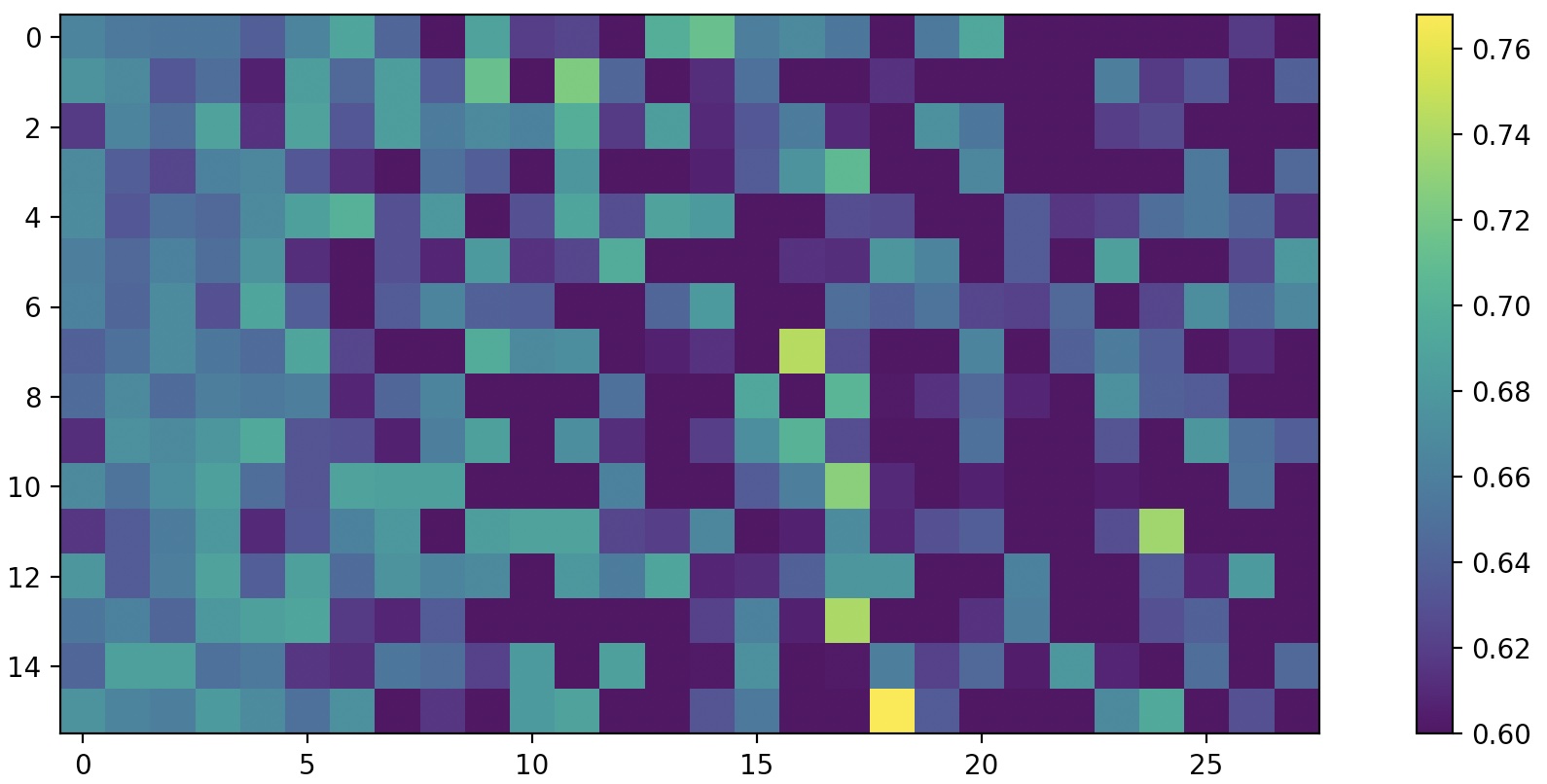}
         \caption{Accuracy on reconstruction of $\hat y$ by interpreting attention as kernel weights.}
     \end{subfigure}
     \hfill
     \begin{subfigure}[b]{0.47\textwidth}
         \centering
         \includegraphics[width=\textwidth]{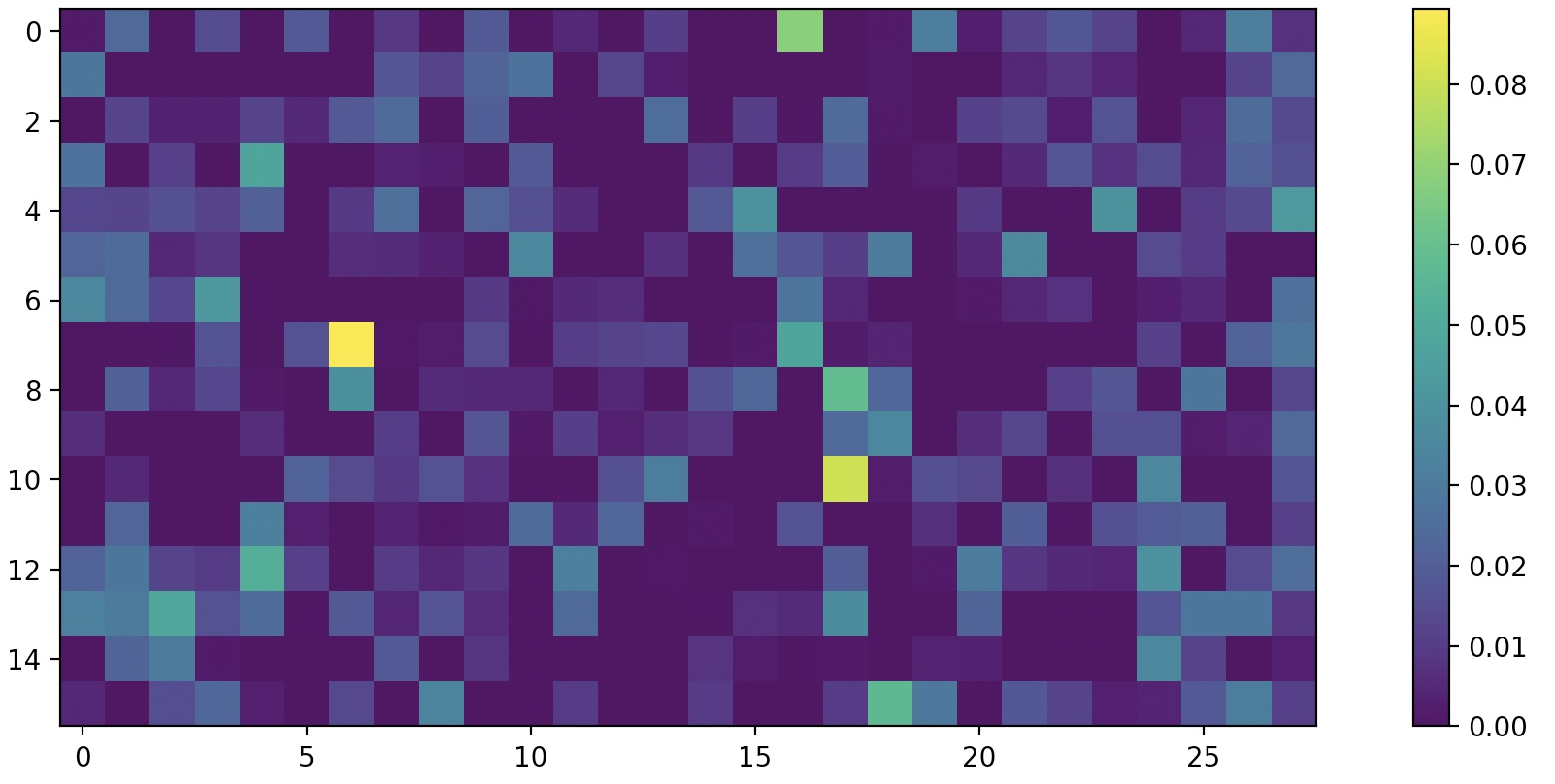}
         \caption{Pearson correlation between attention and logit similarity.}
     \end{subfigure}
\caption{Interpreting attention values from kernerl regression perspective on Tweet Eval (Offensive) dataset.}
\end{figure}

\begin{figure}
     \centering
     \begin{subfigure}[b]{0.47\textwidth}
         \centering
         \includegraphics[width=\textwidth]{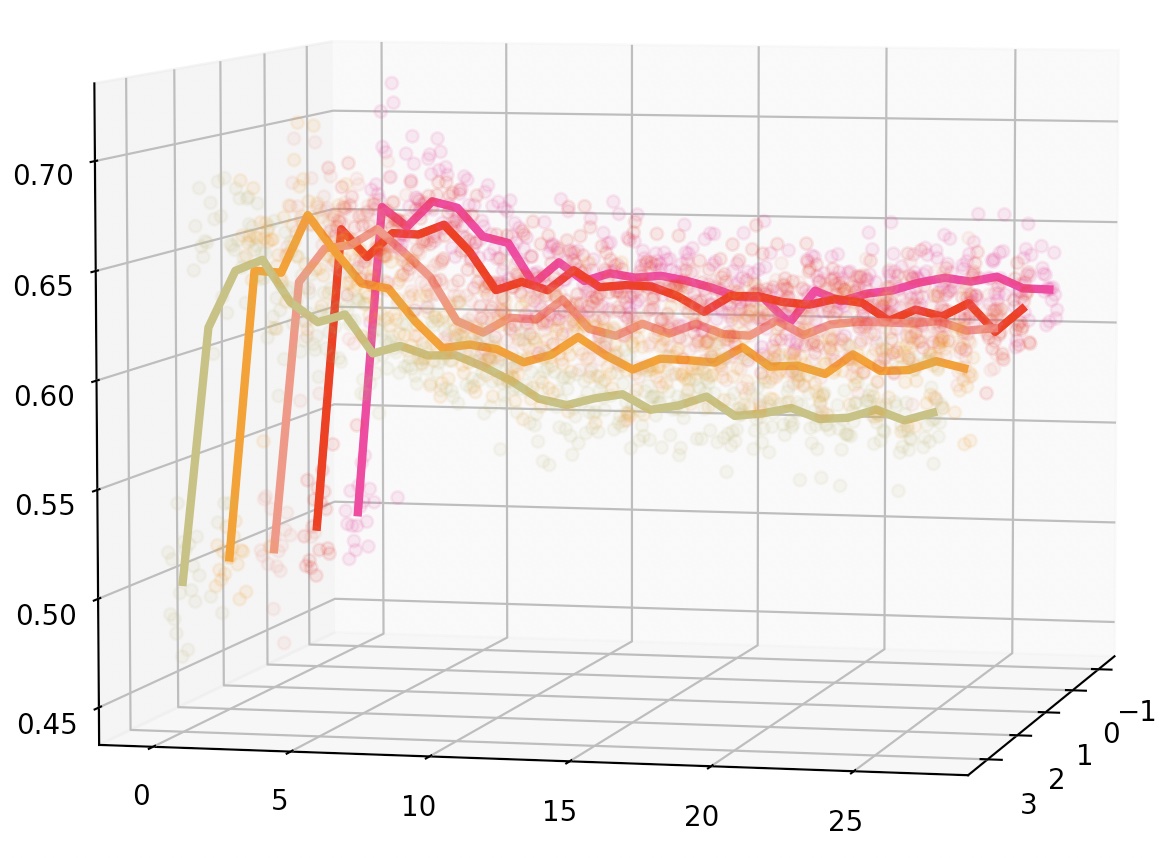}
         \caption{Predicting $\arg\max_o P(o|\bfx_i)$ with \textit{key} vectors.}
     \end{subfigure}
     \hfill
     \begin{subfigure}[b]{0.47\textwidth}
         \centering
         \includegraphics[width=\textwidth]{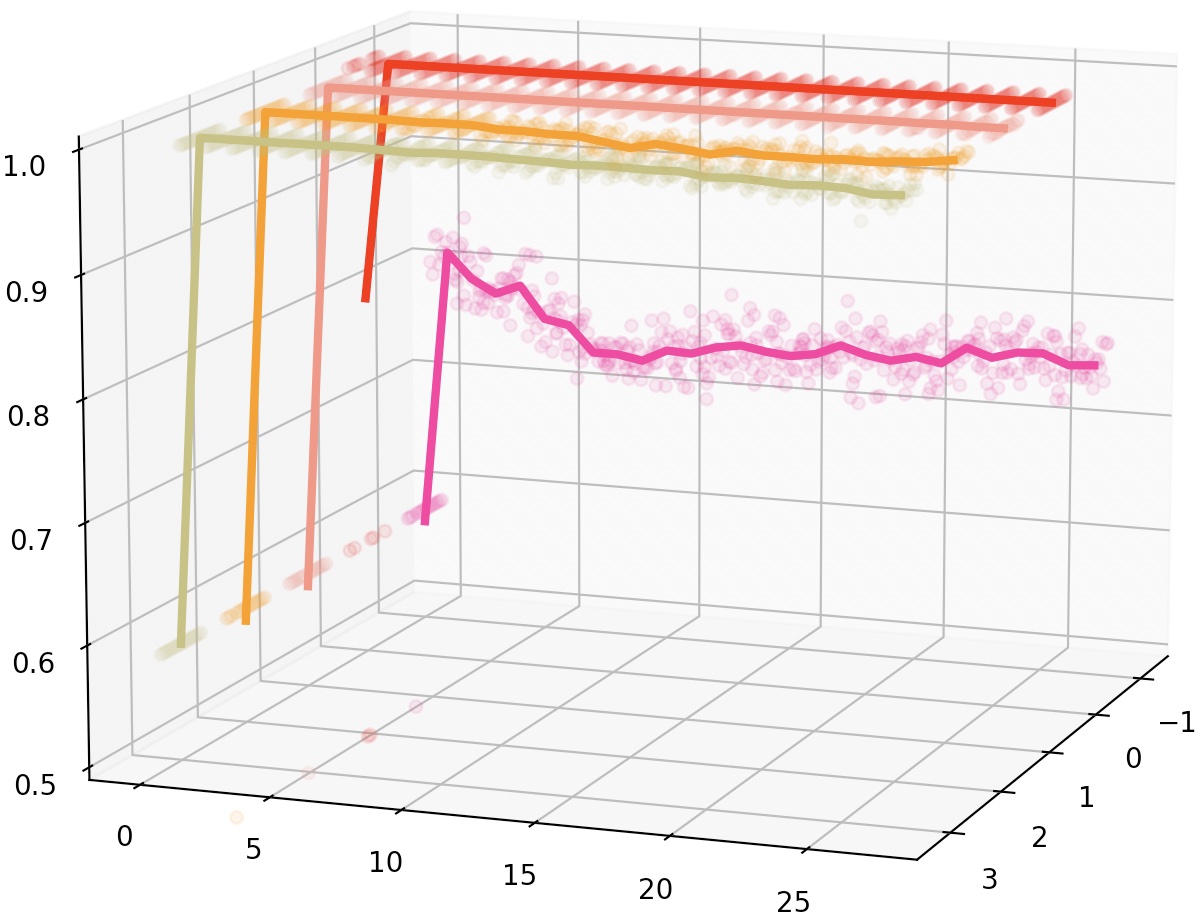}
         \caption{Predicting $y_i$ with \textit{value} vectors.}
     \end{subfigure}
\caption{Investigating information in key and value vectors on Tweet Eval (Offensive) dataset.}
\end{figure}


\newpage

\subsection{MNLI}

\begin{figure*}[ht]
\centering
\includegraphics[width=\textwidth]{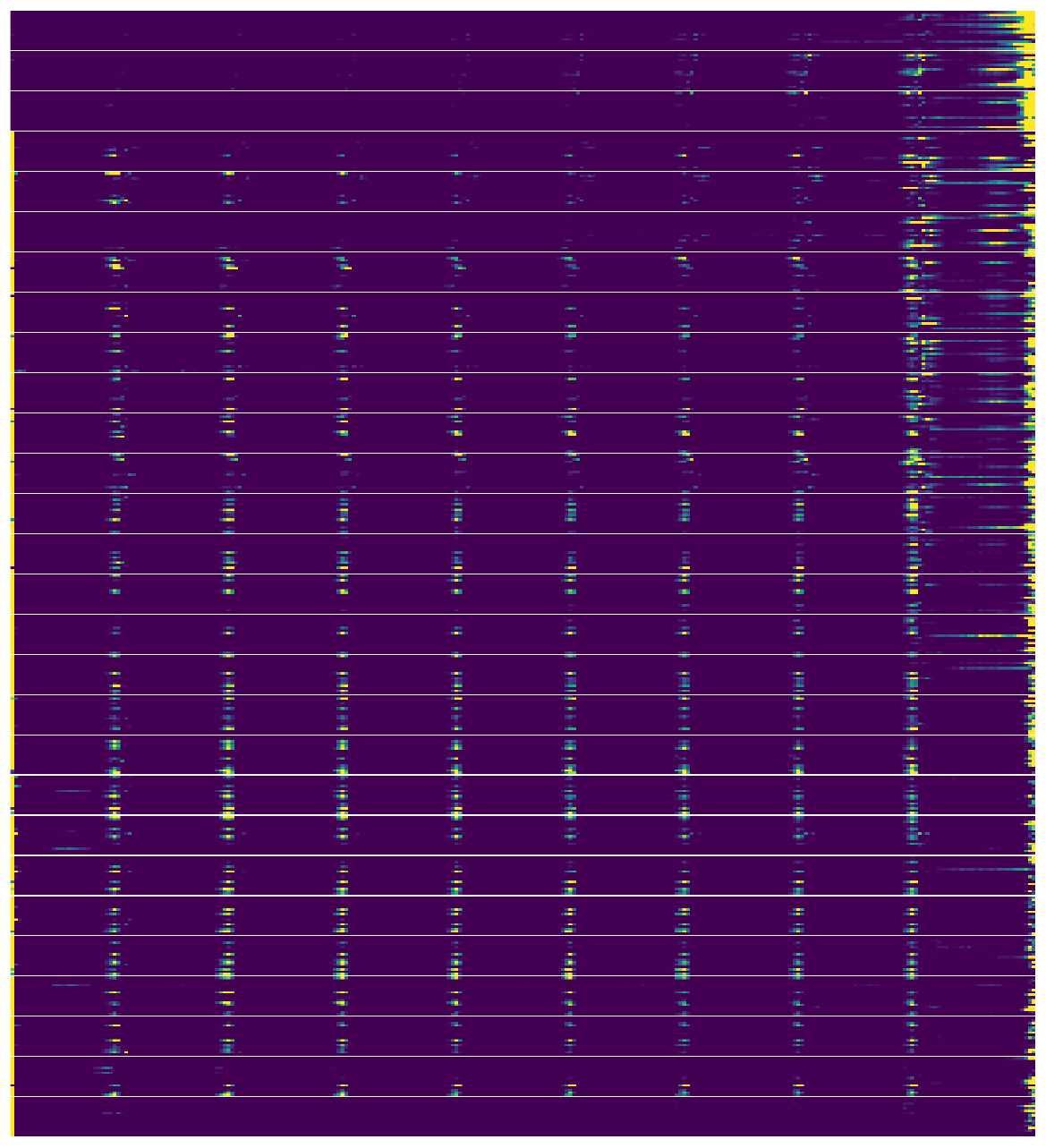}
\caption{Averaged attention map over MNLI test set.}
\end{figure*}

\begin{figure}
     \centering
     \begin{subfigure}[b]{0.47\textwidth}
         \centering
         \includegraphics[width=\textwidth]{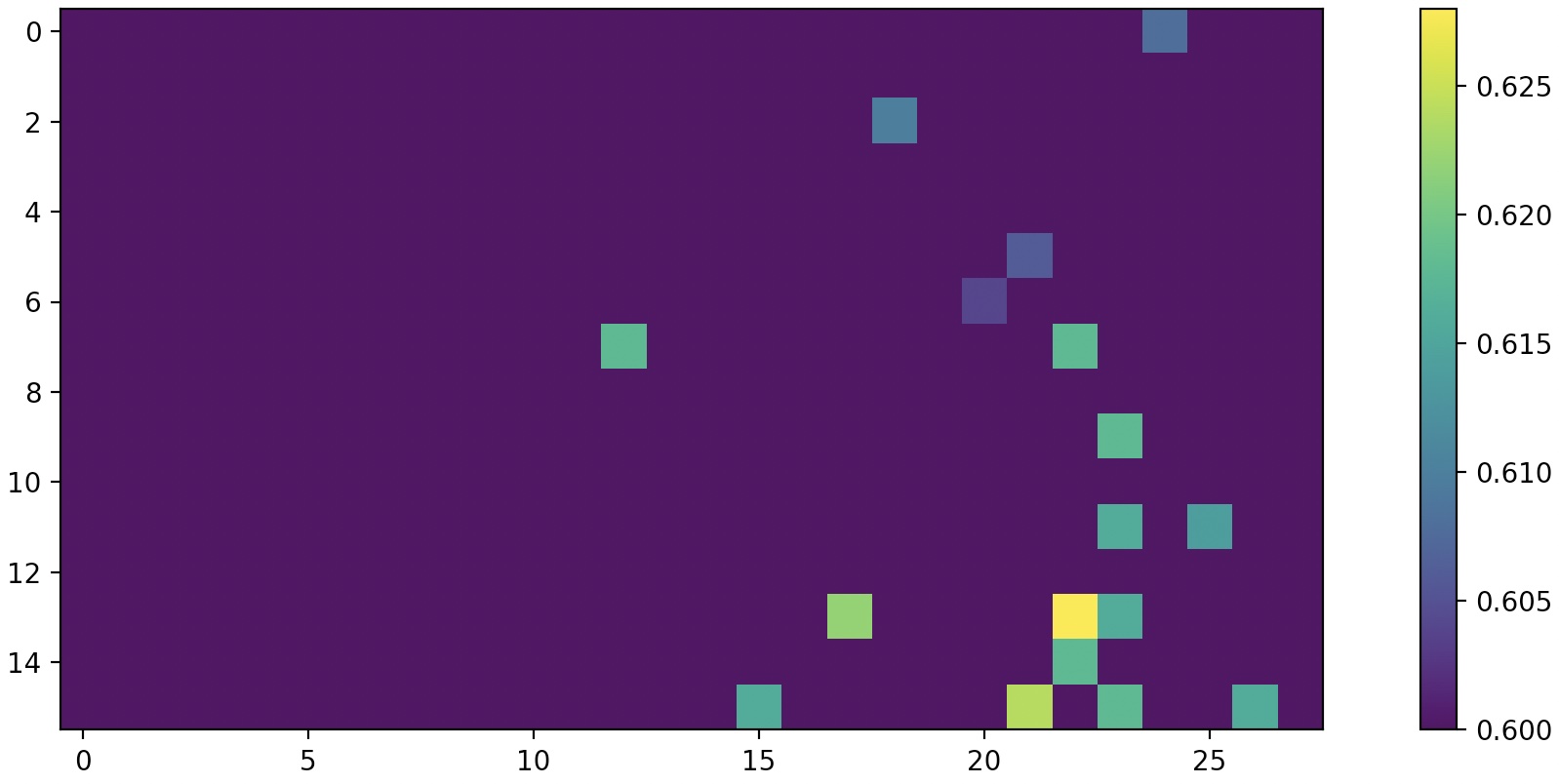}
         \caption{Accuracy on reconstruction of $\hat y$ by interpreting attention as kernel weights.}
     \end{subfigure}
     \hfill
     \begin{subfigure}[b]{0.47\textwidth}
         \centering
         \includegraphics[width=\textwidth]{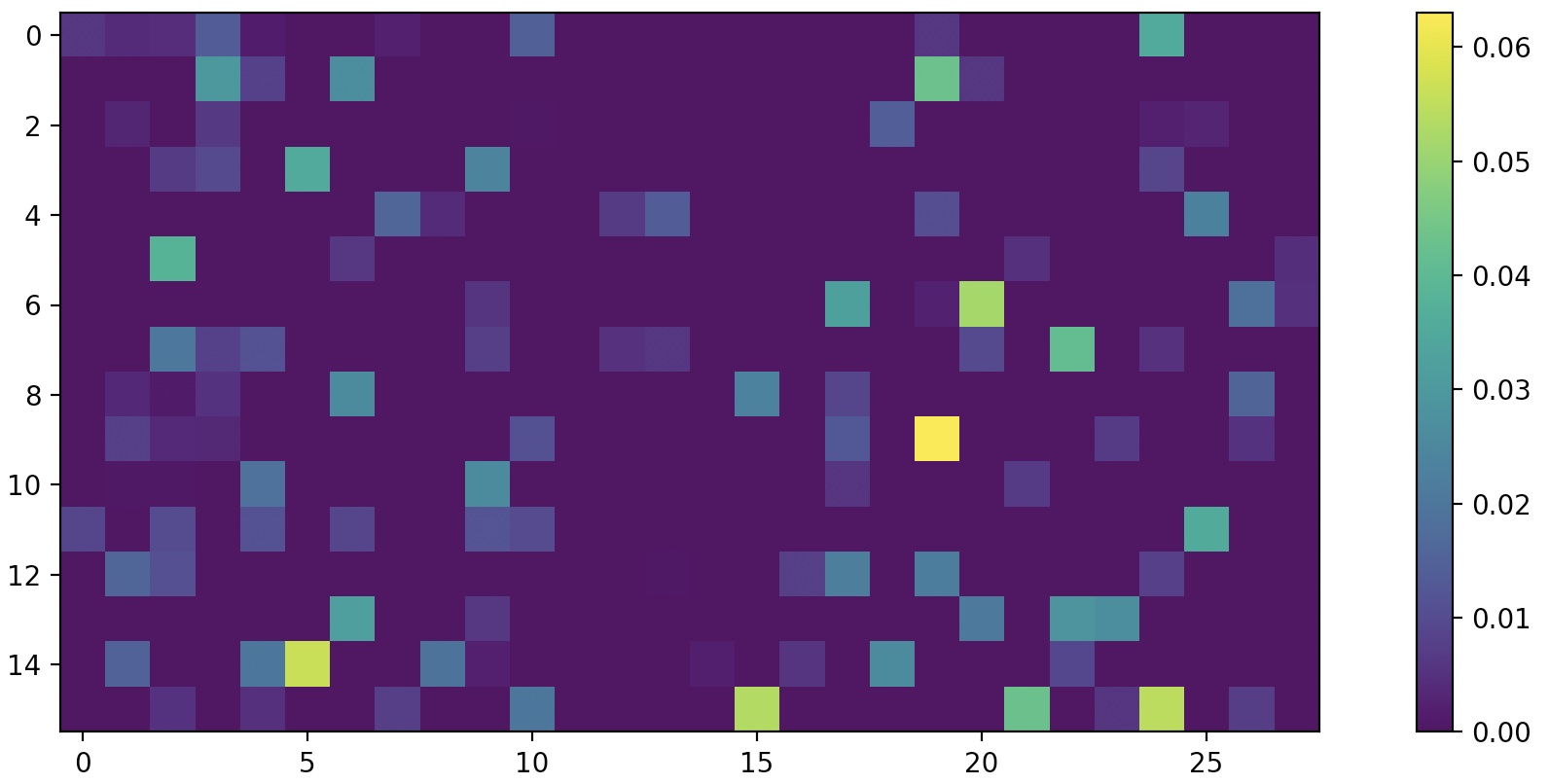}
         \caption{Pearson correlation between attention and logit similarity.}
     \end{subfigure}
\caption{Interpreting attention values from kernerl regression perspective on MNLI dataset.}
\end{figure}

\begin{figure}
     \centering
     \begin{subfigure}[b]{0.47\textwidth}
         \centering
         \includegraphics[width=\textwidth]{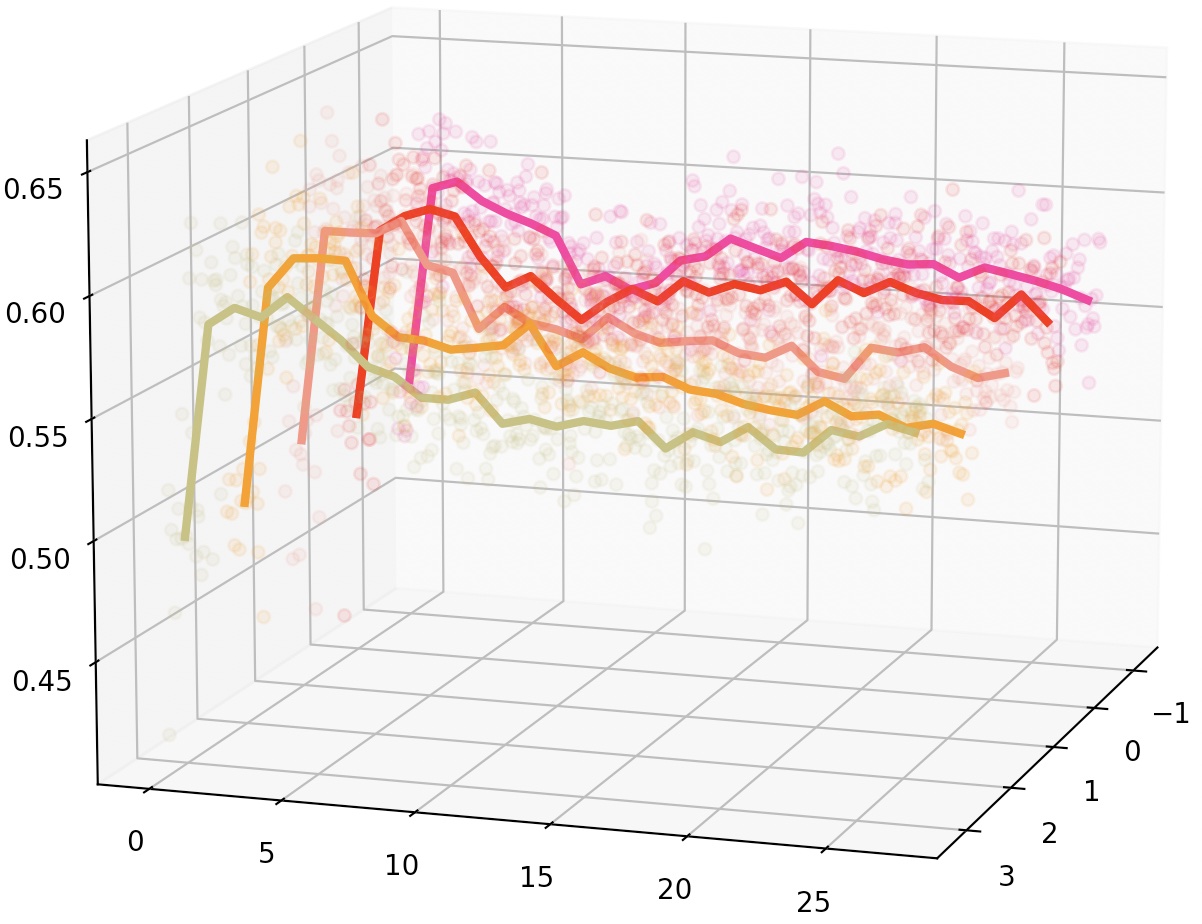}
         \caption{Predicting $\arg\max_o P(o|\bfx_i)$ with \textit{key} vectors.}
     \end{subfigure}
     \hfill
     \begin{subfigure}[b]{0.47\textwidth}
         \centering
         \includegraphics[width=\textwidth]{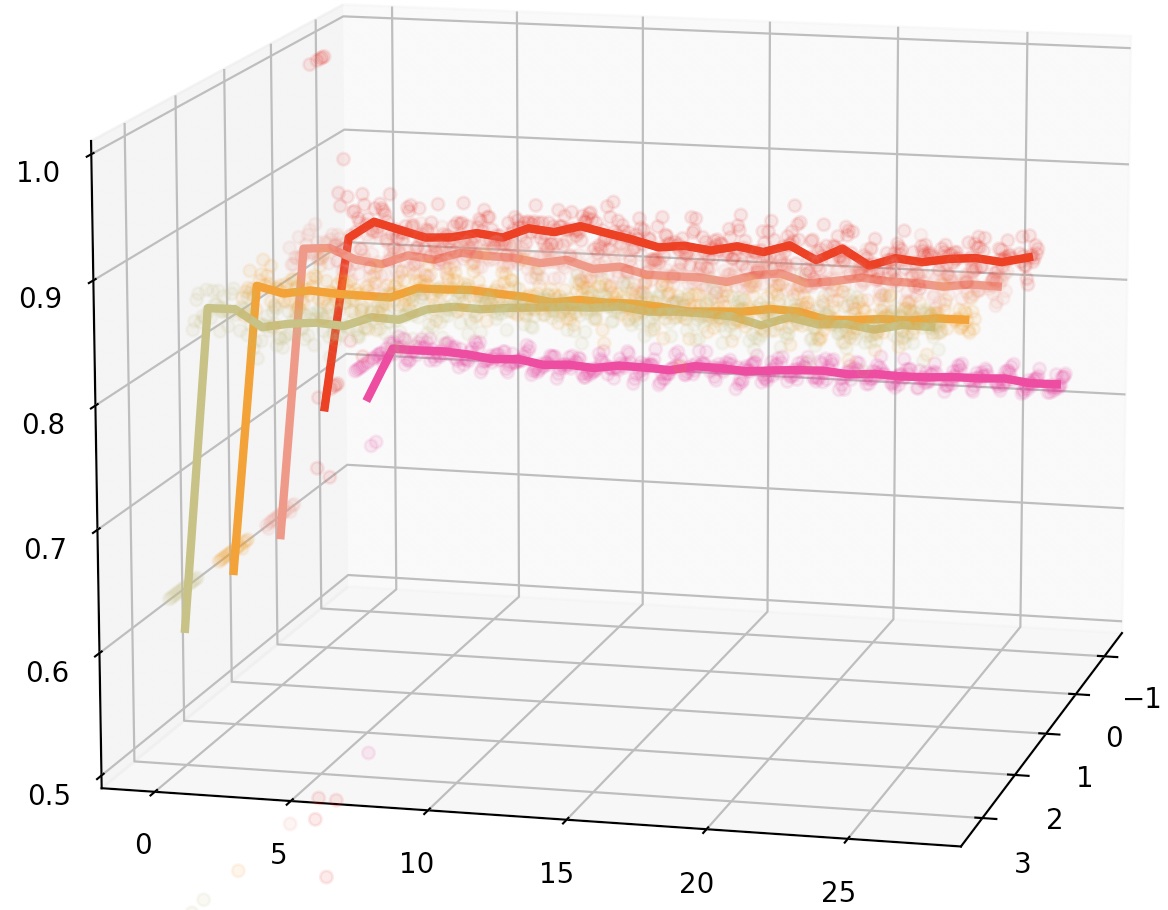}
         \caption{Predicting $y_i$ with \textit{value} vectors.}
     \end{subfigure}
\caption{Investigating information in key and value vectors on MNLI dataset.}
\end{figure}

\newpage
\section{Synthetic Verification}
\label{appsec:synthetic}
To verify the theoretical results, we conduct experiments on a synthetic HMM.
We experimented on randomly parameterized synthetic HMMs with 8 tasks, 80 states, and 100 observations. We vary the number of samples and evaluate the proposed kernel regression on fitting the Bayesian posterior, which is the intuition in Theorem~\ref{thm:kernel}. Results are listed in the Table~\ref{tab:synthetic}. We indeed see a decreasing loss and increasing accuracy with more demonstrative examples. The loss also converges to a non-zero value according to Equation~\ref{eq:convergence}.

\begin{table}[ht]
\centering
\caption{Accuracy and distance of predicting the Bayesian posterior on a synthetic HMM with Equation~\ref{eq:construction}.}
\begin{tabular}{c|cccccccc}
\toprule
\#samples & 1 & 2 & 4 & 8 & 16 & 32 & 64 & 128 \\
\midrule
distance & 1.322 & 0.942 & 0.519 & 0.218 & 0.163 & 0.085 & 0.093 & 0.083 \\
accuracy & 0.337 & 0.532 & 0.745 & 0.901 & 0.926 & 0.968 & 0.961 & 0.964 \\
\bottomrule
\end{tabular}
\label{tab:synthetic}
\end{table}

\newpage
\section{Results on Llama-2 and Llama-3}
\label{appsec:llama2}
To demonstrate the generality of our conclusions across model architectures, we also conduct analysis on Llama-2 model~\citep{touvron2023llama} and Llama-3 8B model~\citep{grattafiori2024llama}. The experiment setting is similar to Section \ref{subsec:attention_distribution} and 
\ref{subsec:prediction_reconstruction}.  Visualizations are as follows. They generally show similar trends with Section~\ref{sec:empirical}.

\begin{figure*}[ht]
\centering
\includegraphics[width=\textwidth]{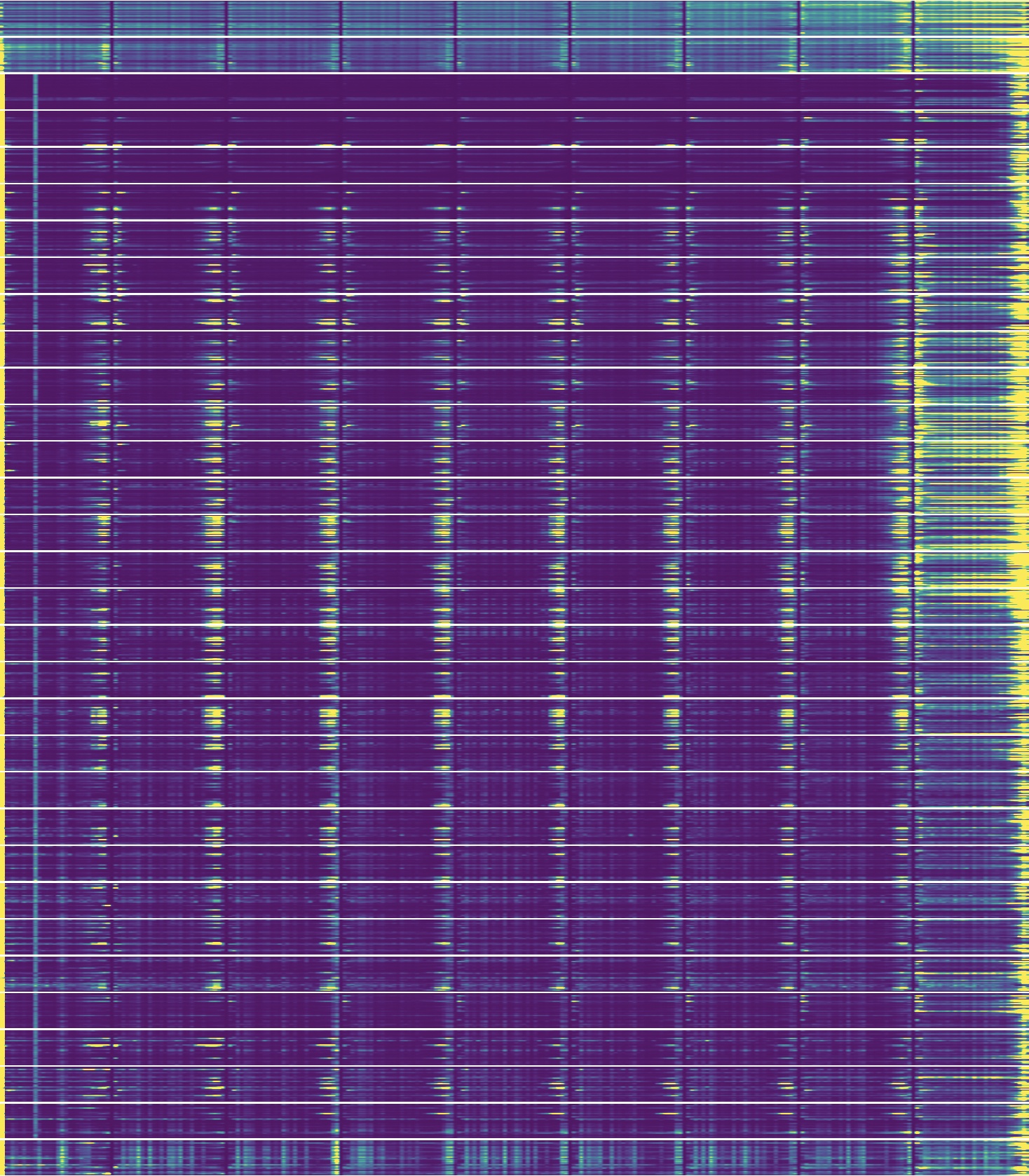}
\caption{Averaged attention map over SST2 dataset on Llama-2 model.}
\end{figure*}

\begin{figure}
     \centering
     \begin{subfigure}[b]{0.47\textwidth}
         \centering
         \includegraphics[width=\textwidth]{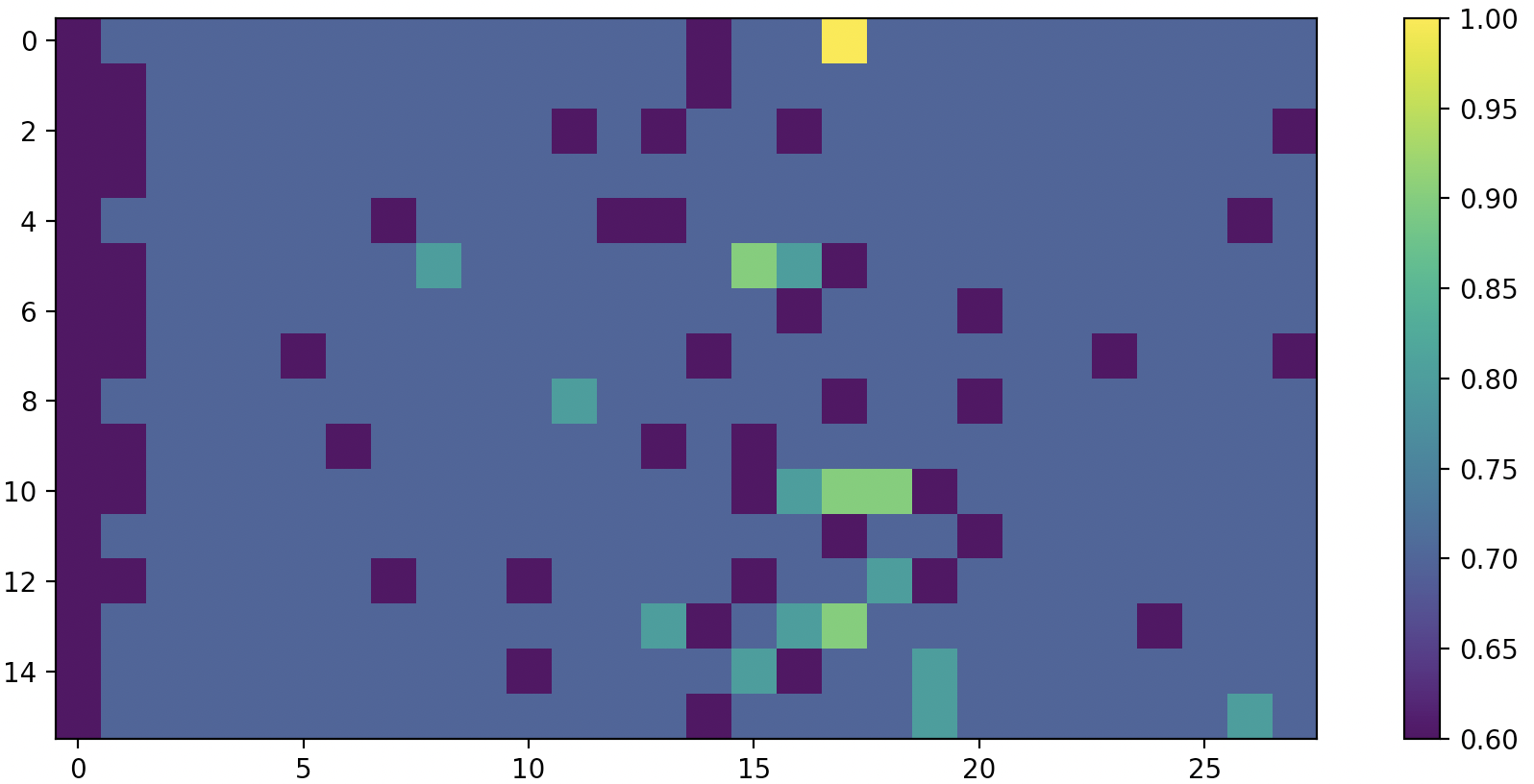}
         \caption{Accuracy on reconstruction of $\hat y$ by interpreting attention as kernel weights.}
     \end{subfigure}
     \hfill
     \begin{subfigure}[b]{0.47\textwidth}
         \centering
         \includegraphics[width=\textwidth]{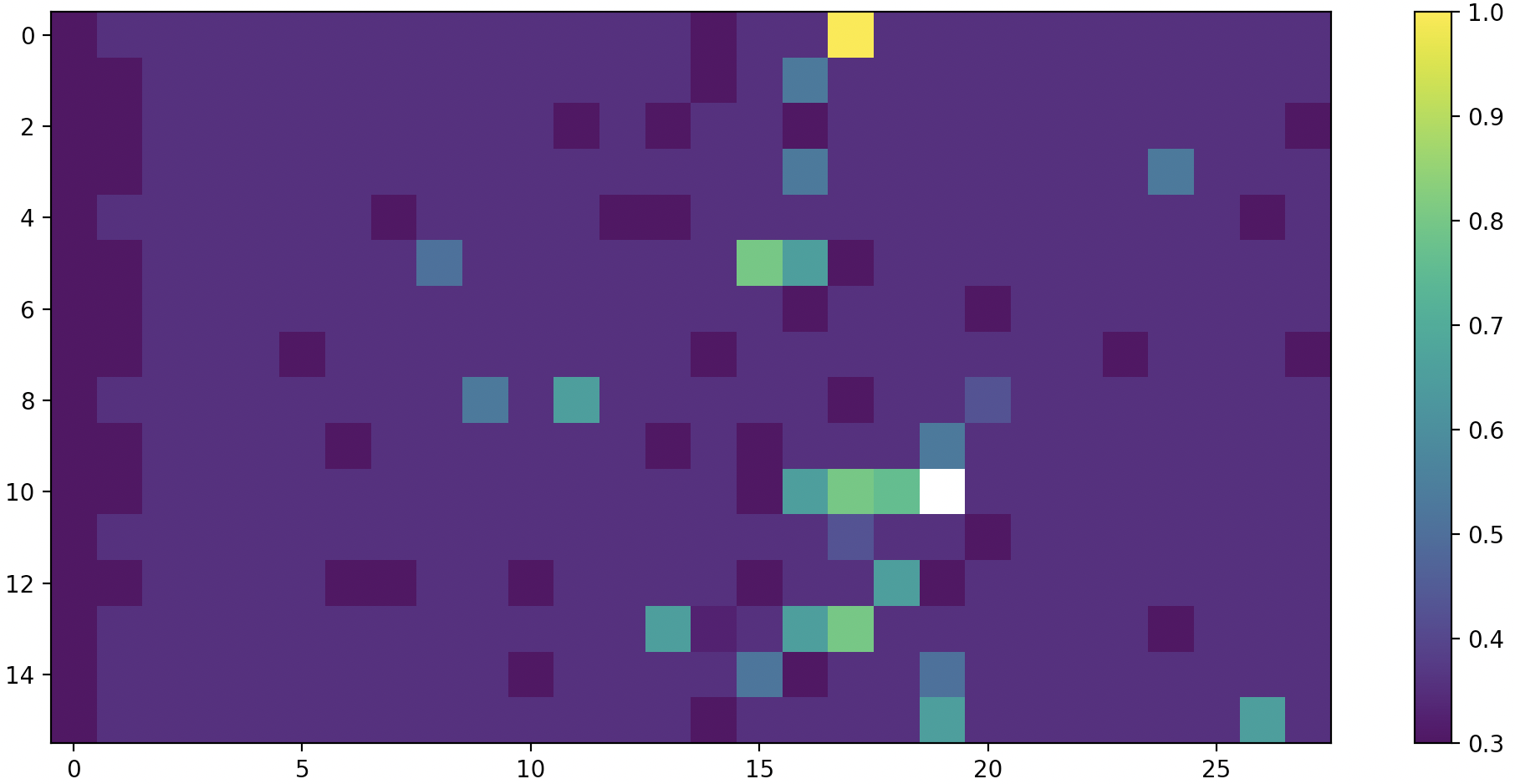}
         \caption{Pearson correlation between attention and logit similarity.}
     \end{subfigure}
\caption{Interpreting attention values from kernerl regression perspective on Llama-2 model.}
\end{figure}

\begin{figure*}[ht]
\centering
\includegraphics[width=\textwidth]{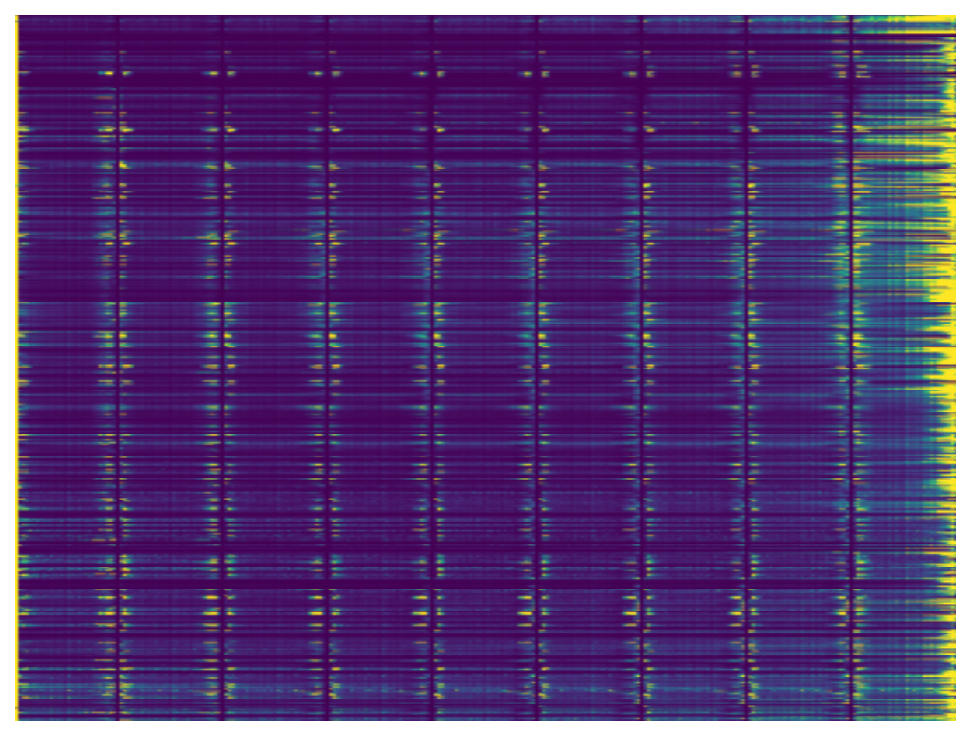}
\caption{Averaged attention map over SST2 dataset on Llama-3 8B model.}
\end{figure*}

\begin{figure}
     \centering
     \begin{subfigure}[b]{0.47\textwidth}
         \centering
         \includegraphics[width=\textwidth]{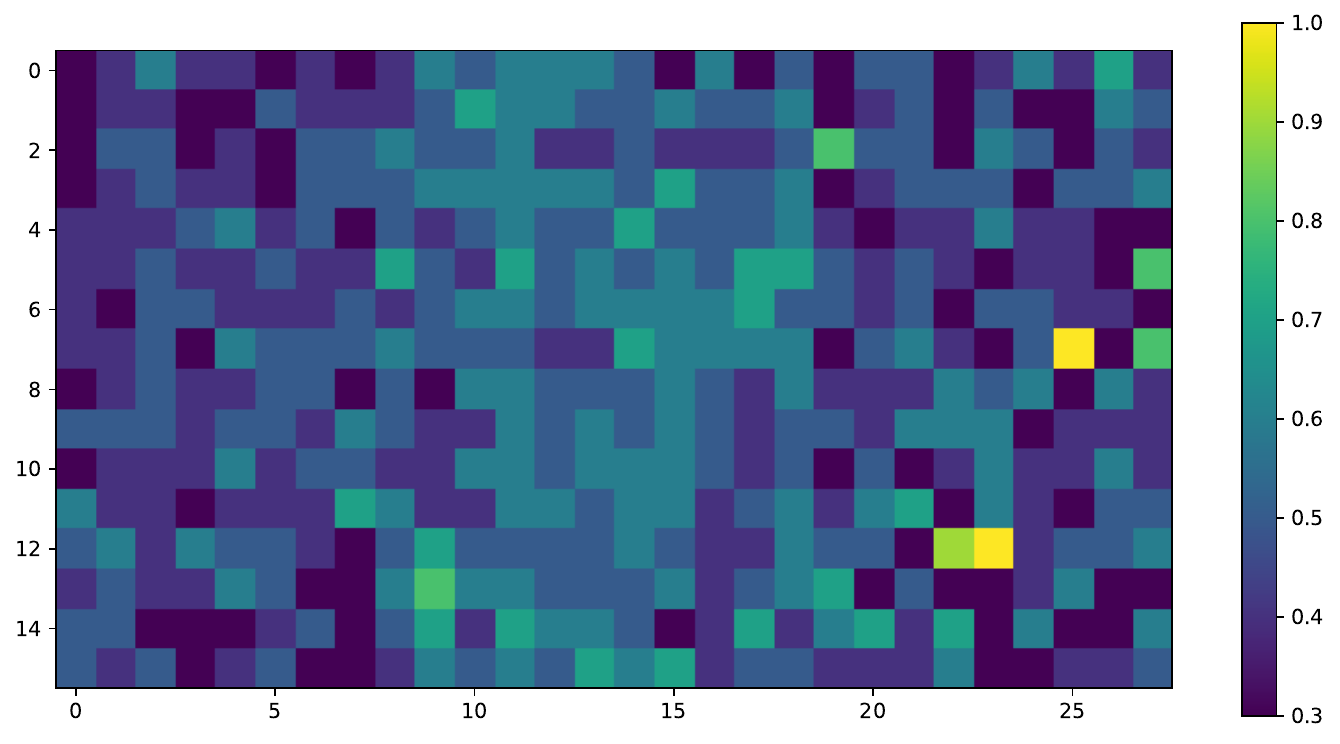}
         \caption{Accuracy on reconstruction of $\hat y$ by interpreting attention as kernel weights.}
     \end{subfigure}
     \hfill
     \begin{subfigure}[b]{0.47\textwidth}
         \centering
         \includegraphics[width=\textwidth]{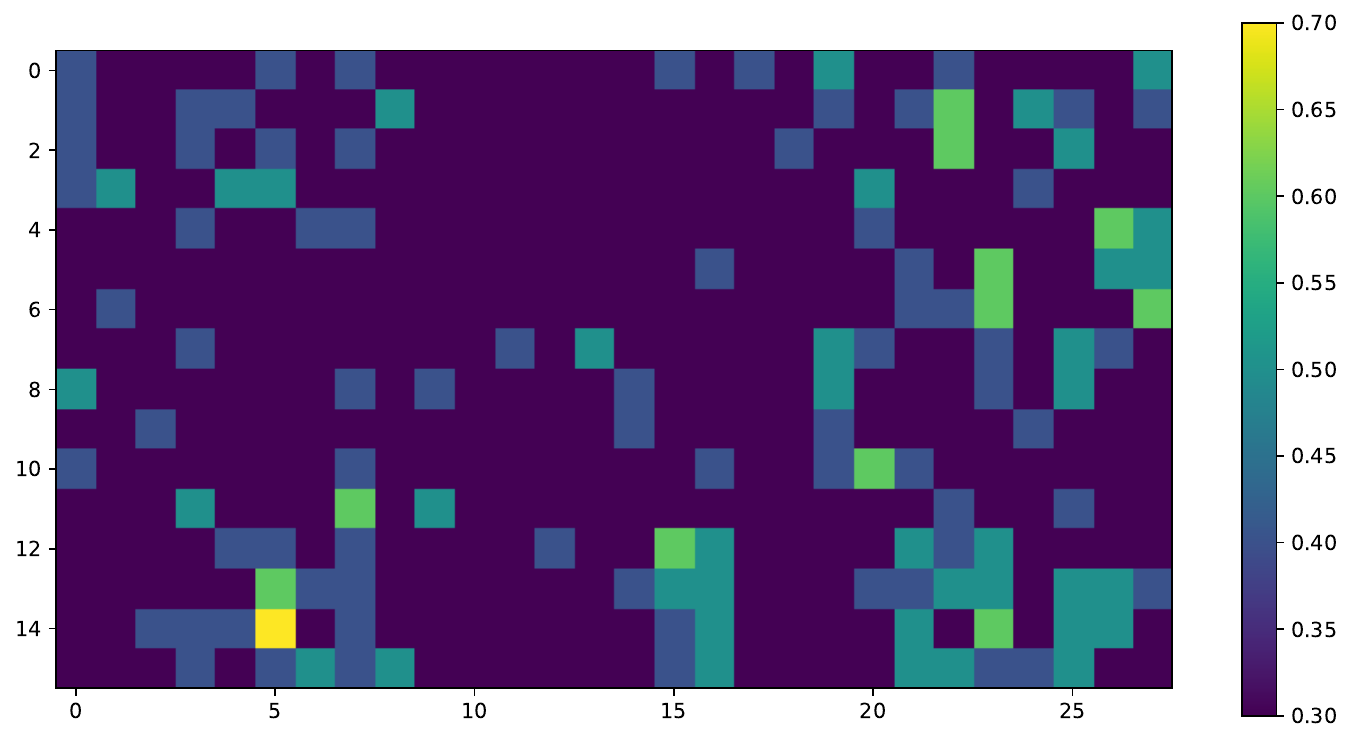}
         \caption{Accuracy of predicting the ground-truth label by interpreting attention as kernel weights.}
     \end{subfigure}
\caption{Interpreting attention values from kernerl regression perspective on Llama-3 8B model.}
\end{figure}

\newpage
\section{Single Datum Visualization}
\label{appsec:individual}
We also plot the visualization on a single datum as follows.  It shows a similar but slightly sparser pattern to Section~\ref{subsec:attention_distribution}.

\begin{figure*}[ht]
\centering
\includegraphics[width=\textwidth]{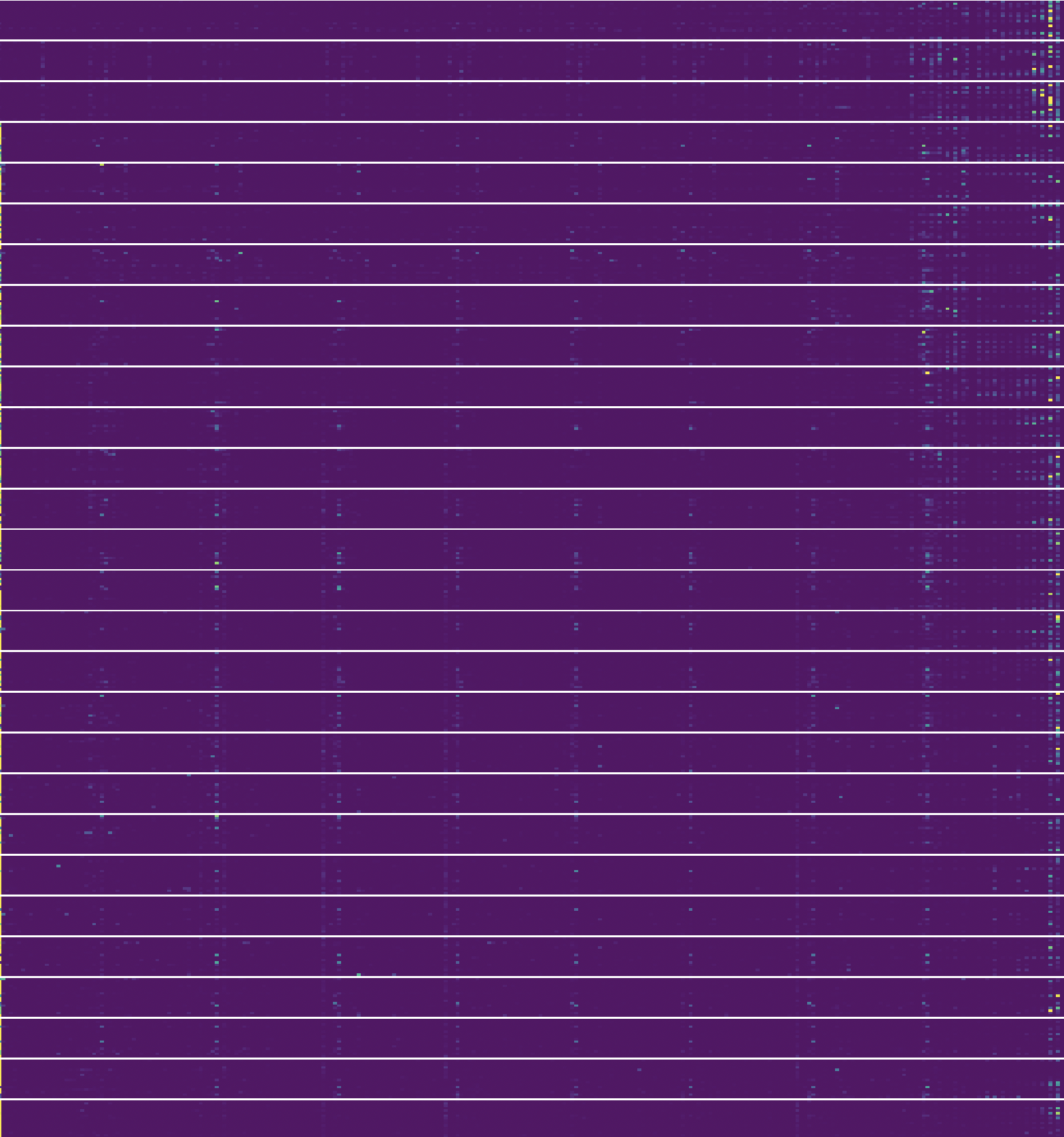}
\caption{The attention map on a single datum.}
\end{figure*}

\end{document}